\documentclass{article}

    \PassOptionsToPackage{numbers, compress}{natbib}

\usepackage[final]{neurips_2021}

\usepackage[utf8]{inputenc} 
\usepackage[T1]{fontenc}    
\usepackage{hyperref}       
\usepackage{url}            
\usepackage{booktabs}       
\usepackage{amsfonts}       
\usepackage{nicefrac}       
\usepackage{microtype}      
\usepackage{amsmath}
\usepackage{xcolor}         
\usepackage{multirow}
\usepackage{graphicx}
\usepackage[font=small,skip=2pt]{caption}
\usepackage{color}
\usepackage[font=small,skip=2pt]{subcaption}
\usepackage{xfrac}
\usepackage{enumitem}
\usepackage{nicefrac}
\usepackage{wrapfig}

\newcommand{\mi}{\mathcal{I}}
\newcommand{\mic}{{\mathcal{I}_c}}
\newcommand{\AV}[1]{{\color{blue}AV: #1}}
\newcommand{\wn}[1]{{\color{orange}[wn: #1]}}

\newtheorem{lemma}{Lemma}

\title{Controllable and Compositional Generation \\ with Latent-Space Energy-Based Models }

\author{%
  Weili~Nie \\
  NVIDIA \\
  \texttt{wnie@nvidia.com} \\
   \And
   Arash Vahdat \\
   NVIDIA \\
   \texttt{avahdat@nvidia.com} \\
   \And
   Anima Anandkumar \\
   Caltech, NVIDIA \\
   \texttt{anima@caltech.edu} \\
}

\begin{document}

\maketitle

\vspace{-6pt}
\begin{abstract}
  Controllable generation 
  is one of the key requirements for successful adoption of deep generative models in real-world applications, 
  but it still remains as a great challenge.
 In particular, the compositional ability to generate novel concept combinations is out of reach for most current models.   In this work, we use energy-based models (EBMs) to handle compositional generation over a set of attributes. To make them scalable to high-resolution image generation, we introduce an EBM in the latent space of a pre-trained generative model such as StyleGAN.
We propose a novel EBM formulation representing the joint distribution of data and attributes together, and we show how sampling from it is formulated as solving an ordinary differential equation (ODE).
   Given a pre-trained generator, all we need for controllable generation is to train an attribute classifier. Sampling with ODEs is done efficiently in the latent space and is robust to hyperparameters. Thus, our method is simple, fast to train, and efficient to sample.
   Experimental results show that our method outperforms the state-of-the-art in
  both conditional sampling and sequential editing.
   In compositional generation, our method excels at  zero-shot 
   generation of unseen attribute combinations. Also, by composing energy functions with logical operators, this work is the first to achieve such compositionality in generating photo-realistic images of resolution 1024$\times$1024.
   Code is available at \url{https://github.com/NVlabs/LACE}.

\end{abstract}

\vspace{-10pt}
\section{Introduction}
\label{sec:intro}
\vspace{-7pt}



Deep generative learning has made tremendous progress in image synthesis. 
High-quality and photorealistic images can now be generated with generative adversarial networks (GANs) \citep{goodfellow2014generative, karras2019style, Karras2020stylegan2}, score-based models \citep{ho2020denoising,song2021scorebased}, and variational autoencoders (VAEs) \citep{kingma2013auto, vahdat2020NVAE}. However, a key requirement for the success of these generative models in many real-world applications is the controllability of the generation process.
Controllable generation,
in particular the compositional ability to generate novel concept combinations, %
still remains a great challenge to current methods.

A common approach for achieving controllable generation is to train a \textit{conditional} model \citep{mirza2014conditional,kingma2014semi,choi2018stargan,park2019semantic,nie2020semi}, where the conditional information, such as semantic attributes, is specified during training to guide the generation. There are two major drawbacks with this approach: (i) 
Since the set of attributes is fixed and pre-defined, it is difficult to introduce new attributes to an existing model. This, in turn, limits the model's compositional ability for unseen attributes and their novel combinations \citep{NEURIPS2020_49856ed4}. 
(ii) Training conditional generative models from scratch for new attributes
is computationally expensive, and it can be challenging to reach to the same image quality of the uncontrollable counterparts~\citep{choi2018stargan,park2019semantic}. 


Recent works have attempted to overcome these issues by first training an unconditional generator, and then converting it to a conditional model with a small cost
~\citep{jahanian2019steerability,harkonen2020ganspace,abdal2020styleflow,patashnik2021styleclip}. 
This is often achieved by discovering semantically meaningful directions in the latent space of the unconditional model. 
This way, one would pay the most computational cost only once for training the unconditional model.
However, these approaches often struggle with compositional generation, in particular with rare combination of the attributes and the introduction of new attributes.
An appealing solution to the compositionality problem is to use energy-based models (EBMs) for controllable generation~\citep{NEURIPS2020_49856ed4,grathwohl2019your,du2020improved}. This is due to the fact that energy functions representing different semantics can be combined together to form compositional image generators. 
However, existing approaches train EBMs directly in the pixel space,
making them slow to sample and difficult to generate high-resolution images~\citep{du2020improved}.



In this paper, 
we leverage the compositionality of EBMs and the generative power of state-of-the-art pre-trained models such as StyleGANs \citep{karras2019style,Karras2020stylegan2} to achieve high-quality controllable and compositional generation.
Particularly, we propose an EBM that addresses the problem of controlling an existing generative model instead of generating images directly with EBMs or improving the sampling quality.

\textbf{Our main contributions are summarized as follows}:
\vspace{-4pt}
\begin{itemize}
[leftmargin=9mm]
    \item We propose a novel formulation of a joint EBM in the latent space of pre-trained generative models for controllable generation. In particular, the EBM training reduces to training a classifier only,
    and it can be applied to any latent-variable generative model.
    \item Based on our EBM formulation, we propose a new sampling method that relies on ordinary different equation (ODE) solvers. The sample quality is robust to hyperparameters. 
    \item We show that our method is fast to train and efficient to sample, and it outperforms previous methods 
    in both conditional sampling and sequential editing.
    
    \item 
    We show strong compositionality with zero-shot generation on unseen attribute combinaions, and with logical compositions of energy functions in generating high-quality images.
\end{itemize}
\vspace{-4pt}

Specifically, we build an EBM in the joint space of data and attributes where the marginal data distribution is denoted by an implicit distribution (e.g., a pre-trained GAN generator), and the conditional distribution of attributes, given data, is represented by an attribute classifier. Using the reparameterization trick, we show that our EBM formulation becomes an energy function in the latent space where the latent distribution is known (i.e., a standard Gaussian). This way, we only need to train the classifier in the data space and do the sampling in the latent space. Because our method only requires training a classifier to add controllability,
it is conceptually simple and fast-to-train. 

For sampling, most existing EBMs rely on 
the Langevin dynamics (LD)
sampling which is often computationally expensive and sensitive to the choice of sampling parameters. Instead, we build on the recent score-based model~\citep{song2021scorebased}, and define the sampling process in the latent space using the corresponding probability flow ODEs, induced by the reverse diffusion process. Thus, we rely on an ODE solver for sampling from our EBM formulation in the latent space. With its adaptive step size, the ODE sampling in the latent space is efficient and robust to the sampling parameters. 

In experiments, we show our method achieves  higher image quality and better controllability in both the conditional sampling and sequential editing, compared to various baselines, including StyleFlow \citep{abdal2020styleflow} and JEM \citep{grathwohl2019your}. For the training time, we show our method is $25\times$ faster than StyleFlow on the FFHQ data. For the inference time, our sampling is at least 49$\times$ and 876$\times$ faster than EBMs \citep{grathwohl2019your,NEURIPS2020_49856ed4} and score-based models \citep{song2021scorebased} in the pixel space, respectively, on CIFAR-10. 
More importantly, 
our method excels on zero-shot generation with unseen attribute combinations where StyleFlow almost completely fails. 
By composing energy functions with logical operators~\citep{NEURIPS2020_49856ed4}, our method is the first to show such compositionality in generating photo-realistic images of resolution 1024$\times$1024.

\vspace{-8pt}
\section{Method}
\vspace{-7pt}

In this section, we will first give an overview of EBMs and then propose our new EBM formulation in the latent space, based on which, we introduce a new sampling method through the ODE solver. 


\vspace{-6pt}
\subsection{Energy-based models} 
\vspace{-5pt}

EBMs \citep{lecun2006tutorial} represent data $x \in \mathbb{R}^d$ by learning an unnormalized probability distribution $p_\theta(x) \propto e^{-E_\theta(x)}$, 
where $E_\theta(x): \mathbb{R}^d \to \mathbb{R}$ is the energy function, parameterized by a neural network. 
To train an EBM on a data distribution $p_{\text{data}}$ with maximum likelihood estimation (MLE), we can estimate the gradient of the data log-likelihood $L(\theta)= \mathbb{E}_{x \sim p_{\text{data}}}[\log p_\theta(x)]$ as \citep{hinton1999prod}:
\begin{align}
    \nabla L(\theta) = \mathbb{E}_{x \sim p_{\text{data}}} [\nabla_\theta E_\theta(x)] - \mathbb{E}_{x \sim p_{\theta}} [\nabla_\theta E_\theta(x)]
\end{align}
To sample $x$ from $p_\theta$ for training and inference, Langevin dynamics (LD) \citep{welling2011bayesian} is applied as follows,
\vspace{-3pt}
\begin{align} \label{LD}
    x_0 \sim p_0(x), \quad x_{t+1} = x_t - \frac{\eta}{2} \nabla_x E_\theta(x_t) + \epsilon_t, \quad \epsilon_t \sim N(0, \eta I)
\end{align}
where $p_0$ is the initial distribution, $\eta$ is the step size. When $\eta \to 0$ and $t \to \infty$, $x_t$ is guaranteed to follow the probability $p_\theta(x)$ under some regularity conditions \citep{song2021train}.

\vspace{-6pt}
\subsection{Modelling joint EBM in the latent space}
\label{sec:modelling}
\vspace{-4pt}

To formulate our EBM for controllable generation, we consider the following setting: We have the data $x \in \mathcal{X} \subseteq \mathbb{R}^d $ and its attribute code $c = \{c_1, \cdots, c_n\} \in \mathcal{C} \subseteq \mathbb{R}^n$, which is a $n$-dimensional vector with each entry $c_i$ representing either a discrete $m_i$-class attribute where $c_i \in \{0,\cdots, m_i - 1\}$, or a continuous attribute where $c_i \in \mathbb{R}$. 
The goal is to learn a generative model and semantically control its generated samples by manipulating its conditioning attribute code $c$. 
To begin with,
we define the joint generative model on both the data and attribute as: 
\vspace{-3pt}
\begin{align}\label{joint_dist}
    p_\theta(x, c) := p_g(x) p_\theta(c|x) \propto p_g(x) e^{-E_\theta(c|x)}
\end{align}
where $p_g(x)$ is an implicit distribution defined by a pre-trained generator $g$ (such as GANs) in the form $x = g(z)$ with the latent variable $z \in \mathcal{Z}$. And, $p_\theta(c|x)$ is conditional distribution on attributes given $x$, modeled by a conditional energy function $E_\theta(c|x)$.

In this paper, we assume that the generator $g$ is fixed and we only train the energy function $E_\theta(c|x)$. We also assume the attributes are conditionally independent, i.e., $E_\theta(c|x)=\sum_{i=1}^n E_\theta(c_i|x)$. Since, our goal is to preserve the image generation quality, we design the joint in Eq.~(\ref{joint_dist}) such that the marginal data distribution $p_\theta(x)$ from our joint satisfies $p_\theta(x) = p_g(x)$.
This is obtained with an energy function that is normalized up to a constant. 
Thus, we define: 
\begin{align}\label{e_cond_dis_cont}
    E_\theta(c_i |x) = 
    \begin{cases}
    -f_i(x;\theta)[c_i] 
    + \log\sum_{c_i}\exp{(f_i(x;\theta)[c_i])}
    & \text{if $c_i$ is discrete}, \\
    \frac{1}{2 \sigma^2}( c_i - f_i(x;\theta) )^2 & \text{if $c_i$ is continuous} \\
    \end{cases}
\end{align}
where 
$f_i(x;\theta)$ is the output of a multi-class classifier mapping from $\mathcal{X}$ to $\mathbb{R}^{m_i}$ if the $i$-th attribute is discrete or a regression network mapping from $\mathcal{X}$ to $\mathbb{R}$ if it is continuous.
Without loss of generality, we will always call $f_i(x;\theta)$ a classifier.
Here, $\sigma^2$ is a hyperparameter to adjust the continuous energy. {In practice, we set $\sigma^2=0.01$ after normalizing all the continuous attributes to $[0, 1]$}.

We cannot use Eq.~(\ref{joint_dist}) for sampling, since $p_g(x)$ is defined implicitly in GAN generators. However, in Appendix \ref{app:derivation}, we show that using the reparameterization trick \citep{che2020GanIsEBM,xiao2020vaebm}, sampling $x$ from $p_\theta(x, c)$ is reparameterized to sampling $z$ from $p_\theta(z, c)\!\propto\!e^{-E_\theta ( c | g(z) ) + \log p(z)}$,
and then transferring the $z$ samples to the data space using generator $x\!=\!g(z)$. 
In most generative models, the prior distribution $p(z)$ is a standard Gaussian. Thus, the joint becomes
$p_\theta(z, c)\!\propto\!e^{-E_\theta(z,c)}$
with the energy function:
\vspace{-3pt}
\begin{align} \label{energy_final}
\small
    E_\theta(z,c) = \sum_{i} E_\theta ( c_i | g(z) ) + \frac{1}{2} \|z\|^2_2
\end{align}
where each conditional energy function $E_\theta ( c_i | g(z) )$ is given by Eq.~(\ref{e_cond_dis_cont}). For sampling, we can run LD of Eq.~(\ref{LD}) in the $z$-space using the energy function above. At the end of a chain, we pass the final $z$ samples to the generator $g$ to get the $x$ samples. 
Since our latent-space EBM is formulated for controllable generation, we term it as \textbf{LA}tent \textbf{C}ontrollable \textbf{E}BM (LACE) throughout this work.

Inspecting the joint energy $E_\theta(z, c)$ in Eq.~(\ref{energy_final}) shows that when the unconditional generator $g$ is fixed, the only trainable component is the classifier $f_i(x;\theta)$ that represents $E_\theta ( c_i | x )$.
Therefore, unlike the previous joint EBMs \citep{grathwohl2019your} that train both $p_\theta(x)$ and $p_\theta(c|x)$, our method only needs to train a classifier for the conditional $p_\theta(c|x)$, and does not require sampling with LD in the $x$-space during training. That is, for controllable generation, we can train the classifier only in the $x$-space and do the LD sampling in the $z$-space.
This fact makes our method conceptually simple and easy to train.

Note that the considered $x$-space for training the classifier does not have to be the pixel space. Instead, we can choose any intermediate layer of the pre-trained generator as the $x$-space.
Take StyleGAN2 \citep{Karras2020stylegan2} as an example, if we train the classifier in the $w$-space, then $g(z)$ in Eq. (\ref{energy_final}) actually corresponds to the mapping network of StyleGAN2, and once we obtain the $w$ samples via the LD sampler, we can pass them to the synthesis network of StyleGAN2 to get the final images.
In other words, since data $x$ is a deterministic function of $z$, the distribution of attributes is uniquely defined by $z$. We have the freedom to define $p_\theta ( c_i | z)$ by applying the classifier on top of $z$ or any representation extracted from $z$ including the generator $g(z)$ or its intermediate representations such as $w$-space in StyleGAN2.

\vspace{-6pt}
\subsection{Sampling through an ODE solver}
\label{sec:ode}
\vspace{-4pt}

Sampling with LD requires hand-crafted tricks to speed up and stabilize its convergence \citep{du2019implicit,grathwohl2019your}.
Even in the latent space, our experiments show that LD tends to be sensitive to its hyperparameters. In this section, we introduce another sampling method, called \textit{ODE sampler}, that relies on an ODE solver. Our method with the LD and ODE sampler is termed as \textit{LACE-LD} and \textit{LACE-ODE}, respectively.


Our idea is inspired by the prior work \citep{song2021scorebased}, which shows that controllable generation can be achieved by solving a conditional reverse-time SDE. 
Specifically,
if we consider a \textit{Variance Preserving (VP) SDE} \citep{song2021scorebased}, the forward SDE is defined as 
$ dx = -\frac{1}{2}\beta(t) x dt + \sqrt{\beta(t)} dw $,
where $w$ is a standard Wiener process and $x_0 \sim p_{\text{data}}$ is a data sample. The scalar time-variant diffusion coefficient $\beta(t)$ has a linear form $\beta(t) = \beta_{\min} + (\beta_{\max} - \beta_{\min}) t$, where $t \in [0, T]$. 
Then, the
conditional sampling from $p_0(x|c)$ is equivalent to solving the following reverse VP-SDE:
\vspace{-3pt}
\begin{align}\label{eq:reverse_sde}
    dx = -\frac{1}{2}\beta(t)[x + 2\nabla_x \log p_t(x, c)]dt + \sqrt{\beta(t)} d\bar{w}
\end{align}
where $\bar{w}$ is a standard reverse Wiener process when time flows backwards from $T$ to 0, and $p_t(x, c)$ is the join data and attribute distribution at time $t$. Song et al.~\cite{song2021scorebased} show that there exist a corresponding ODE for the SDE in Eq.~(\ref{eq:reverse_sde}) which can be used for sampling. In Appendix \ref{app:derivation_ode}, using the reparameterization trick, we show that the ODE in the latent space for our model takes a simple form:
\vspace{-3pt}
\begin{align}\label{ode_final}
    \begin{split}
        dz = \frac{1}{2} \beta(t) \sum_{i} \nabla_z E( c_i| g(z) ) dt
    \end{split}
\end{align}
with negative time increments from $T$ to 0. To generate a sample $x$ conditioned on $c$, all we need is to draw $z(T)\!\sim\!N(0, I)$ and solve the ODE in Eq.~(\ref{ode_final}) using a black-box solver. The final $z(0)$ samples are transferred to the data space through the generator $x\!=\!g(z(0))$. Our main insight in Appendix \ref{app:derivation_ode} that leads to the simple ODE is that the latent $z$ in most generative models follows a standard Gaussian distribution, and diffusing it with VP-SDE will not change its distribution, i.e., $p_t{(z(t))}\!=\!\mathcal{N}(0, I)$.


\vspace{-4pt}
\paragraph{Remarks} There are some key observations from our ODE formulation in Eq. (\ref{ode_final}):

\textit{(i) Connections to gradient flows. } Similar to \citep{santambrogio2017euclidean,ansari2020refining} that build connections between SDE/ODE and gradient flows, our ODE sampler can be seen as a gradient flow that refines a latent $z$ from a random noise to a $z$ vector conditioned on an attribute vector. While \cite{ansari2020refining} relied on the Euler-Maruyama solution of the SDE, we convert our generative SDE to an ODE. 
Adaptive discretization with a higher order method (e.g., Runge-Kutta~\citep{chen2018neuralode}) is often preferred when solving ODEs. But our ODE formulation also works well with the first-order Euler method, and its performance lies in-between LACE-LD and LACE-ODE (See Appendix \ref{app:abl_euler}).

\textit{(ii) Advantages of  ODE in the latent space.} If the SDE/ODE is built in the pixel space as in \cite{song2021scorebased}, it requires 1) estimating the score function $\nabla_x \log p_t(x(t))$, and 2) training a \textit{time-variant} classifier $p_t(c|x(t))$, both of which make its training and inference challenging. Instead, our ODE sampler is much simpler: we do not need to train any score function. We only train a {time-invariant} classifier. 
Compared with the LD sampler that uses a fixed step size and is sensitive to many hyperparameters (e.g., step size, noise scale and number of steps), our ODE sampler is adaptive in step sizes and only needs to tune the tolerances, making it more efficient and robust to hyperparameters.


\vspace{-8pt}
\section{Experiments}
\label{sec:exp}
\vspace{-6pt}

In this section, we show the effectiveness of our method in conditional sampling, sequential editing and compositional generation, and we also perform an ablation study on the sampling method.

\vspace{-6pt}
\paragraph{Experimental setting}
We use StyleGAN-ADA~\citep{NEURIPS2020_8d30aa96} as the pre-trained model for experiments on CIFAR-10~\citep{krizhevsky2009learning}, and StyleGAN2~\citep{Karras2020stylegan2} for experiments on FFHQ~\citep{karras2019style}.
We train the classifier in the $w$-space, where our method works best (see ablation studies in Appendix \ref{app:which_space}).
To train the latent classifier in the 
$w$-space, we first generate (image, 
$w$) pairs from StyleGAN, and then label each 
$w$ latent by annotating its paired image with an image classifier (see data preparation in Appendix \ref{app:exp_setting}). 

For LACE-ODE, we use the `dopri5' solver \citep{chen2018neuralode} with the tolerances of (1e-3, 1e-3), and we set $T=1$, $\beta_{\min} = 0.1$ and $\beta_{\max} = 20$.
For LACE-LD, the step size $\eta$ and the standard deviation $\sigma$ of $\epsilon_t$ are chosen separately for faster training \citep{grathwohl2019your}, where the number of steps $N=100$, step size $\eta=0.01$ and standard deviation $\sigma=0.01$.
For metrics, we use (i) \textit{conditional accuracy} (\textit{ACC}) to measure the controllability, where we generate images using  randomly sampled attribute codes, and pass them to a pre-trained image classifier to predict the attribute codes, and (ii) \textit{FID} to measure the generation quality
and diversity 
\citep{heusel2017gans}. See Appendix \ref{app:exp_setting} for more details.

\vspace{-6pt}
\subsection{Conditional sampling}
\vspace{-4pt}

\paragraph{Baselines}
For comparison, we consider a set of baselines: StyleFlow \citep{abdal2020styleflow}, JEM \citep{grathwohl2019your}, Conditional EBM (Cond-EBM) \citep{NEURIPS2020_49856ed4}, and SDEs \citep{song2021scorebased}. 
We also propose \textit{Latent-JEM}, a variant of JEM modelled in the latent space, and \textit{LACE-PC}, which replaces the ODE sampler with the Predictor-Corrector (PC) sampler that solves the reverse SDE 
(Eq.~\ref{eq:reverse_sde})~\citep{song2021scorebased}. See the Appendix \ref{app:baselines} for more details.

\begin{figure}[t]
\vspace{-0.6cm}
  \begin{minipage}[b]{0.49\textwidth}
    \centering
    \captionof{table}{\small Comparison of our method and baselines for conditional sampling on CIFAR-10.
    For notations, Train -- training time, Infer -- inference time (m: minute, s: second), which refer to the single GPU time for generating a batch of 64 images,
    $\eta$ is the LD step size, and $N$ is the number of predictor steps in the PC sampler. 
    }
\footnotesize\addtolength{\tabcolsep}{-4pt}
  \begin{subtable}[t]{\textwidth}
    \centering
    \begin{tabular}{c|cccc}
        \hline
         Methods & Train & Infer & FID$\downarrow$ & ACC$\uparrow$ \\
         \hline
          JEM \citep{grathwohl2019your} &
          2160m & 135s & 52.35 & 0.645
          \\
          Cond-EBM \citep{NEURIPS2020_49856ed4} & 
          2280m & 24.5s & 41.72 & 0.792 
          \\
          VP-SDE \citep{song2021scorebased} & 
          52800m & 438s & 19.13 & 0.643
          \\
          VE-SDE \citep{song2021scorebased} & 
          52800m & 448s  & \textbf{2.97} & 0.662 
          \\
\hline
        Latent-JEM ($\eta$=0.1) & 
        21m & 0.63s 
        & 8.75 & 0.950
        \\
        Latent-JEM ($\eta$=0.01) & 
        21m & 0.63s 
        & 5.65 & 0.821
        \\
\hline
\hline
        LACE-PC ($N$=100) &
        \textbf{4m} & 0.84s & \textbf{2.99} & 0.747
        \\
        LACE-PC ($N$=200) &
        \textbf{4m} & 1.86s  & \textbf{2.94} & 0.722
        \\
        LACE-LD &
        \textbf{4m} & 0.68s & 4.30 & 0.939
        \\
        LACE-ODE & 
        \textbf{4m} & \textbf{0.50s} & 6.63 & \textbf{0.972}
        \\
         \hline
    \end{tabular}
\end{subtable}
      \label{baselines_cifar10}
  \end{minipage}
    \hfill
  \begin{minipage}[b]{0.49\textwidth}
    \centering
    \includegraphics[width=\linewidth]{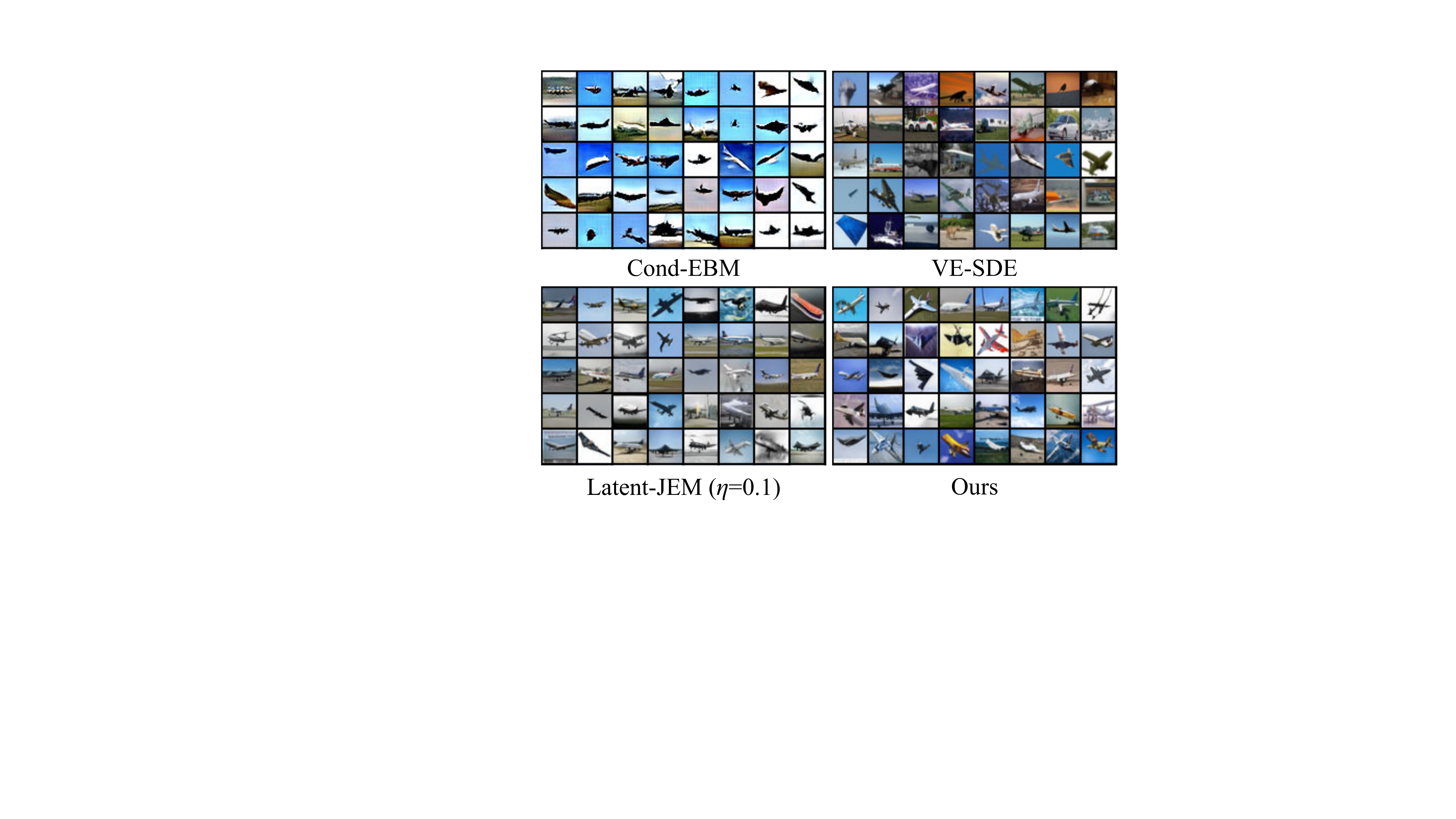}
    \captionof{figure}{\small Conditionally generated images of our method (LACE-ODE) and baselines on the plane class of CIFAR-10. See Appendix \ref{app:cond_cifar} for more results. 
    }
    \label{cifar10_visual}
  \end{minipage}
  \vspace{-9pt}
\end{figure}

\vspace{-6pt}
\paragraph{CIFAR-10} 
The results of our method against baselines on CIFAR-10 are shown in Table \ref{baselines_cifar10} and Figure \ref{cifar10_visual}. 
Our method requires only four minutes to train, and it takes less than one second to sample a batch of 64 images on a single NVIDIA V100 GPU, which significantly outperforms previous EBMs (at least 49$\times$ faster) and score-based models (at least 876$\times$ faster) in the pixel space. 
Latent-JEM that works in the latent space is also slower in training as it performs the LD for each parameter update.

For the conditional sampling performance,
LACE-ODE and LACE-LD largely outperform all the baselines in precisely controlling the generation while maintaining the relatively high image quality. 
In particular, 
LACE-PC performs similarly to VE-SDE: they can achieve better image quality but have a problem with precisely controlling the generation (ACC$\leq$0.75).
Besides, we observe that EBMs in the latent space achieve much better overall performance than EBMs in the pixel space.

\vspace{-6pt}
\paragraph{FFHQ}
To test on FFHQ,
we use 10k ($w$, $c$) pairs created by \cite{abdal2020styleflow} for training, 
where 
$w$
is sampled from the $w$-space of StyleGAN2,
and 
$c$ 
is the attribute code.
Unless otherwise specified, we use truncation $\psi=0.7$ for our method.
Following the evaluation protocol from \citep{abdal2020styleflow}, we use 1k generated samples
from StyleGAN2 to compute the FID. Note that given the small sample size, FID values tend to be high.
For the reference, the original unconditional StyleGAN2 with 1k samples has FID 20.87.

\begin{figure}[t]
\vspace{-0.6cm}
  \centering
  \begin{subfigure}[c]{\textwidth}
		\centering
		\includegraphics[width=\linewidth]{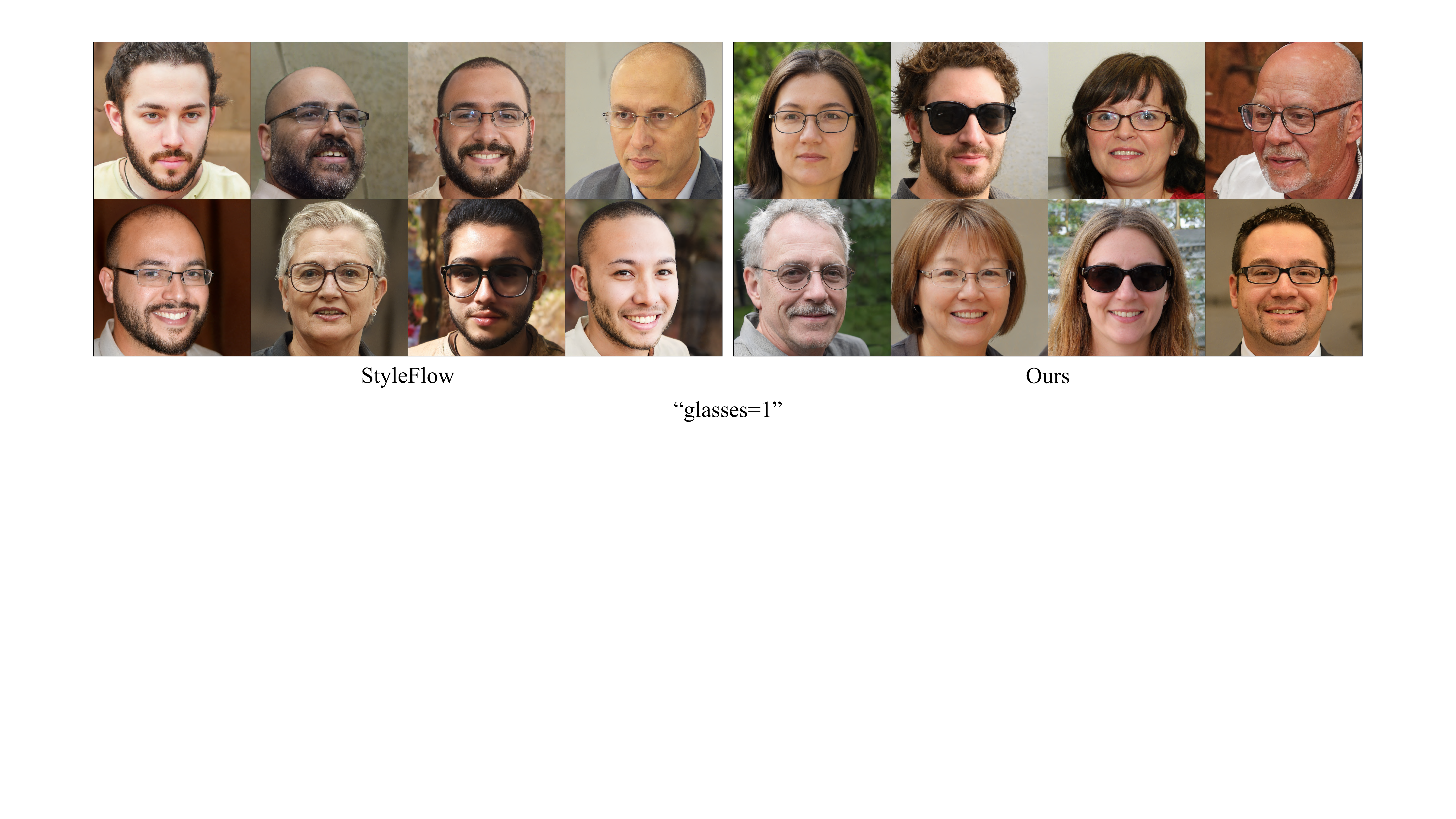}
		\caption{ \underline{\texttt{glasses=1}} (\textit{Left}: StyleFlow, \textit{Right}: Ours)}
  \end{subfigure}
  \begin{subfigure}[c]{\textwidth}
		\centering
		\includegraphics[width=\linewidth]{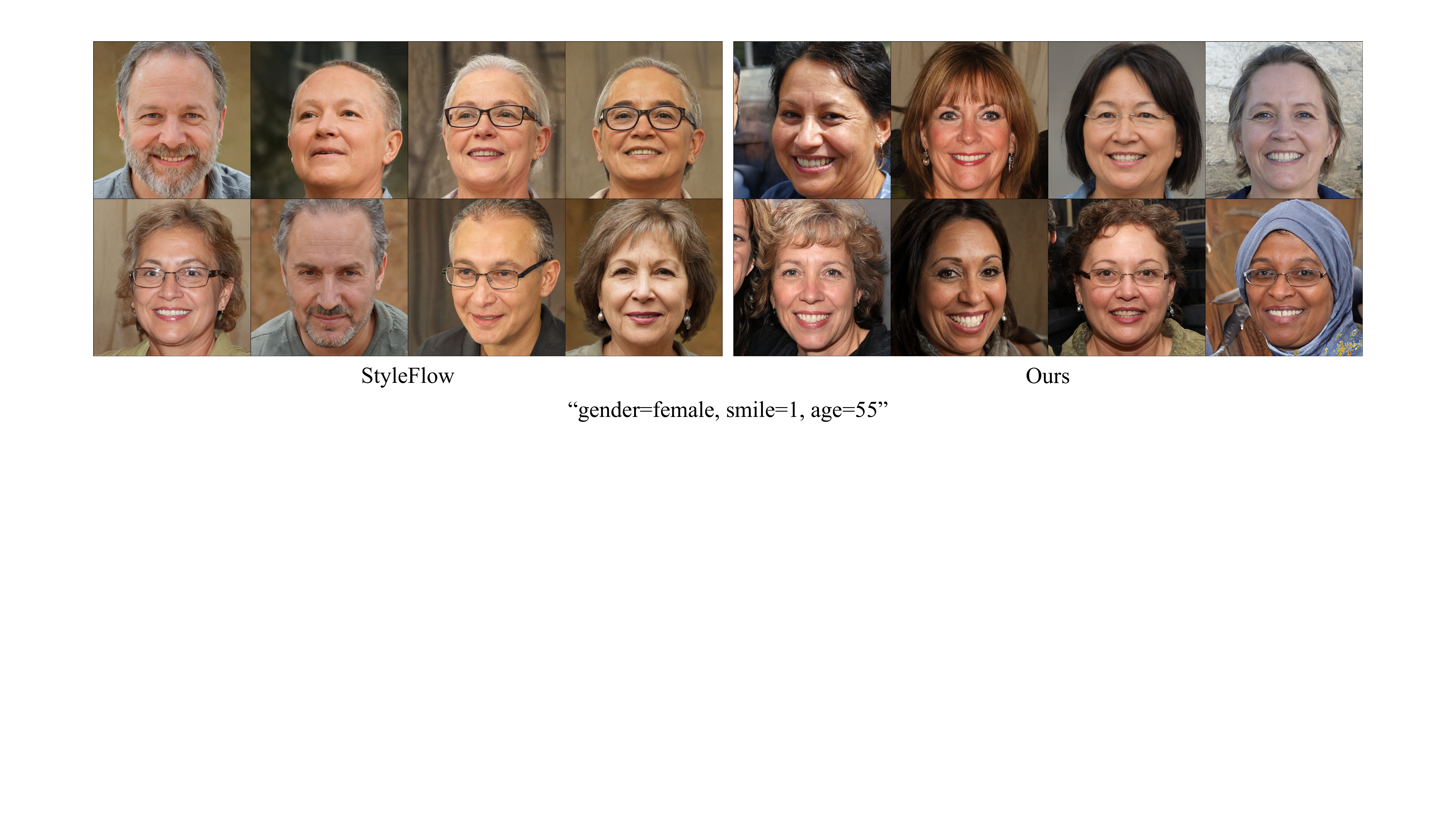}
		\caption{ \underline{\texttt{gender=female,smile=1,age=55}} (\textit{Left}: StyleFlow, \textit{Right}: Ours)}
  \end{subfigure}
  \caption{\small Conditionally generated images of our method (LACE-ODE) and StyleFlow \citep{abdal2020styleflow} on the \texttt{glasses} and \texttt{gender\_smile\_age} of FFHQ, respectively.
  Visually, our method achieves higher image generation quality, more diversity and better controllability than StyleFlow. }
  \vspace{-10pt}
  \label{ffhq_visual}
\end{figure}

The results of comparing our method with baselines are shown in Table \ref{baselines_ffhq} and Figure \ref{ffhq_visual}, where we condition on \texttt{glasses} and \texttt{gender\_smile\_age}, respectively. 
For the training time, our method only takes 2 minutes, which is around $25\times$ faster than StyleFlow, and $5\times$ faster than Latent-JEM. 
Our inference time increases with the number of attributes to control.
For instance, LACE-ODE needs similar inference time with StyleFlow (0.68s vs. 0.61s) on \texttt{glasses}, but more inference time than StyleFlow (4.81s vs. 0.61s) on \texttt{gender\_smile\_age}. 
In practice, we could adjust the ODE tolerances,
allowing for trade-offs between inference time and overall performance.
Besides, we can also optimize the network to further reduce the inference time
(see results in Appendix \ref{app:cond_ffhq}).

For the controllability, both LACE-ODE and LACE-LD outperform the baselines by a large margin.
Also, the generation quality of our method is on par with our proposed baselines, and much better than the prior work StyleFlow. For instance, LACE-ODE has much lower FID than StyleFlow (24.52 vs. 43.55). 
Latent-JEM achieves better FID but always has worse controllability.
These quantitative results could be verified by the visual samples in Figure \ref{ffhq_visual}, where our method achieves high-quality controllable generation. However, StyleFlow has difficulty with (i) the full controllability in different cases, and (ii) the lack of image diversity specifically when conditioning on more attributes. 

\begin{table}[t]
\vspace{-5pt}
    \caption{\small Comparison between our method and baselines for conditional sampling on the \texttt{glasses} and \texttt{gender\_smile\_age} of FFHQ, respectively. For notations, Train -- training time, Infer -- inference time (m: minute, s: second), which refer to the single GPU time for generating a batch of 16 images, 
    $\eta$ is the LD step size, and $N$ is the number of predictor steps in the PC sampler. 
    }
    \vspace{1pt}
\begin{subtable}[t]{\textwidth}
    \centering
    \footnotesize\addtolength{\tabcolsep}{-2pt}
    \begin{tabular}{c|c|*{3}{c}|*{5}{c}}
        \hline
         
      \multirow{2}{*}{Methods} &
      \multirow{2}{*}{Train} &
      \multicolumn{3}{c|}{\texttt{glasses}} &
      \multicolumn{5}{c}{\texttt{gender\_smile\_age}} \\
      \cline{3-10} 
      & & Infer &
      FID$\downarrow$ &
      $\text{ACC}_{\text{gl}}$$\uparrow$ & Infer & FID$\downarrow$ &
      $\text{ACC}_{\text{ge}}$$\uparrow$ & $\text{ACC}_{\text{s}}$$\uparrow$ & $\text{ACC}_{\text{a}}$$\uparrow$  
      \\
         \hline
          StyleFlow \citep{abdal2020styleflow} &
          50m & \textbf{0.61s} & 42.08 & 0.899 
          & \textbf{0.61s} & 43.88 & 0.718 & 0.870 & 0.874 
          \\
        \hline
        Latent-JEM ($\eta$=0.1) & 
        15m & 0.69s & 22.83 & 0.765 & 
        0.93s & 22.74 & 0.878 & 0.953 & 0.843
        \\
        Latent-JEM ($\eta$=0.01) & 
        15m & 0.69s & 21.58 & 0.750 
        & 0.93s & \textbf{21.98} & 0.755 & 0.946 & 0.831
        \\
\hline
\hline
        LACE-PC ($N$=100) &
        \textbf{2m} & 1.29s & 21.48 & 0.943  
        & 2.65s & 24.31 & 0.951 & 0.922 & 0.896  
        \\
        LACE-PC ($N$=200) &
        \textbf{2m} & 2.20s & 21.38 & 0.925
        & 4.62s & 23.86 & 0.949 & 0.914 & 0.894
        \\
        LACE-LD &
        \textbf{2m} & 1.15s & \textbf{20.92} & \textbf{0.998} 
        & 2.40s & 22.97 & 0.955 & 0.960 & \textbf{0.913}
        \\
        LACE-ODE & 
        \textbf{2m} & {0.68s} & \textbf{20.93} & \textbf{0.998} 
        & 4.81s & 24.52 & \textbf{0.969} & \textbf{0.982} & \textbf{0.914}
        \\
         \hline
    \end{tabular}
\end{subtable}
    \label{baselines_ffhq}
    \vspace{-12pt}
\end{table}

\vspace{-6pt}
\subsection{Sequential editing}
\vspace{-4pt}

In sequential editing, we semantically edit the images by changing an attribute each time without affecting other attributes and face identity. 
Given a sequence of attributes $\{c_1, \cdots, c_n\}$, we adapt our method for sequential editing by relying on the compositionality of energy functions. We define the joint energy function of the $i$-th edit as
\begin{align*}
\small
    E^{\text{seq}}_\theta(z, \{c_j\}_{j=1}^i) :=
    E_\theta(z, \{c_j\}_{j=1}^i) 
    + \mu \tilde{d}(z,z_{i-1}) + \gamma \sum\nolimits_{j > i}
    d \left( f_j(g(z); \theta), f_j(g(z_{i-1}); \theta) \right)
\end{align*}
where the joint energy function $E_\theta(z, \{c_j\}_{j=1}^i)$
is from
Eq.~(\ref{energy_final}), and 
$f_j(\cdot; \theta)$ is the classifier output for the $j$-th attribute,
$\tilde{d}(z,z_{i-1}) = \left\| g(z) - g(z_{i-1}) \right\|_2^2 + \left\| z - z_{i-1}) \right\|_2^2$  prevents $z$ from moving too far from the previous $z_{i -1}$, and the last term penalizes $z$ for changing other attributes, with $d(\cdot,\cdot)$ defined as the squared L2 norm.
Note that in the $i$-th edit, we use $z_{i-1}$ as the new initial point of the sampling for a faster convergence. By default, we set $\mu=0.04$ and $\gamma=0.01$.

We compare our method against StyleFlow \citep{abdal2020styleflow}, the state-of-the-art in sequential editing, with
two additional metrics that quantify the disentanglement of the edits: (i) the \textit{face identity loss} (\textit{ID}) \citep{abdal2020styleflow,patashnik2021styleclip}, which calculates the distance between the image embeddings before and after editing to measure the identity preservation. 
(ii) the \textit{disentangled edit strength} (\textit{DES}), defined as $\text{DES} = \frac{1}{n}\sum_{i=1}^n \text{DES}_i$ and $\text{DES}_i = \mathbb{E}_{p_\theta(x)} [\Delta_i - \max_{j \neq i} |\Delta_j|] $, where $\Delta_i = \frac{\text{ACC}_i - \text{ACC}_{0i}}{1 -  \text{ACC}_{0i}}$ denotes the \textit{normalized} ACC improvement, with $\text{ACC}_i$ and $\text{ACC}_{0i}$ being the ACC score of $i$-th attribute after and before the $i$-th edit, respectively. 
Intuitively, the maximum DES is achieved when each edit precisely control the considered attribute only but leave other attributes unchanged, leading to good disentanglement. 

\begin{figure}[t]
  \centering
  \begin{subfigure}[c]{0.82\textwidth}
		\centering
		\includegraphics[width=\linewidth]{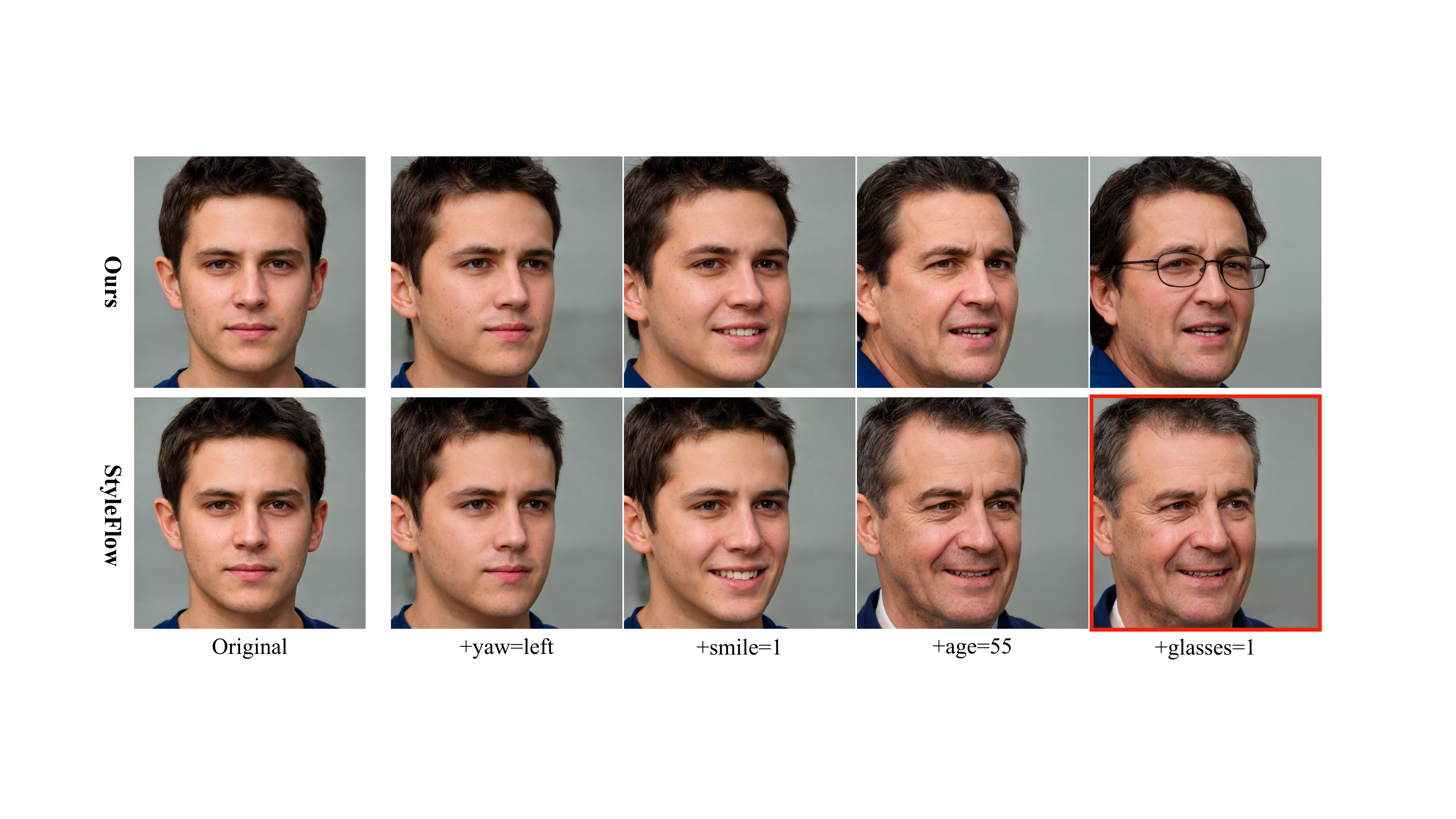}
  \end{subfigure}
  \begin{subfigure}[c]{0.82\textwidth}
		\centering
		\includegraphics[width=\linewidth]{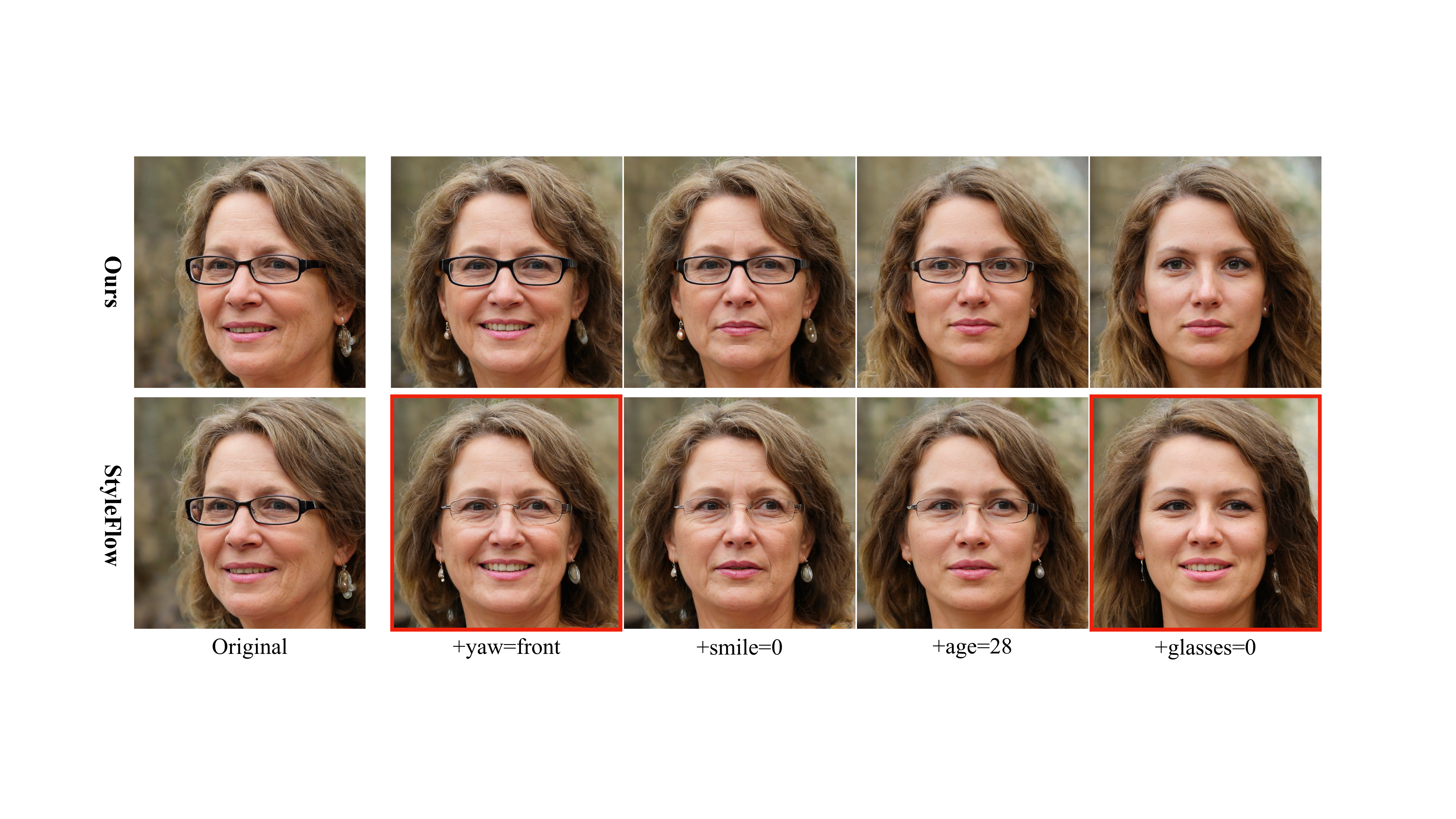}
  \end{subfigure}
  \caption{\small Sequentially editing images with our method (LACE-ODE) and StyleFlow \citep{abdal2020styleflow} on FFHQ with a sequence of [\texttt{yaw}, \texttt{smile}, \texttt{age}, \texttt{glasses}]. 
  Our method can successfully perform each edit while less affecting the other attributes. On the contrary, StyleFlow may unintentionally modify/lose glasses, largely change the face identity, or have a smiling face when it is not supposed to.
  }
  \vspace{-10pt}
  \label{ffhq_seq_visual}
\end{figure}

Table \ref{tab_seq_edit} shows the quantitative results, where the truncation coefficient $\psi=0.5$, and we apply the subset selection strategy as proposed in StyleFlow \citep{abdal2020styleflow} to alleviate the background change and the reweighting of energy functions (see Appendix \ref{app:exp_setting} for details) for better disentanglement. Our method outperforms StyleFlow in terms of disentanglement (DES), identity preservation (ID), image quality (FID) and controllability (ACC).
This is confirmed by the qualitative results in Figure \ref{ffhq_seq_visual}, where StyleFlow usually has the following issues: (i) \textit{changing unedited attributes}, such as accidentally modifying or losing glasses and changing  face identities, and (ii) \textit{having incomplete edits}, for example, the smiling face still appears after setting \texttt{smile}=0 (last row in Figure \ref{ffhq_seq_visual}). On the contrary, our method suffers less from the above problems. 
Moreover, as the sequential editing in our method is  defined by simply composing energy functions, it can incorporate novel attributes in a plug-and-play way.

\begin{table}[t]
\vspace{-0.2cm}
    \caption{\small Comparison between our method and StyleFlow \citep{abdal2020styleflow} for sequential editing on FFHQ, where we edit each attribute in the sequence of [\texttt{yaw}, \texttt{smile}, \texttt{age}, \texttt{glasses}]. Note that here we only show
    the final performance \textit{after all edits}, and the results of every individual edit 
    and ablation studies are deferred to Appendix \ref{app:seq}.
    }
    \vspace{1pt}
\begin{subtable}[t]{\textwidth}
    \centering
    \footnotesize\addtolength{\tabcolsep}{0pt}
    \begin{tabular}{c|ccccccc}
        \hline
         
      \multirow{2}{*}{Methods} &
      \multicolumn{7}{c}{All (\texttt{yaw\_smile\_age\_glasses}) } \\
      \cline{2-8} 
      & $\text{DES}$$\uparrow$ & ID$\downarrow$ & FID$\downarrow$ & $\text{ACC}_{\texttt{y}}$$\uparrow$ & $\text{ACC}_{\texttt{s}}$$\uparrow$ &  $\text{ACC}_{\texttt{a}}$$\uparrow$ & $\text{ACC}_{\texttt{g}}$$\uparrow$ \\
         \hline
          StyleFlow \citep{abdal2020styleflow} &
         0.569 & {0.549} & 44.13 &
         \textbf{0.947} &
         0.773 &
         0.817 &
         0.876
          \\
        LACE-ODE & 
        \textbf{0.735} & \textbf{0.501}  & \textbf{27.94} & 0.938 &
        \textbf{0.956} & \textbf{0.881} & \textbf{0.997}
        \\
         \hline
    \end{tabular}
\end{subtable}
    \label{tab_seq_edit}
    \vspace{-8pt}
\end{table}

\vspace{-6pt}
\subsection{Compositional generation}
\label{sec:comp_gen}
\vspace{-3pt}

\paragraph{Zero-shot generation}

\begin{figure}[t]
\vspace{-12pt}
  \centering
  \begin{subfigure}[c]{\textwidth}
		\centering
		\includegraphics[width=\linewidth]{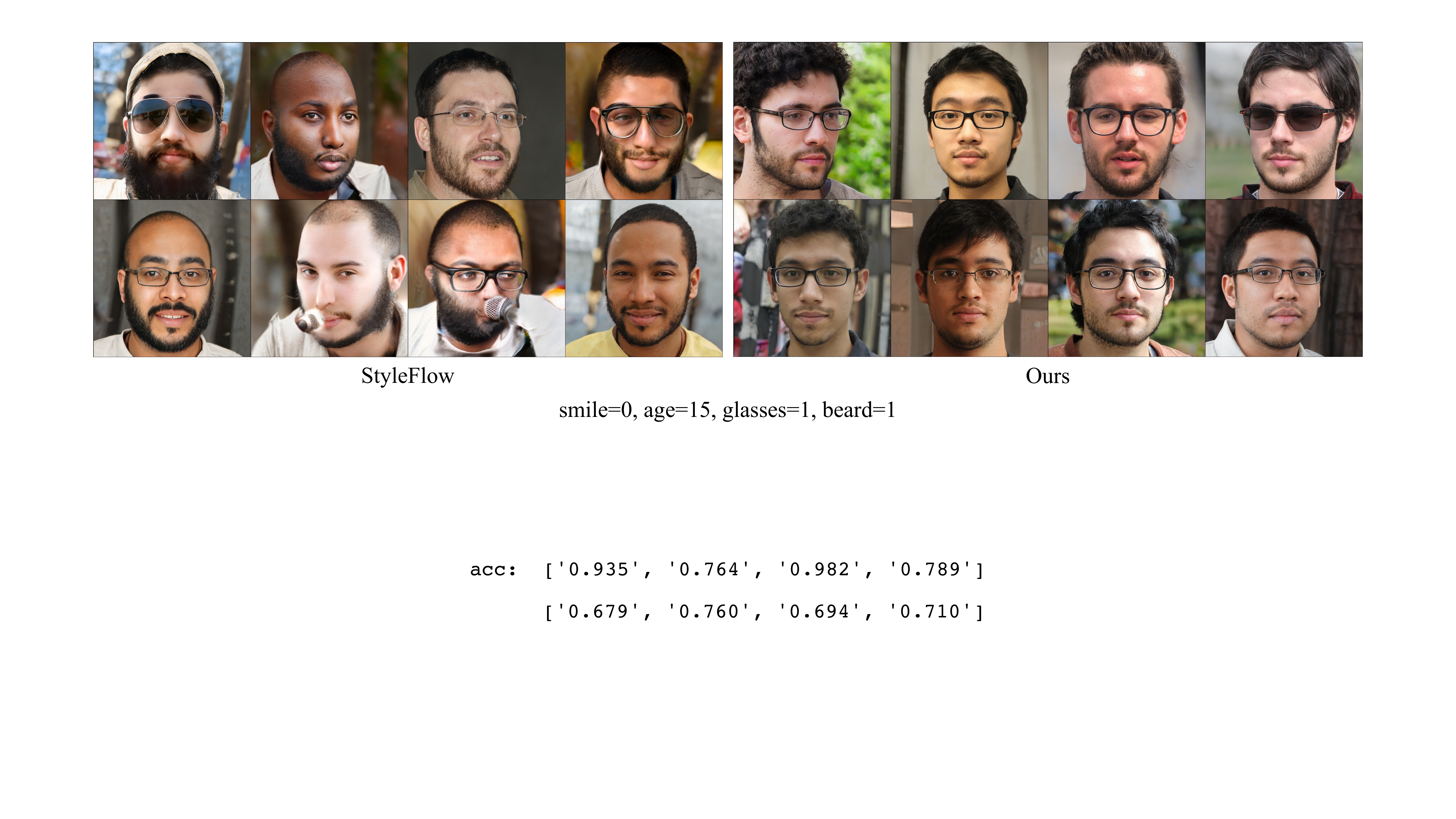}
		\caption{\small 
		\underline{\texttt{beard=1,smile=0,glasses=1,age=15}}
		(\textit{Left}: StyleFlow, \textit{Right}: Ours)
		}
  \end{subfigure}
  \begin{subfigure}[c]{\textwidth}
		\centering
		\includegraphics[width=\linewidth]{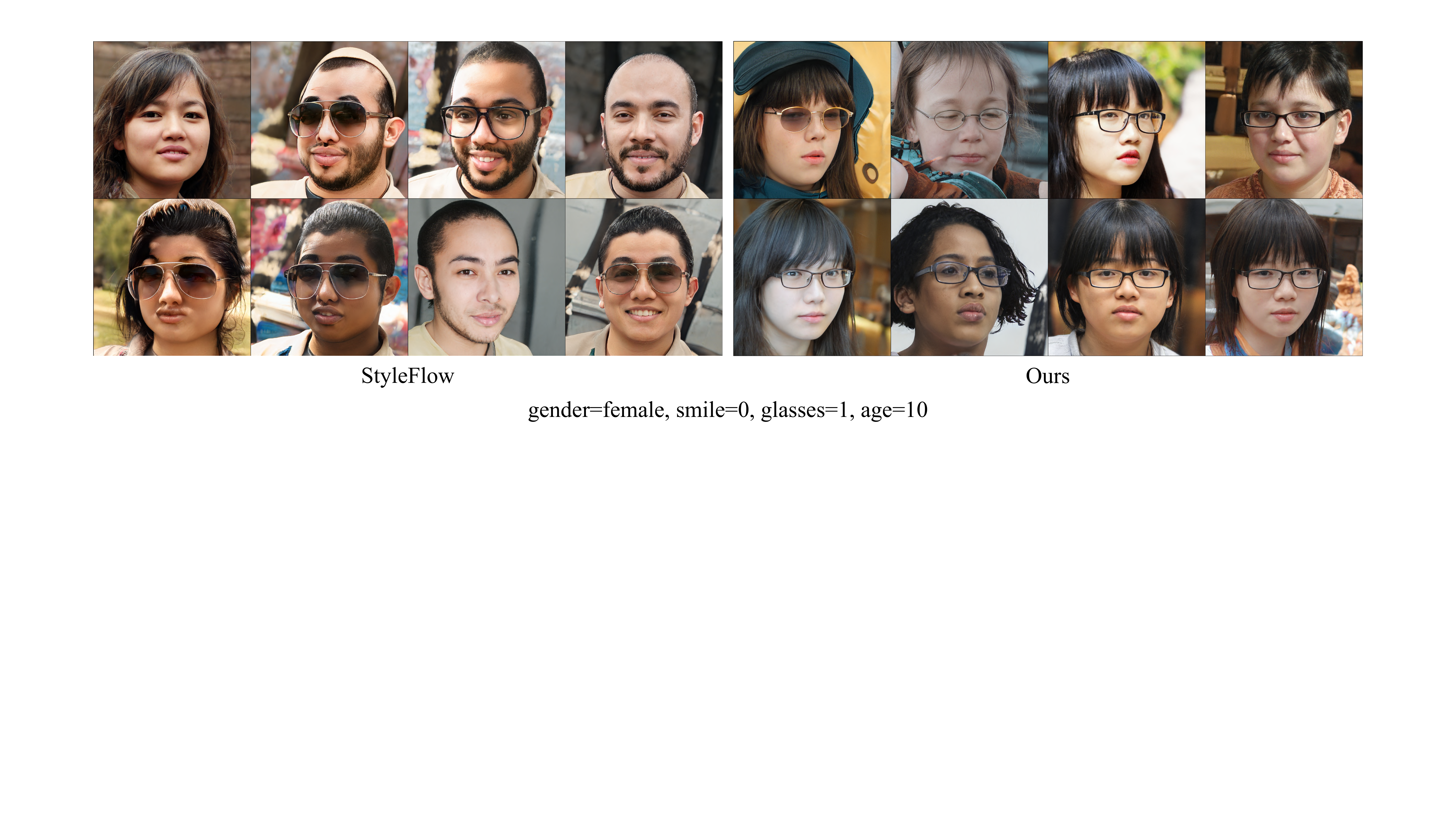}
		\caption{\small \underline{\texttt{gender=female,smile=0,glasses=1,age=10}} (\textit{Left}: StyleFlow, \textit{Right}: Ours)}
  \end{subfigure}
  \caption{\small Zero-shot conditional sampling results of our method and StyleFlow \citep{abdal2020styleflow} on FFHQ, where none of the above two attribute combinations is seen in the training set. Our method performs well on zero-shot generation while StyleFlow almost fails by either generating low-quality images or missing the conditioning information. }
  \label{zero_shot}
    \vspace{-14pt}
\end{figure}

The goal is to generate novel images conditioned on \textit{unseen combinations} of attributes that are not present in the training data, which is used for evaluating a model's compositionality in controllable generation.
We compare our method against StyleFlow on zero-shot generation  in Figure \ref{zero_shot}, where we condition on (a) \{\texttt{beard=1, smile=0, glasses=1, age=15}\} and (b) \{\texttt{gender=female, smile=0, glasses=1, age=10}\}, respectively.
We can see our method still performs well in zero-shot generation while StyleFlow suffers from a severe deterioration of image quality and diversity, and almost completely fails in the controllability. Quantitatively, the ACC scores of our method are much larger than StyleFlow, as we have 
0.935 vs. 0.679 (\texttt{smile}), and 0.982 vs. 0.694 (\texttt{glasses}) in the setting (a), and 0.906 vs. 0.408 (\texttt{smile}), and 0.939 vs. 0.536 (\texttt{glasses}) in the setting (b).
These results clearly demonstrate the strong compositionality of our method.

\vspace{-4pt}
\paragraph{Compositions of energy functions}

EBMs have shown great potential in concept compositionality with various ways of composing energy functions \citep{NEURIPS2020_49856ed4}.
Our method can inherit their compositionality, while achieving high image quality. 
Inspired by \cite{NEURIPS2020_49856ed4}, we also consider composing energy functions with three logical operators: conjunction (AND), disjunction (OR) and negation (NOT). 
In particular, given two attributes $\{c_1, c_2\}$, we can define the joint energy function of each logical operator as
\begin{align*}
\small
    \begin{split}
        E(z, \{c_1\; \text{\tiny AND}\; c_2\}) & := E(c_1|g(z)) + E(c_2|g(z)) + \nicefrac{1}{2} \|z\|^2_2 \\
        E(z, \{c_1\; \text{\tiny OR}\; c_2\}) & :=
        -\log \left(e^{\beta - E(c_1|g(z))} + e^{-E(c_2|g(z))} \right)
        + \nicefrac{1}{2} \|z\|^2_2 \\
        E(z, \{c_1\; \text{\tiny AND}\; (\text{\tiny NOT}\; c_2) \}) & := E(c_1|g(z)) - \alpha E(c_2|g(z)) + \nicefrac{1}{2} \|z\|^2_2
    \end{split}
\end{align*}
where the AND operator actually boils down to the conditional sampling with multiple attributes, and $\alpha, \beta>0$ 
are tunable hyperparameters to balance the importance of different energy functions. 
For more complex compositionality, we can recursively apply these logical operators.

\begin{figure}[t]
\vspace{-12pt}
  \centering
  \begin{subfigure}[c]{\textwidth}
		\centering
		\includegraphics[width=\linewidth]{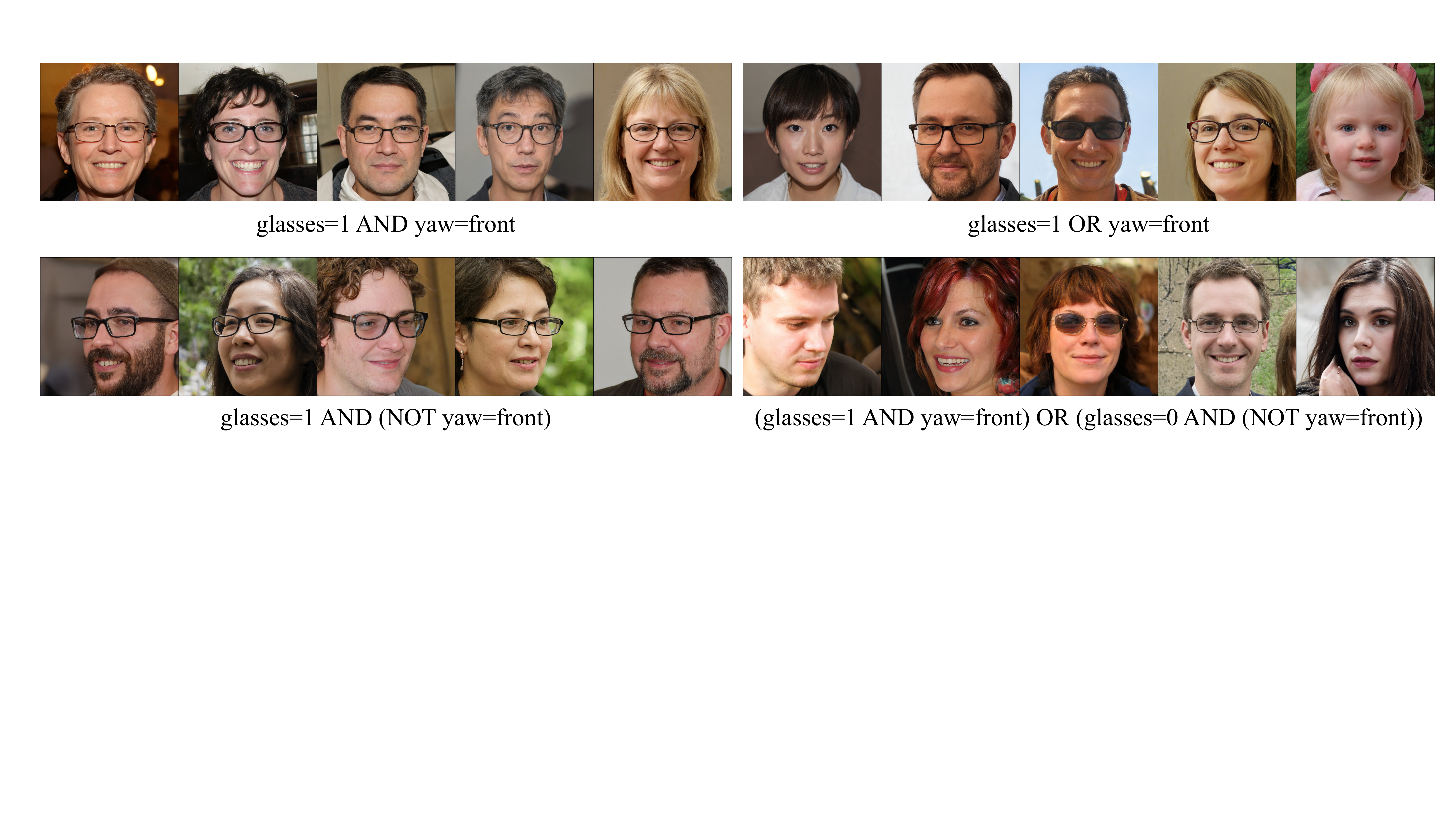}
  \end{subfigure}
  \caption{\small Compositions of energy functions in our method with different logical operators: conjunction (AND), disjunction (OR), negation (NOT), and their recursive combinations on FFHQ of resolution 1024$\times$1024. 
  }
  \vspace{-10pt}
  \label{ffhq_comb_visual}
\end{figure}

Figure \ref{ffhq_comb_visual} demonstrates the concept compositionality of our method for \texttt{glasses} and \texttt{yaw} where we set $\beta = \ln 20$, and $\alpha = \min(\frac{0.1}{|E(z, c_2)|}, 1)$. 
We can see the generated images not only precisely follow the rule of the given logical operators, but are also sufficiently diverse to cover all possible logical cases.
For instance, given \{\texttt{glasses=1 OR yaw=front}\}, some images have glasses regardless of the yaw while other images satisfy \texttt{yaw=front} regardless of the glasses. Our method also works well for the recursive combinations of logical operators, as shown in the bottom-right of Figure \ref{ffhq_comb_visual}. To the best of our knowledge, our method is the first to show such strong compositionality when controllably generating photo-realistic images of resolution $1024 \times 1024$.

\vspace{-6pt}
\subsection{Ablation study on the ODE and LD sampler}
\label{sec:ablation}
\vspace{-4pt}

\begin{wrapfigure}{r}{0.37\textwidth}
  \centering
  \vspace{-25pt}
    \includegraphics[trim=2cm 0cm 2cm 2cm,clip=true,width=0.37\textwidth]{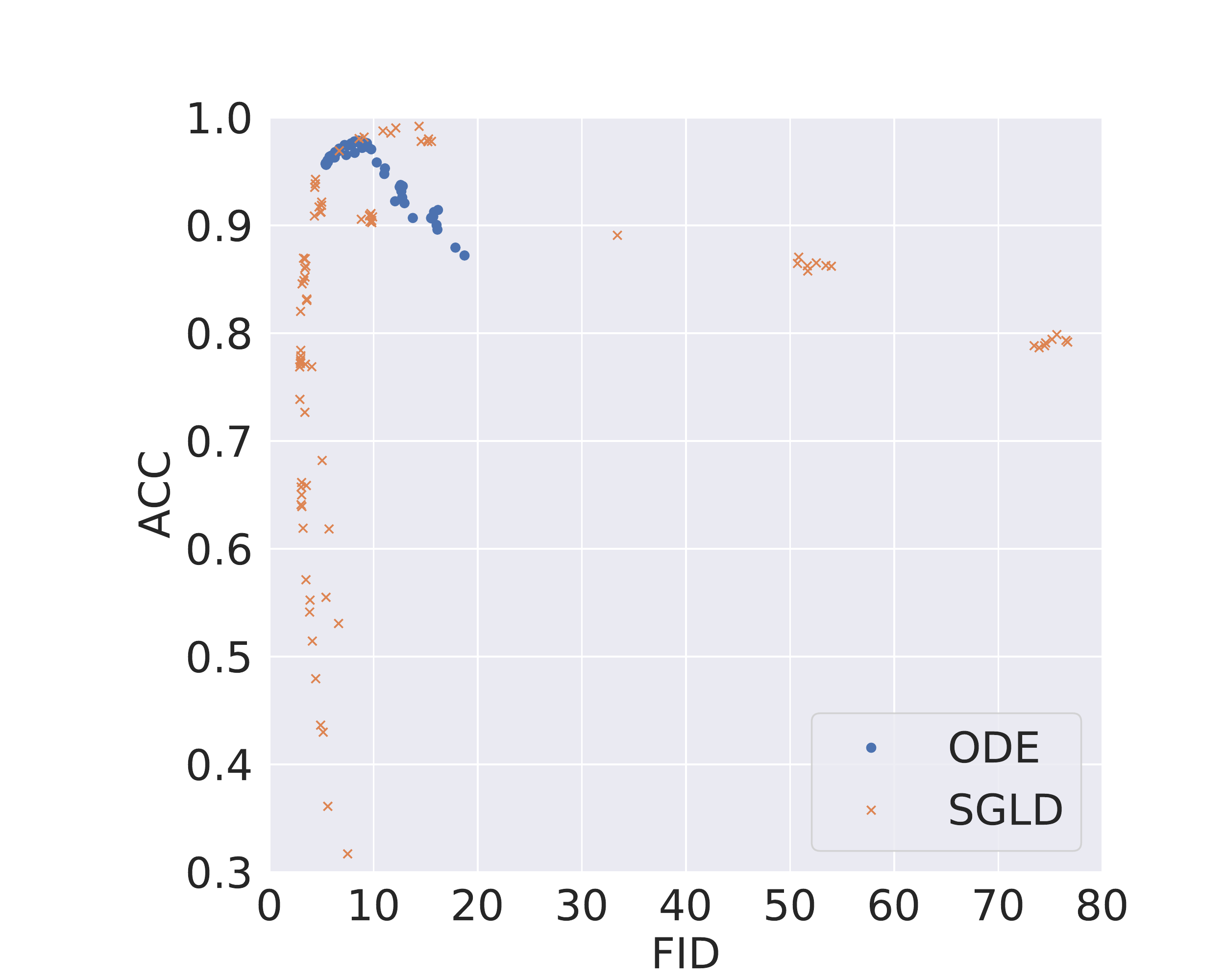}
  \caption{\small The ACC-FID distributions for the ODE and LD sampler, where each dot denotes 
    each hyperparameter setting.}
    \label{which_sampler}
    \vspace{-10pt}
\end{wrapfigure}
Here, we carefully examine the ODE and LD samplers
with a grid search in a large range of hyperparameter settings (see Appendix \ref{app:ode_ld} for details). 
Figure \ref{which_sampler} shows the ACC and FID trade-off of the two samplers on CIFAR-10, where each dot in the figure refers to the result of a particular hyperparameter setting. 
After grid research, 
there are 81 and 104 hyperparameter settings for the ODE and LD sampler, respectively.

We can see that (i) the best ACC-FID scores (on the top-left of Figure \ref{which_sampler}) of the two samplers heavily overlap, implying that they perform equally well in their best tuned hyperparameter settings.
(ii) 
With the concentrated ACC-FID scores for the ODE sampler on the top-left, we can see that the ODE sampler is more stable and less sensitive to the choice of hyperparameters than the LD sampler. (iii) When focusing on the top-left with $\text{ACC} \geq 95\%$ and $\text{FID} \leq 10$, 
the ODE sampler needs much smaller (less than $21\%$) number of minimum average steps than the LD sampler 
(41.8 vs. 200). 
It shows that given the similar performance, the ODE sampler tends to be more efficient due to its adaptivity in step sizes.

\vspace{-6pt}
\subsection{Results on other high-resolution images}
\vspace{-5pt}

\begin{figure}[t]
  \centering
  \begin{subfigure}[c]{0.49\textwidth}
		\centering
		\includegraphics[width=\linewidth]{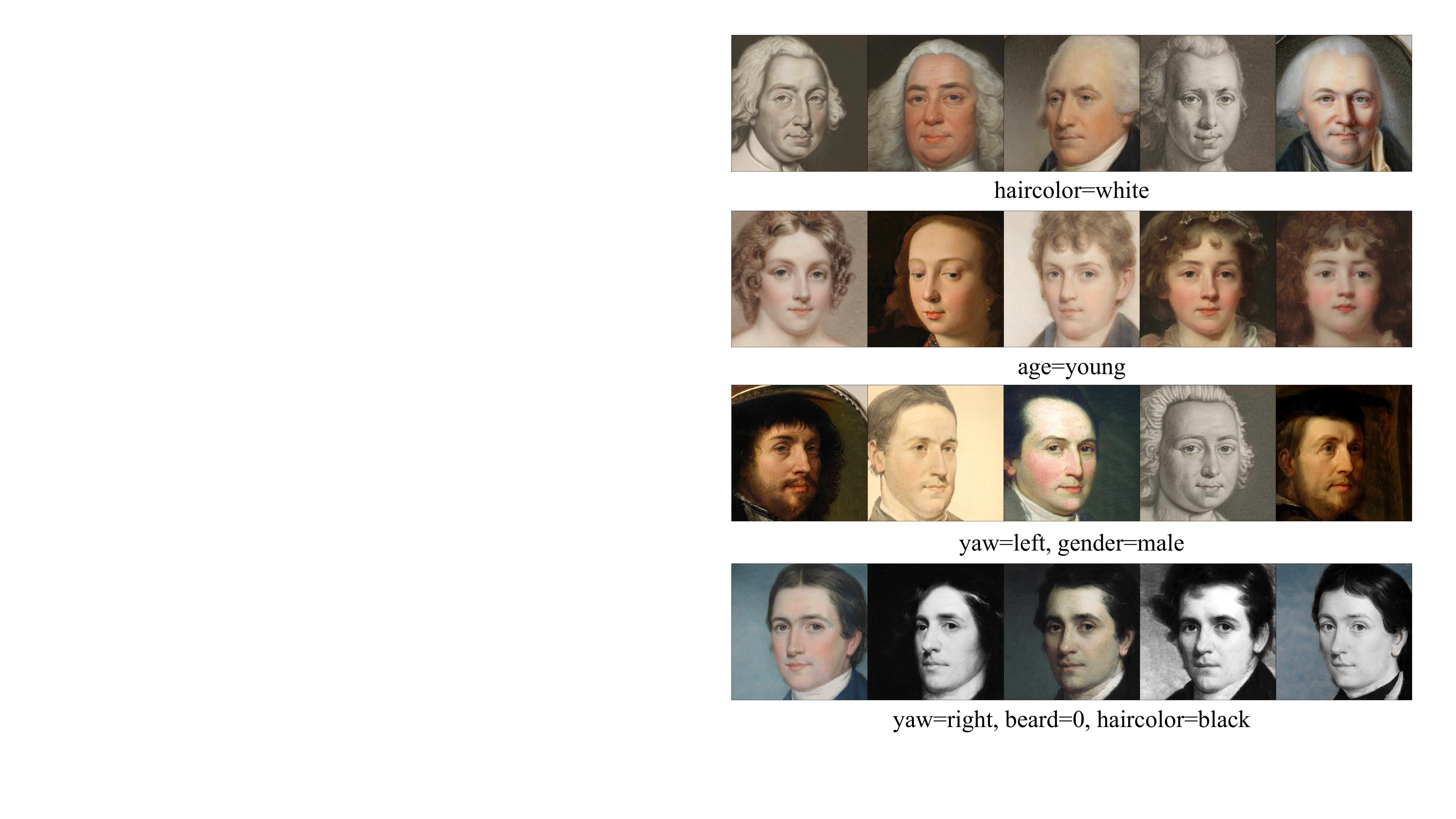}
		\caption{\small MetFaces of resolution 1024$\times$1024}
  \end{subfigure}
  \begin{subfigure}[c]{0.49\textwidth}
		\centering
		\includegraphics[width=\linewidth]{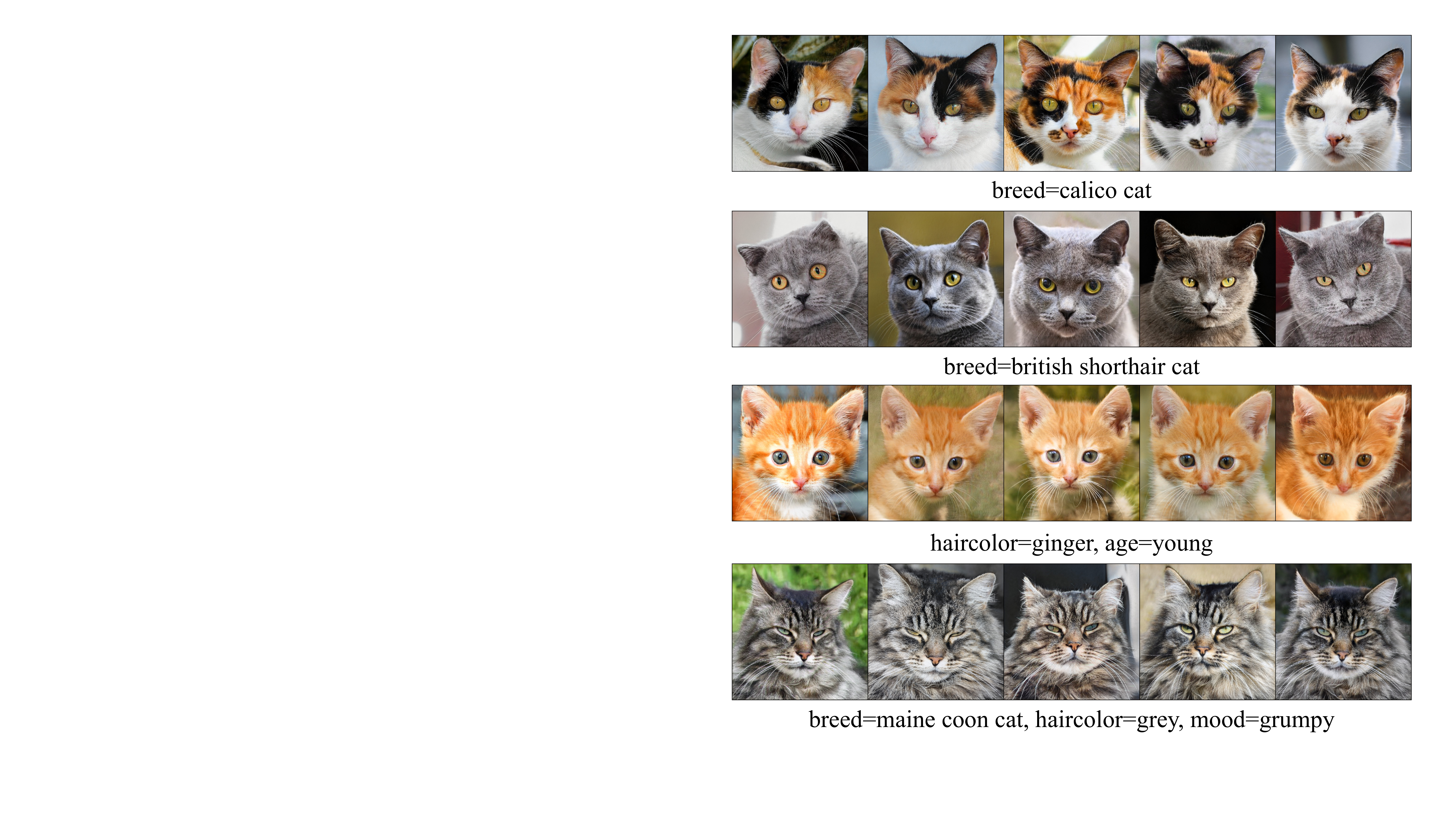}
		\caption{\small AFHQ-Cats of resolution 512$\times$512}
  \end{subfigure}
  \caption{\small Conditional sampling of our method (LACE-ODE) on (a) MetFaces of resolution 1024$\times$1024 and (b) AFHQ-Cats of resolution 512$\times$512, respectively. The generated images are conditioned on either a single attribute or a combination of multiple attributes.
  }
  \vspace{-12pt}
  \label{metfaces_afhq}
\end{figure}

\paragraph{MetFaces \cite{NEURIPS2020_8d30aa96}}
By applying the image classifier pre-trained on the FFHQ data to label the painting faces data (i.e., MetFaces), we train a latent classifier in the $w$-space of StyleGAN-ADA on MetFaces. Accordingly, our method can effectively control the generation of painting faces with one or multiple attributes, as shown in Figure \ref{metfaces_afhq}(a). It demonstrates the robustness of our method to the classification noise and its generalization ability regarding the domain gap between photos and paintings.

\vspace{-7pt}
\paragraph{AFHQ-Cats \cite{choi2020stargan}}

Since the original AFHQ-Cats dataset does not contain the ground-truth attributes (i.e., breed, haircolor, and age, etc.), we have to rely on extra annotators for efficient labeling. Inspired by \cite{gabbay2021image}, we apply the CLIP \cite{radford2021learning} to annotate the AFHQ-Cats data by designing proper prompts that contain the controlling attributes. Similarly, we then train a latent classifier in the $w$-space of StyleGAN-ADA on AFHQ-Cats to apply our method for controllable generation. As shown in Figure \ref{metfaces_afhq}(b), we can effectively control the generation of cat images based on the CLIP annotations.

\vspace{-10pt}
\section{Related work}
\vspace{-7pt}

\paragraph{Conditional generative models} Significant efforts for conditional sampling and image editing have been invested in conditional generative models. In these models, various conditional information can be used to guide the generation process, such as class labels \citep{mirza2014conditional,kingma2013auto}, attribute codes \citep{chen2016infogan,nie2020semi}, source images \citep{choi2018stargan}, text descriptions \citep{reed2016generative}, and semantic maps \citep{park2019semantic}. Training these models is costly, in particular for generating high-resolution images. Additionally, adding or modifying the conditional information requires re-training the whole network. 
On the contrary, our method is based on pre-trained unconditional models, and it requires little effort to introduce new conditional information. 

\vspace{-7pt}
\paragraph{Latent code manipulation for image editing} An alternative approach to conditional GANs is to manipulate the latent code of pre-trained unconditional models, in particular GANs. Some works explore linear manipulations of latent code \citep{jahanian2019steerability,harkonen2020ganspace,shen2021closedform}, while others consider more complicated nonlinear manipulations \citep{nguyen2016synthesizing,goetschalckx2019ganalyze,abdal2020styleflow,patashnik2021styleclip}. Similar to our work, these methods can discover the directions that often correspond to meaningful semantic edits, and need not train conditional generative models from scratch. However, they tend to have issue with compositionality.
For instance, StyleFlow \citep{abdal2020styleflow} specifies a fixed length of attributes during training, limiting its ability to generalize to new attributes and novel combinations (Section~\ref{sec:comp_gen}).
Notably, \cite{nguyen2016synthesizing,goetschalckx2019ganalyze} share the similar idea of optimizing latent variables of generators with a classifier. 
But with the intuition of using generative models as a powerful prior, their classifiers are defined in the pixel space only, making their sampling more challenging. Instead, our method is formulated in a principled way from the EBM perspective (that unifies these methods) and can be defined in various spaces (pixel / $w$ / $z$) of a generator.

\vspace{-8pt}
\paragraph{EBMs and score-based models}
EBMs~\citep{lecun2006tutorial} and score-based models~\citep{song2019generative} have been widely used for image generation~\citep{xie2016theory, NEURIPS2020_49856ed4, song2020improved, song2021scorebased}, due to their flexibility in probabilistic generative modeling. 
Recently, many works have applied them for controllable and compositional generation. \cite{grathwohl2019your} proposes the joint EBM and shows its ability in class-conditional generation. \cite{NEURIPS2020_49856ed4} explores the compositionality of conditional EBMs in controllable generation. \cite{song2021scorebased} proposes a unified framework of score-based models from the SDE perspective, and performs controllable generation by using both a conditional reverse-time SDE and its corresponding ODE. 
Unlike our method that works in the latent space, they all formulate energy functions or score functions in the pixel space, making them both more challenging to train and orders of magnitude slower to sample for high-resolution images.

Also, many other works have built EBMs or score-based models in the latent space of generative models, most of which, however, focus on improving generation quality with EBMs as a structural prior~\citep{ aneja2021ncpvae, arbel2020generalized, han2017alternating, VahdatMBKA18, vahdat2021score, xiao2020vaebm, vahdat2018dvae}. More similar to our work, \cite{xie2021cooperative,pang2020learning} considers using the conditional EBMs as a latent prior (conditioning on labels or attributes) of a generator for controllable generation. A key difference is that they require inferring latent variables from data or sampling from the EBMs using LD during training while we do not need to. This advantage distinguishes our method from these prior methods regarding training speed and inference efficiency.

\vspace{-10pt}
\section{Discussions and limitations}
\label{sec:discussions}
\vspace{-7pt}

We proposed a novel formulation of a joint EBM in the latent space of pre-trained generative models for controllable generation. 
Based on our formulation, all we need for controllability is to train an attribute classifier, and sampling is done in the latent space. Moreover, 
we proposed a more stable and adaptive sampling method by formulating it as solving an ODE. Our experimental results showed that our method is fast to train and efficient to sample, and it outperforms state-of-the-art techniques in both conditional sampling and sequential editing. 
With our strong performance in the zero-shot generation with unseen attribute combinations and compositions of energy functions with logical operators, our method also demonstrated compositionality in generating high-quality images. 

One limitation of this work is that the generation quality is limited by the generative power of the underlying pre-trained generator. 
Since our EBM can be applied to any latent-variable model including GANs and VAEs, it is interesting to apply our method to different generative models based on downstream tasks.
Another limitation is that training the attribute classifier requires 
the availability of attribute labels. 
We believe that advanced semi-supervised and self-supervised techniques \citep{sohn2020fixmatch,chen2020simple} can improve the classifier training while reducing the dependency on labels.


\vspace{-10pt}
\section{Broader impact}
\label{broad_impact}
\vspace{-8pt}

The method presented in this work enables high-quality controllable image synthesis, built upon pre-trained generative models (e.g., StyleGAN2~\citep{Karras2020stylegan2}). Technically, it inherits the compositionality of EBMs and overcomes their difficulty in generating high-resolution images. Since our method does not train any conditional generator from scratch, it significantly reduces the computational cost of training generators for new conditioning attributes. 
 
On the application side, it shares with other image synthesis tools similar potential benefits and risks, which have been discussed extensively in~\citep{Bailey2020thetools,vaccari2020deepfakes}. Our method does not produce new images but only guides the generation process of existing generators. Thus, it inherits potential biases from the pre-trained generators, e.g. StyleGAN2 trained on the FFHQ dataset~\citep{karras2019style}. On the other hand, by providing a semantic control of image generation with strong compositionality, our method can be used to discover and proactively reduce unknown biases in existing generators.

{\small

\bibliography{references}
\bibliographystyle{apalike}
}

\newpage

\appendix

\section{Appendix}

\vspace{-6pt}
\subsection{Derivation of the joint distribution in latent space}
\label{app:derivation}
Our proof closely follows \citep{che2020GanIsEBM} from the rejection sampling perspective. Before the derivation, we introduce the following lemma about the resulting probability distribution in rejection sampling:

\begin{lemma} \label{app:lem}
 (Lemma 1 in \citep{che2020GanIsEBM}) Given a probability distribution $p(x)$ where $x \in \mathcal{X}$ and a measurable function $r: \mathcal{X} \to [0, 1]$, the rejection sampling with the proposal distribution $p(x)$ and acceptance probability $r(x)$ generates $x$ samples following from the distribution $q(x)$, which satisfies $q(x) = p(x) r(x) / Z_0$ where $Z_0 = \mathbb{E}_{x \sim p(x)}[r(x)]$.
\end{lemma}

\emph{Proof: } See the proof of Lemma 1 in \citep{che2020GanIsEBM}. \hfill $\square$

Recall that we define the generative model as
\begin{align} \label{app:generative_model}
    p_\theta(x, c) = p_g(x) e^{-E_\theta(c|x)} / Z
\end{align}
where $p_g(x)$ is an implicit distribution defined by a pre-trained generator $g$ in the form of $x = g(z)$ with the latent variable $z \in \mathcal{Z}$ and the data $x \in \mathcal{X}$, and $Z$ is a normalization constant.

When conditioned on the attribute $c$, we have $p_\theta(x | c) = \frac{p_\theta(x, c)}{p_\theta(c)}$. Similar to \citep{che2020GanIsEBM}, if we do rejection sampling with the proposal distribution $p_g(x)$ and the acceptance probability $\frac{p_\theta(x| c)}{M(c) p_g(x)}$, where $M(c)$ is a constant (w.r.t. $x$) satisfying $M(c) \geq \frac{p_\theta(x|c)}{p_g(x)}$, we get samples $x$ from  $p_\theta(x|c)$. 
As $p_g(x)$ is an implicit distribution induced by the generator $x = g(z)$, this rejection sampling on $p_g(x)$ is equivalent to a rejection sampling on the prior $p(z)$ in the latent space and then applying $x = g(z)$.

Specifically, the corresponding rejection sampling in the latent space has the proposal distribution $p(z)$ and the acceptance probability 
\begin{align} \label{app:acc_prob}
    \small
    r_\theta(z, c) = \frac{p_\theta(g(z)| c)}{M(c) p_g(g(z))} = \frac{e^{-E_\theta(c|g(z))}}{M(c) p_\theta(c) Z}
\end{align}
where the second equation follows from Eq. (\ref{app:generative_model}). From Lemma \ref{app:lem}, we have that conditioned on the attribute $c \sim p_\theta(c)$, the resulting probability $p_\theta(z| c)$ of the accepted $z$ samples in the above rejection sampling procedure satisfies
\begin{align} \label{app:pzc}
    p_\theta(z| c) = p(z) r_\theta(z, c) / Z'(c)
\end{align}
where $Z'(c) := \mathbb{E}_{z \sim p(z)}[r_\theta(z, c)]$. By substituting Eq. (\ref{app:acc_prob}) into Eq. (\ref{app:pzc}), we have 
\begin{align} \label{app:pzc_v2}
    \small
    p_\theta(z| c) = \frac{ p(z) e^{-E_\theta(c|g(z))} } { p_\theta(c) Z''(c) }
\end{align}
where $Z''(c) := M(c) ZZ'(c)$, and from Eq. (\ref{app:acc_prob}) we further have
\begin{align}
\small
    \begin{split}
         Z''(c) & = \mathbb{E}_{z \sim p(z)}[e^{-E_\theta(c|g(z))}] / p_\theta(c) \\
        & = \mathbb{E}_{x \sim p_g(x)}[e^{-E_\theta(c|x)}] / p_\theta(c) \\
        & = \frac{Z}{p_\theta(c)} \int p_\theta(x, c) dx \\
        & = Z
    \end{split}
\end{align}
where the second equation follows from the change of variables with $x=g(z)$, and the third equation follows from marginalizing $x$ in Eq. (\ref{app:generative_model}).
Thus, Eq. (\ref{app:pzc_v2}) yields
\begin{align}
    p_\theta(z, c) =  p(z) e^{-E_\theta(c|g(z))} / Z
\end{align}
which concludes the derivation. 
 \hfill $\square$

 \subsection{Derivation of Eq. (\ref{ode_final})}
\label{app:derivation_ode}

Song et al.~\cite{song2021scorebased} define a forward diffusion process that maps samples $x_0 \sim p_{\text{data}}$ to $x_T \sim p_T = N(0, I)$ using the VP-SDE:
\begin{align}
dx = -\frac{1}{2}\beta(t) x dt + \sqrt{\beta(t)} dw    
\end{align}
for $t \in [0, T]$. They also show that a generative SDE can be defined by:
\begin{align}
dx = -\frac{1}{2}\beta(t)[x + 2\nabla_x \log p_t(x)]dt + \sqrt{\beta(t)} d\bar{w}
\end{align}
with time flowing backward from $T$ to 0 and reverse standard Wiener process $\bar{w}$. For conditional generation, the generative SDE above becomes:
\begin{align} \label{eq:reverse_sde2}
dx = -\frac{1}{2}\beta(t)[x + 2\nabla_x \log p_t(x, c)]dt + \sqrt{\beta(t)} d\bar{w}
\end{align}
where $p_t(x, c)$ is the join distribution of data and attribute at time $t$.

Song et al.~\cite{song2021scorebased} show that there exists an equivalent ODE whose trajectories share the same joint probability densities $p_t(x(t), c)$ with the reverse SDE defined in Eq. (\ref{eq:reverse_sde2}):
\begin{align}
    dx = -\frac{1}{2} \beta(t) \left[ x + \nabla_x \log p_t(x, c) \right] dt
\end{align}
where the idea is to use the Fokker-Planck equation~\citep{oksendal2003stochastic} to transform an SDE to an ODE (see Appendix D.1 for more details in~\citep{song2021scorebased}).

Song et al.~\cite{song2021scorebased} define $\nabla_x \log p_t(x(t), c) := \nabla_x \log p_t(x(t)) + \nabla_x \log p_t(c|x(t))$ which requires (i) estimating the score function $\nabla_x \log p_t(x(t))$, and (ii) training a \emph{time-variant} classifier $p_t(c|x(t))$ in the $x$-space for $\nabla_x \log p_t(c|x(t))$, both of which make the training and inference challenging. 
Instead, in our framework, we solve the ODE in the $z$-space and transfer the $z$ samples to the data space with the generator $g$. 
Specifically with $p_t(z(t), c) \propto e^{-E_t ( c | g(z(t)) ) + \log p_t(z(t))}$, we have 
\vspace{-3pt}
\begin{align}
    \begin{split}
        dz &
        = -\frac{1}{2} \beta(t) \left[ z + \nabla_z \log p_t(z) - \nabla_z E_t( c| g(z) ) \right] dt
    \end{split}
\end{align}
Since the distribution of latent variable $z$ in most generative models satisfies $p_0(z(0)) = \mathcal{N}(0, I)$, diffusing it with VP-SDE will not change its distribution at time $t$, i.e., $p_t(z(t)) = \mathcal{N}(0, I)$~\cite{song2021scorebased}. 
Since $p_t(z(t))$ is time-invariant and the generator $g$ is fixed, the classifier $p_t(c|g(z(t)))$ receives input $g(z(t))$ with a time-invariant distribution. Thus, 
we assume that the classifier is also time-invariant, and so is the energy function $E_t(c| g(z(t))) := E(c| g(z(t)))$. 
Therefore, the above ODE becomes: 
\vspace{-3pt}
\begin{align}
    \begin{split}
        dz &= \frac{1}{2} \beta(t) \nabla_z E( c| g(z) ) dt = \frac{1}{2} \beta(t) \sum_{i=1}^n \nabla_z E( c_i| g(z) ) dt
    \end{split}
\end{align}
which concludes the derivation. 
 \hfill $\square$


\vspace{-4pt}
\subsection{More details of experimental setting}
\label{app:exp_setting}

\vspace{-3pt}
\paragraph{Training hyperparameters} 
For the classifier $f_i(x; \theta)$, we use a four-layer MLP as the network architecture when it is trained in the $z$-space or the $w$-space of StyleGAN2, as shown in Table \ref{tab:clf_arc}. We train the classifier with the Adam optimizer \citep{kingma2014adam} for 100 epochs using a staircase decay schedule.  
When we consider the classifier in the pixel space, we directly use the pre-trained WideResNet-28-10 \citep{zagoruyko2016wide} with no batch normalization as the network architecture.

\begin{table}[ht]
\small
\centering
\vspace{-6pt}
\caption{\small The four-layer MLP architecture as the attribute classifier when it is trained in the $z$-space or the $w$-space of StyleGAN2. Its output dimension depends on the number of class predictions for the targeted attributes.
}
\vspace{2pt}
\begin{tabular}{l}
\hline
 input: $z \in \mathbb{R}^{512}$ or $w \in \mathbb{R}^{512}$
 \\
Linear 384, LeakyReLU                                          \\
Linear 256, LeakyReLU                                        \\
Linear 128, LeakyReLU
    \\ 
Linear \#logits
    \\
    \hline
\end{tabular}
\label{tab:clf_arc}
\vspace{-8pt}
\end{table}

\vspace{-5pt}
\paragraph{Inference hyperparameters}
By default, we use the `dopri5' ODE solver with the absolute and relative tolerances (\texttt{atol}, \texttt{rtol}) being set to (1e-3, 1e-3) in LACE-ODE. The time-variant diffusion coefficient $\beta(t)$ in the ODE sampler has the form that $\beta(t) = \beta_{\min} + (\beta_{\max} - \beta_{\min}) t$, where $\beta_{\min} = 0.1$ and $\beta_{\max} = 20$, and $t \in [0, 1]$. 
Similar to the prior work~\citep{grathwohl2019your}, for the LD sampling in LACE-LD,  the step size $\eta$ and the standard deviation of $\epsilon_t$ are chosen separately for faster training, although it results in a biased sampler \citep{grathwohl2019your,du2019implicit}. By default, we set the number of steps to be 100, step size $\eta=0.01$ and standard deviation of $\epsilon_t$ to be 0.01 in LACE-LD.

We use StyleGAN-ADA as the pre-trained generator for experiments on CIFAR-10
and StyleGAN2 for experiments on FFHQ. StyleGAN-ADA shares the same network architecture with StyleGAN2, where the truncation trick can be applied in the $w$-space for better image quality \citep{karras2019style}. Since our EBM formulation does not modify the generator architecture, we can also apply the truncation in the $w$-space during sampling. By default, we use the truncation coefficient $\psi=0.7$ in our method. 

\vspace{-5pt}
\paragraph{Hardware}
We ran all experiments on one single NVIDIA V100 GPU with 32GB memory size.

\vspace{-5pt}
\paragraph{Data preparation}
To train attribute classifier in the $z$-space or the $w$-space of StyleGAN2, we first need a training set of the pairs $(z, c)$ or $(w, c)$, where $c$ represents the class label in CIFAR-10 and the attribute code in FFHQ. 
For the experiments on CIFAR-10, we first sample 50k $z$ latent variables from the standard Gaussian and pass them to the StyleGAN2 generator to get 50k $w$ samples (i.e., the output of the mapping network) and 50k images (i.e., the output of the synthesis network). 
Then, we label the 50k images using 
a pre-trained DenseNet \citep{huang2017densely} image classifier with an error rate 4.54\% on CIFAR-10. Accordingly, we get 50k $(z,c)$ pairs and 50k $(w,c)$ pairs as the training sets.

For the experiments on FFHQ, we directly use the 10k $(w, c)$ pairs created by the StyleFlow paper \citep{abdal2020styleflow} to train the attribute classifier. 
We use 12 attributes in most of our experiments, including 5 discrete attributes: \texttt{smile}, \texttt{glasses}, \texttt{gender}, \texttt{beard}, \texttt{haircolor}, and 7 continuous attributes: \texttt{yaw}, \texttt{age}, \texttt{pitch}, \texttt{bald}, \texttt{width}, \texttt{light0}, \texttt{light3}. Usually for binary discrete attributes, we denote ``1'' as the presence and ``0'' as the absence. For instance, \texttt{glasses}=1 means wearing glasses and \texttt{glasses}=0 means no glasses. Similarly, \texttt{smile}=1 means smiling and \texttt{smile}=0 means no smiling. For continuous attributes, we normalize their values to the range of $[0, 1]$.

\paragraph{Metrics}
To quantify the model performance in controllable generation, we mainly use the following two metrics: (i) \textit{conditional accuracy} (\textit{ACC}) to measure the controllability, and (ii) \textit{FID} to measure the generation quality and diversity \citep{heusel2017gans}.
Specifically, the ACC score is calculated as follows: We first generate $N_a$ images based on randomly sampled attributes codes, and then pass these generated images to a pre-trained image classifier to predict the attribute codes. Accordingly, the ACC score reflects how accurately the predicted attribute codes match the sampled ground-truth ones. 

For the experiments on CIFAR-10 and FFHQ, these two metrics are evaluated in slightly different ways. First, for the ACC score on CIFAR-10, we use the aforementioned DenseNet pre-trained on CIFAR-10 
as the image classifier. For each class, we uniformly sample 1k images, meaning the total generated images $N_a=10\text{k}$. The final ACC score is then the averaged accuracy of the predicted class labels over 10 classes. For the FID score on CIFAR-10, we uniformly sample 5k images in each class and then use the total 50k images to calculate the FID. 

Second, for the ACC score on FFHQ, we use the MobileNet \citep{howard2017mobilenets} as the network backbone of the image classifier due to its small size and effectiveness in the recognition of face attributes. To improve the generalization ability of the image classifier, we first train the MobileNet with the Adam optimizer for 10 epochs on the CelebA dataset \citep{liu2015faceattributes}, and then fine-tune it on the 10k generated FFHQ image and attribute pairs for another 50 epochs. 
Similarly, we \textit{uniformly} sample the attribute codes from the set of all possible combinations, and generate 1k images to compute the ACC. The final ACC score is then the averaged accuracy of the predicted attribute codes over all the sampled attribute combinations. Note that for the continuous attributes $c_i \in \mathbb{R}$, such as \texttt{yaw} and \texttt{age}, we normalize their values to the range of $[0,1]$, and the ACC for each continuous attribute is represented by $1- | \hat{c}_i - c_i |$ instead, where $\hat{c}_i$ is the predicted continuous attribute from the MobileNet image classifier. 

For the FID score on FFHQ, we follow from StyleFlow \citep{abdal2020styleflow} that uses 1k generated samples StyleGAN2 to compute the FID. 
Note that the resulting FID scores are not comparable to those reported in the original StyleGAN2 paper, as it uses 50k real FFHQ images to evaluate the FID. 
For the reference, the original unconditional StyleGAN2 with our evaluation protocol has FID=20.87$\pm$0.11.
Besides, different from CIFAR-10 where each class has equally distributed samples, the attribute distribution in the FFHQ data is heavily imbalanced. For instance, the number of images with \texttt{glasses}=0 is at least 5$\times$ larger than that with \texttt{glasses}=1. The number of images with \texttt{smile}=1 is at least 3$\times$ larger than that with \texttt{smile}=0.
Thus, if we uniformly sample attributes as before, the resulting generated image distribution will largely deviate from the reference data distribution, making the FID score incorrectly reflect the generation quality. 
To remedy this, we randomly sample attribute codes \textit{from the training set} instead to generate 1k images for the FID evaluation on FFHQ.

\vspace{-3pt}
\subsection{More details of baselines}
\label{app:baselines}

We use different baselines for comparing with our method in controllable generation. The first set of baselines is the EBMs in the pixel space:

\vspace{-5pt}
\paragraph{JEM \citep{grathwohl2019your}}
It proposes the joint EBM framework of modelling the data and labels in the pixel space. 
Its training is composed of two parts: $p_\theta(c|x)$ for the classifier training and $p_\theta(x)$ for the generative modelling. In both training and inference of JEM, the LD sampling is applied to draw samples from $p_\theta(x)$. We use the default hyperparameter settings in \citep{grathwohl2019your} to report its results. 

\vspace{-5pt}
\paragraph{Cond-EBM \citep{NEURIPS2020_49856ed4}}
Based on conditional EBMs, it proposes different ways of composing the energy functions with logical operators for compositional generation.
To train conditional EBMs, it also applies the LD sampling to draw samples from $p_\theta(x|c)$. We use the default hyperparameter settings in \citep{NEURIPS2020_49856ed4} to report its results. Particularly during inference, to improve the generation quality, we apply the following tricks \citep{NEURIPS2020_49856ed4}: (i) we combine two training checkpoints, and (ii) we run 50 LD steps followed by the data augmentations to get a good initialization of the LD sampling. 

The second set of baselines is the score-based models with SDEs~\citep{song2021scorebased}:
\paragraph{VP-SDE \citep{song2021scorebased}}
In \textit{Variance Preserving (VP) SDE} \citep{song2021scorebased}, the forward SDE is defined as 
\begin{align}
    dx = -\frac{1}{2}\beta(t) x dt + \sqrt{\beta(t)} dw 
\end{align}
where $\beta(t)$ represents a scalar time-variant diffusion coefficient and $w$ is a standard Wiener process. Then, the
conditional sampling from $p_0(x|c)$ is equivalent to solving the following reverse SDE:
\vspace{-3pt}
\begin{align}\label{app_eq:reverse_sde}
    dx = -\frac{1}{2}\beta(t)[x + \nabla_x \log p_t(x, c)]dt + \sqrt{\beta(t)} d\bar{w}
\end{align}
where $\bar{w}$ is a standard Wiener process when time flows backwards from $T$ to 0. 
To sample from Eq. (\ref{app_eq:reverse_sde}), the \textit{Predictor-Corrector (PC)} sampler is proposed in \citep{song2021scorebased}. At each time step, the numerical SDE solver first gives an estimate of the sample
at the next time step, playing the role of a “predictor”. Then, the score-based MCMC approach
corrects the marginal distribution of the estimated sample, playing the role of a “corrector” \citep{song2021scorebased}.

\paragraph{VE-SDE \citep{song2021scorebased}}
In \textit{Variance Exploding (VE)} SDE, the forward SDE is defined as 
\begin{align}
    dx = \sqrt{\frac{d [\sigma^2(t)]}{dt}} dw 
\end{align}
where $\sigma(t)$ represents a sequence of positive noise scales and $w$ is a standard Wiener process. Then, the
conditional sampling from $p_0(x|c)$ is equivalent to solving the following reverse SDE:
\begin{align} \label{app_eq:ve_sde}
    dx = - \nabla_x \log p_t(x, c) d [\sigma^2(t)] + \sqrt{\frac{d [\sigma^2(t)]}{dt}} d \bar{w} 
\end{align}
where $\bar{w}$ is a standard Wiener process when time flows backwards from $T$ to 0. 
According to \citep{song2021scorebased}, the PC sampler can also be applied to sample from Eq. (\ref{app_eq:ve_sde}).

To report the controllable generation results of VP-SDE and VE-SDE, we use the pre-trained models released by the official implementation (\url{https://github.com/yang-song/score_sde}), and also used the default sampling hyperparameters of the PC sampler: the number of predictor steps $N=1000$, the number of corrector steps $M=1$, and the signal-to-noise ratio $r=0.16$.

The last set of baselines is the methods modelled in the latent space of the pre-trained generator:

\vspace{-5pt}
\paragraph{StyleFlow \citep{abdal2020styleflow}}

It applies the conditional continuous normalizing flows (CNFs) to build an invertible mapping between the $z$-space and the $w$-space of StyleGAN2 conditioned on the attribute codes. 
The goal is to enable adaptive latent space vector manipulation by casting the
conditional sampling problem in terms of conditional CNFs using the attributes for conditioning \citep{abdal2020styleflow}. 

The conditional sampling task is straightforward: it sets the attribute code to a desired set of values, and then samples multiple $z$ variables, which are passed to the conditional CNF and the synthesis network of StyleGAN2 to get the final images. The sequential editing task is mainly composed by a sequence of Conditional Forward Editing (CFE) and Joint Reverse Encoding (JRE). Meanwhile, several hand-crafted tricks are applied to improve the editing quality, including the Edit Specific Subset Selection and re-projection of edited $w$ to the $z$-space. See the original paper \citep{abdal2020styleflow} for details. 

To get the reported results, we use the pre-trained models released by the official implementations (\url{https://github.com/RameenAbdal/StyleFlow}). As we keep all the hand-crafted tricks mentioned above, it implies that we actually use the StyleFlow (V2) \citep{abdal2020styleflow} for comparison. 
Besides that, we use the default sampling hyperparameters. In particular, we use the adjoint method to compute the gradients and solve the ODE using `dopri5' ODE solver, where the tolerances are set to 1e-5.

\vspace{-5pt}
\paragraph{Latent-JEM}
This is a baseline we propose by modelling JEM \citep{grathwohl2019your} in the latent space of a pre-trained generator. 
Similarly, the Latent-JEM is modelled in the $w$-space of StyleGAN2. Given the joint distribution of $w$ varaible and attribute code $c$:
\begin{align}\label{app_eq:joint}
    p_\theta(w, c) \propto e^{-E_\theta(w, c)},
\end{align}
then we assume 
$E_\theta(w, c) = \sum_{i=1}^n E_\theta(w, c_i)$ (i.e., the conditional independence assumption) where 
\begin{align}
    E_\theta(w, c_i) = \begin{cases}
    -f_i(x, \theta)[c_i] & \text{if $c_i$ is discrete} \\ 
    \frac{1}{2\sigma^2} (c_i - f_i(x, \theta))^2 & \text{if $c_i$ is continuous} \\ 
    \end{cases}
\end{align}
Similarly, $f_i(x;\theta)$ is the output of a multi-class classifier mapping from $\mathcal{X}$ to $\mathbb{R}^{m_i}$ if the $i$-th attribute is discrete or a regression network mapping from $\mathcal{X}$ to $\mathbb{R}$ if it is continuous.
Note that the original JEM paper \citep{grathwohl2019your} has only considered the discrete case, so here we propose a more generalized framework that also works for the continuous attributes. 

By marginalizing out $c$ in Eq. (\ref{app_eq:joint}), we obtain an unnormalized density model:
\begin{align}\label{app_eq:marg}
    p_\theta(w) \propto e^{-E_\theta(w)},
\end{align}
where the marginal energy function is given by
\begin{align}\label{app_eq:marg_energy}
    E_\theta(w) = -\sum_{i \in \mathcal{I}_{\text{dis}}}\log \sum\nolimits_{c_i} \exp(f_i(x, \theta)[c_i])
\end{align}
where $\mathcal{I}_{\text{dis}}$ is the index set of all discrete attributes. 
Similar to JEM \citep{grathwohl2019your}, when we compute the conditional $p_\theta(c|w)$ via $p_\theta(w, c) / p_\theta(w)$ by dividing Eq. (\ref{app_eq:joint}) to Eq. (\ref{app_eq:marg}), the normalizing constant cancels out, yielding the standard Softmax parameterization for the discrete attributes and the squared L2 norm parameterization for the continuous attributes. 

During training, we follow from \citep{grathwohl2019your} to optimize $p_\theta(c|w)$ using standard cross-entropy and optimize $p_\theta(w)$ using Eq. (\ref{LD}) with the LD where gradients are taken with respect to the marginal energy function (\ref{app_eq:marg_energy}).
In practice, we find a trade-off between the generation quality and controllability in Latent-JEM controlled by the step size $\eta$. Thus, after a grid search, we use both two step sizes: $\eta=0.1$ and $\eta=0.01$ to get the reported results, while the number of LD steps $N=200$ and the standard deviation of noise $\sigma=0.01$ work the best for Latent-JEM. 

Besides, we use the reply buffer of size 10,000 during training and inference, as suggested by \citep{grathwohl2019your}, to improve the results of Latent-JEM on CIFAR-10. 
For the experiments of Latent-JEM on FFHQ, instead of sampling $w$ from an uniform distribution as the initialization point of the LD \citep{grathwohl2019your}, we get a better initialization of $w$ by first randomly sampling $z$ from the standard Gaussian and passing $z$ to the pre-trained mapping network of StyleGAN2. By doing so, the performance of Latent-JEM on FFHQ improves significantly.

\vspace{-5pt}
\paragraph{LACE-PC}
This is another baseline we propose by replacing the ODE sampler with the Predictor-Corrector (PC) sampler from the SDE perspective \citep{song2021scorebased}. We keep the EBM formulation in Eq. (\ref{energy_final}) unchanged.
In experiments, we first perform a grid search on the hyperparameters of the PC sampler: the number of predictor steps $N$, the number of corrector steps $M$ and the signal-to-noise ratio $r$. Similarly, we find a trade-off between generation quality and controllability in LACE-PC, controlled by the the number of predictor steps. Thus, we use both two numbers of predictor steps: $N=100$ and $N=200$ to get the reported results while the number of corrector steps $M=1$ and the signal-to-noise ratio $r=0.05$ work the best for LACE-PC.

\vspace{-5pt}
\subsection{Which space to train the classifier?}
\label{app:which_space}

\begin{figure}[ht]
\vspace{-8pt}
  \begin{minipage}[b]{\textwidth}
    \centering
    \captionof{table}{\small The \textit{FID} and ACC scores of the ODE and LD sampler in different spaces of StyleGAN-ADA on CIFAR-10, where "default" means
      the default hyperparameter setting for each sampler,
      "best\_acc" and "best fid" denote the hyperparameter settings with the best ACC and the best FID, respectively, in grid research.}
\footnotesize\addtolength{\tabcolsep}{-3pt}
  \begin{tabular}{cc|cccccc}
    \hline
    \multirow{2}{*}{Sampler} &
    \multirow{2}{*}{Space} &
      \multicolumn{2}{c}{default} &
      \multicolumn{2}{c}{best\_acc} &
      \multicolumn{2}{c}{best\_fid} \\
      \cline{3-8}
      & & ACC$\uparrow$ & FID$\downarrow$ & ACC$\uparrow$ & FID$\downarrow$ & ACC$\uparrow$ & FID$\downarrow$ \\
      \hline
    \multirow{3}{*}{ODE} &
    $z$ & 0.929 & 7.34 & 0.933 & \bf{7.94} & 0.912 & 6.66 \\
    & $w$ & \bf{0.971} & \bf{6.69} & \bf{0.979} & 8.52 & \bf{0.957} & \bf{5.40} \\
    & $i$ & 0.473 & 20.18 & 0.473 & 20.18 & 0.413 & 9.98 \\
    \hline
    \multirow{3}{*}{LD} &
    $z$ & 0.924 & 10.27 & 0.990 & 23.62 & 0.549 & 2.93 \\
    & $w$ & \bf{0.935} & \bf{4.34} & \bf{0.992} & \bf{14.36} & \bf{0.769} & \bf{2.89} \\
    & $i$ & 0.394 & 10.85 & 0.468 & 74.76 & 0.134 & 3.28 \\
    \hline
  \end{tabular}
      \label{which_space}
  \end{minipage}
  \vspace{-10pt}
\end{figure}


We use StyleGAN-ADA \citep{NEURIPS2020_8d30aa96}  pre-trained on CIFAR-10 \citep{krizhevsky2009learning} to investigate which space works the best to train the classifier. In particular, we compare the performances of our method in three spaces of StyleGAN-ADA: $z$-space, $w$-space and pixel space (or $i$-space). 
The results are shown in Table \ref{which_space} for the ODE and LD sampler, respectively. We can see that in different hyperparameter settings, our method works the best in the $w$-space for both samplers.
The reason why $z$-space works worse is that the classifier in the $z$-space has lower accuracy than that in the more disentangled $w$-space. 
The fact that we get the worst performance in the $i$-space is mainly because of its difficulty in convergence.
Therefore, we focus on the $w$-space to train the classifier for our method.

\vspace{-8pt}
\subsection{More results of conditional sampling on CIFAR-10}
\label{app:cond_cifar}
\vspace{-4pt}

We report the results of our method and baselines on CIFAR-10 with error bars in Table \ref{app:baselines_cifar10}.
Note that in Table \ref{app:baselines_cifar10}, the reported FID is slightly higher than that of the pre-trained StyleGAN-ADA \citep{NEURIPS2020_8d30aa96} (FID: $2.92\pm0.05$).
This is because our goal is to turn an unconditional generative model into a conditional one for better controllable generation, and the controllable sampling process changes the generated data distribution. Specifically, the original StyleGAN-ADA randomly samples the latent $z$ (by following a standard Gaussian) for image generation, while our method controllably samples the latent $z$ (to satisfy the conditional attribute specifications) with the ODE/LD sampler. The resulting data distributions of the two sampling methods will be different, thus making the FID different.

\begin{figure}[ht]
\vspace{-8pt}
    \centering
\begin{subfigure}[c]{\textwidth}
    \centering
        \includegraphics[width=\linewidth]{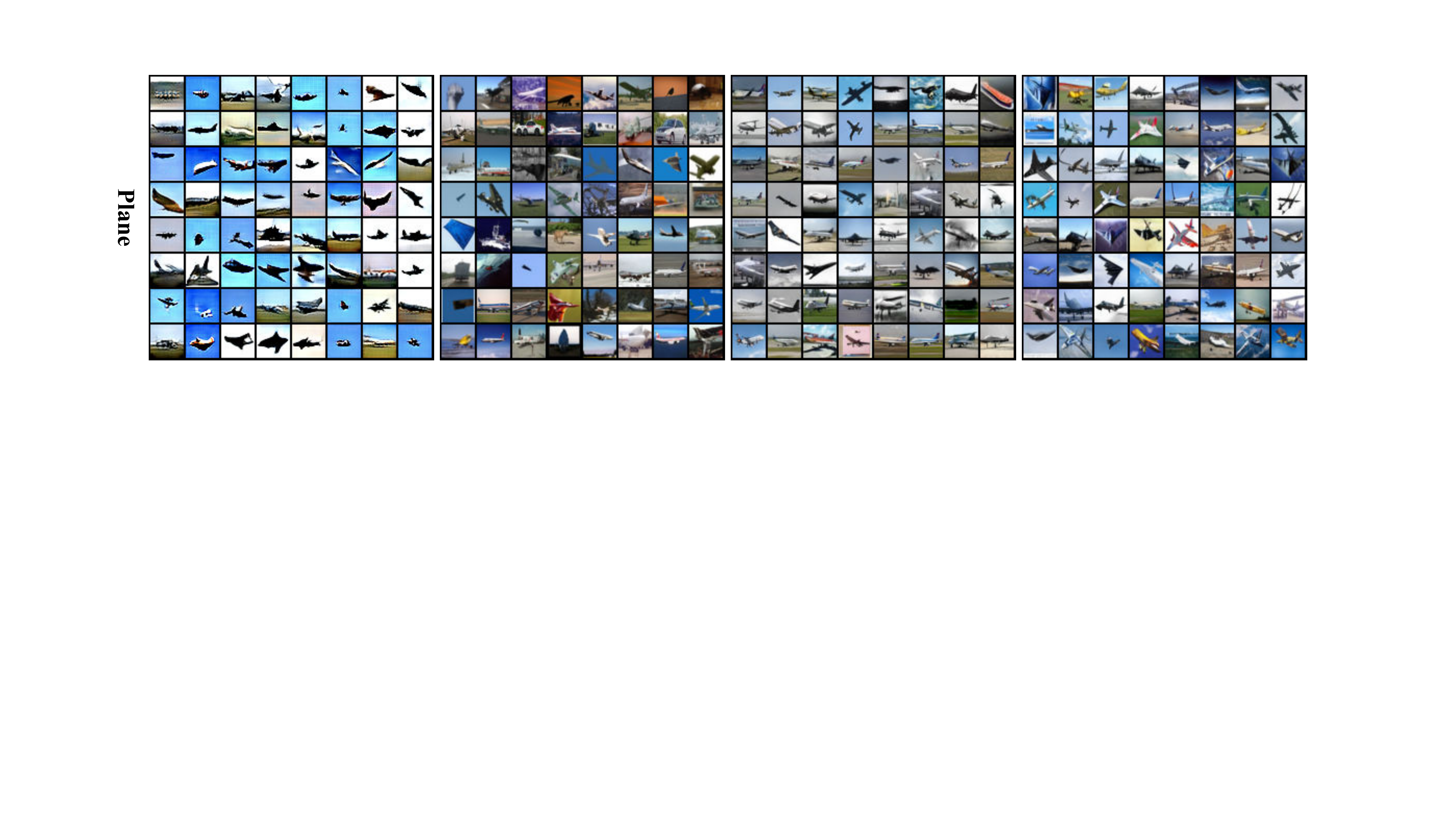}
    \includegraphics[width=\linewidth]{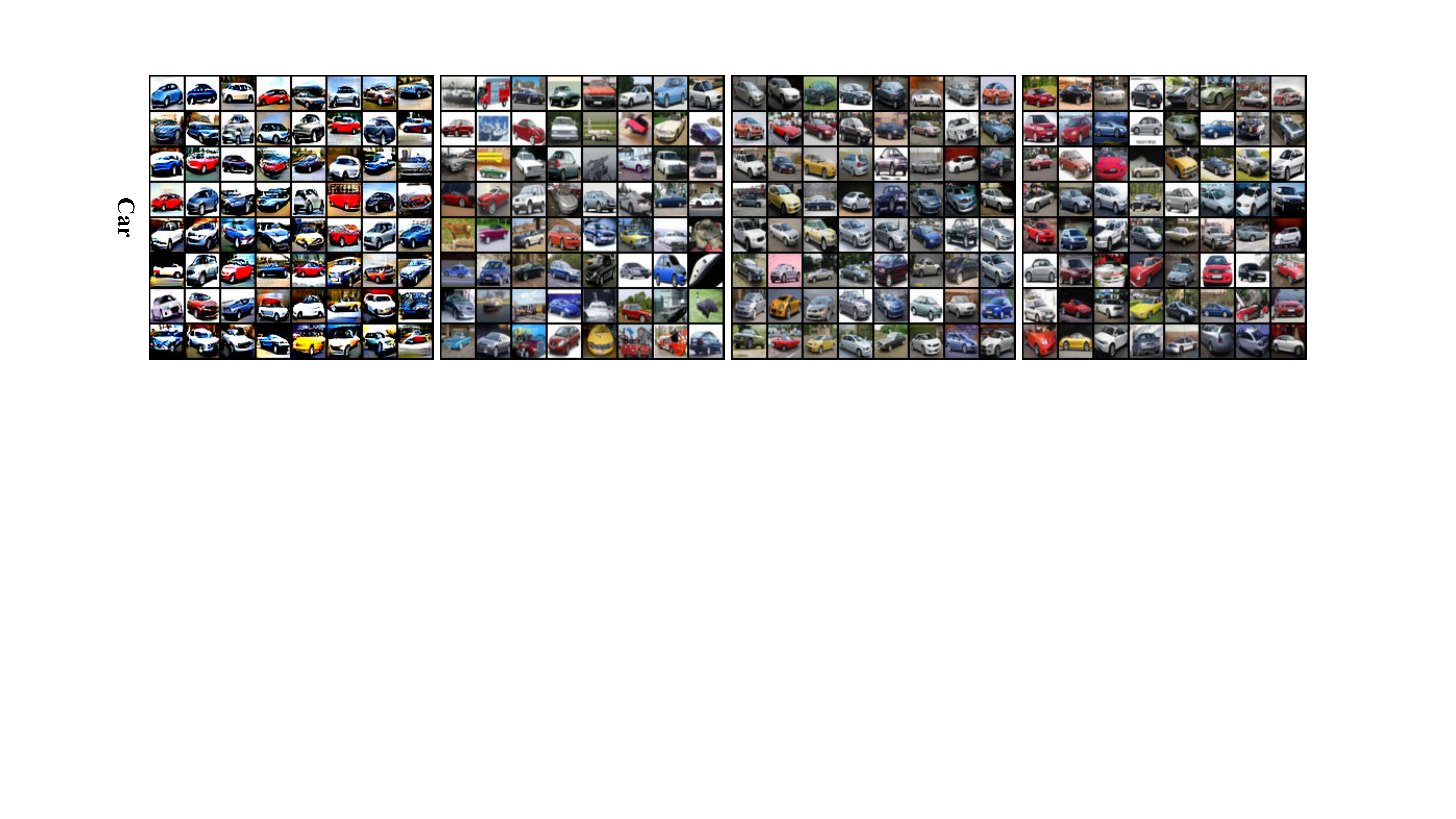}
    \includegraphics[width=\linewidth]{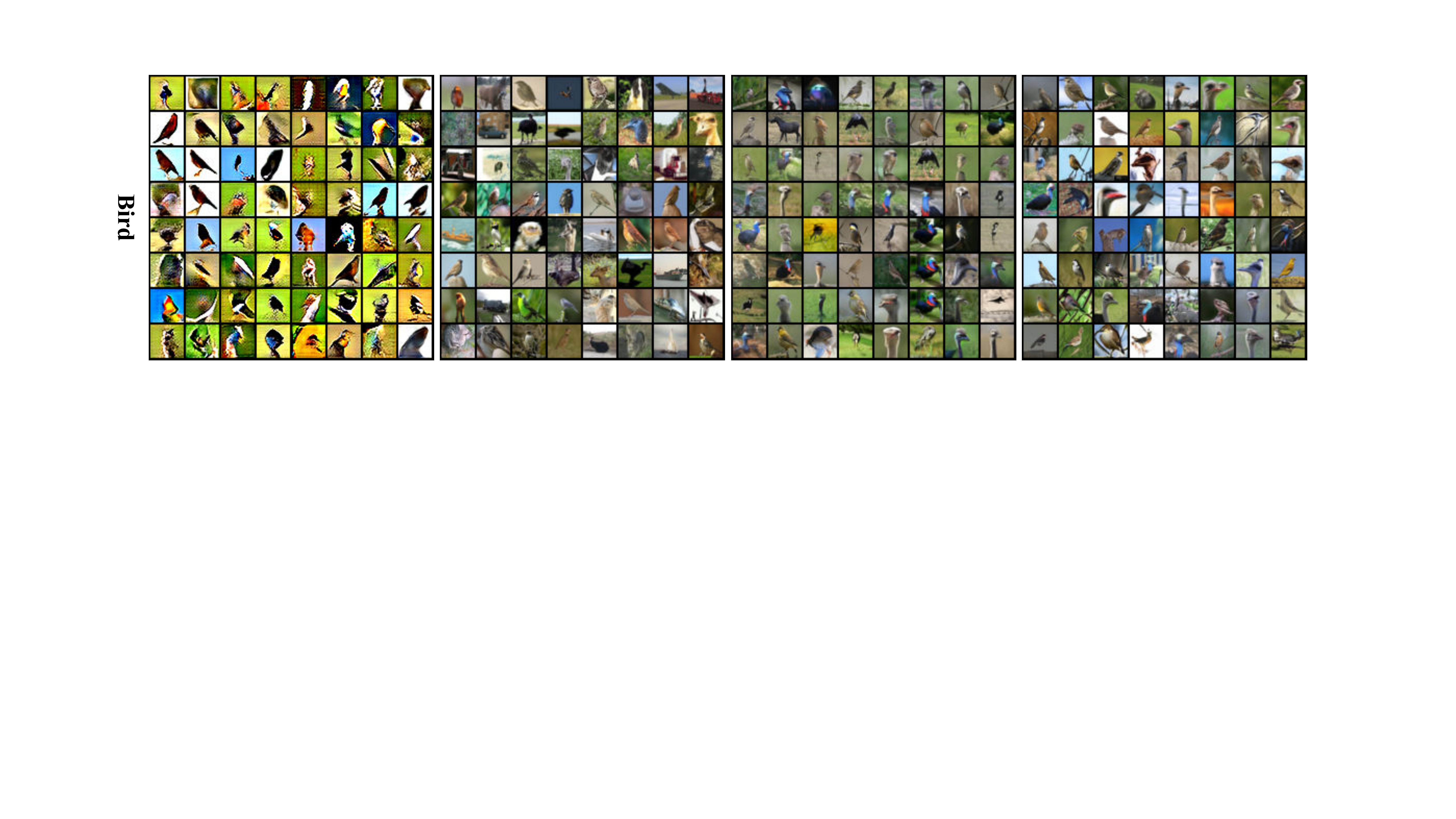}
    \includegraphics[width=\linewidth]{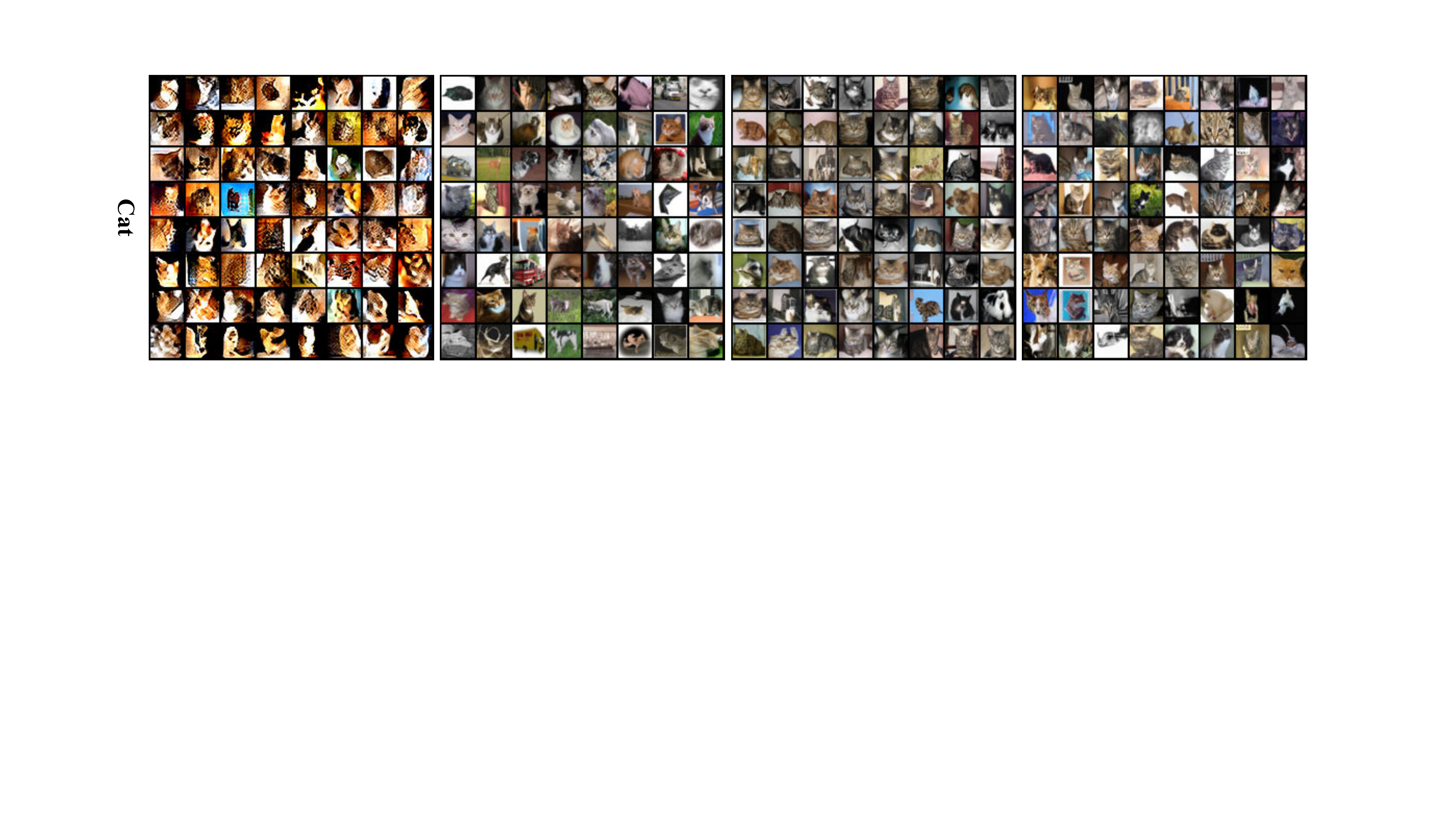}
    \includegraphics[width=\linewidth]{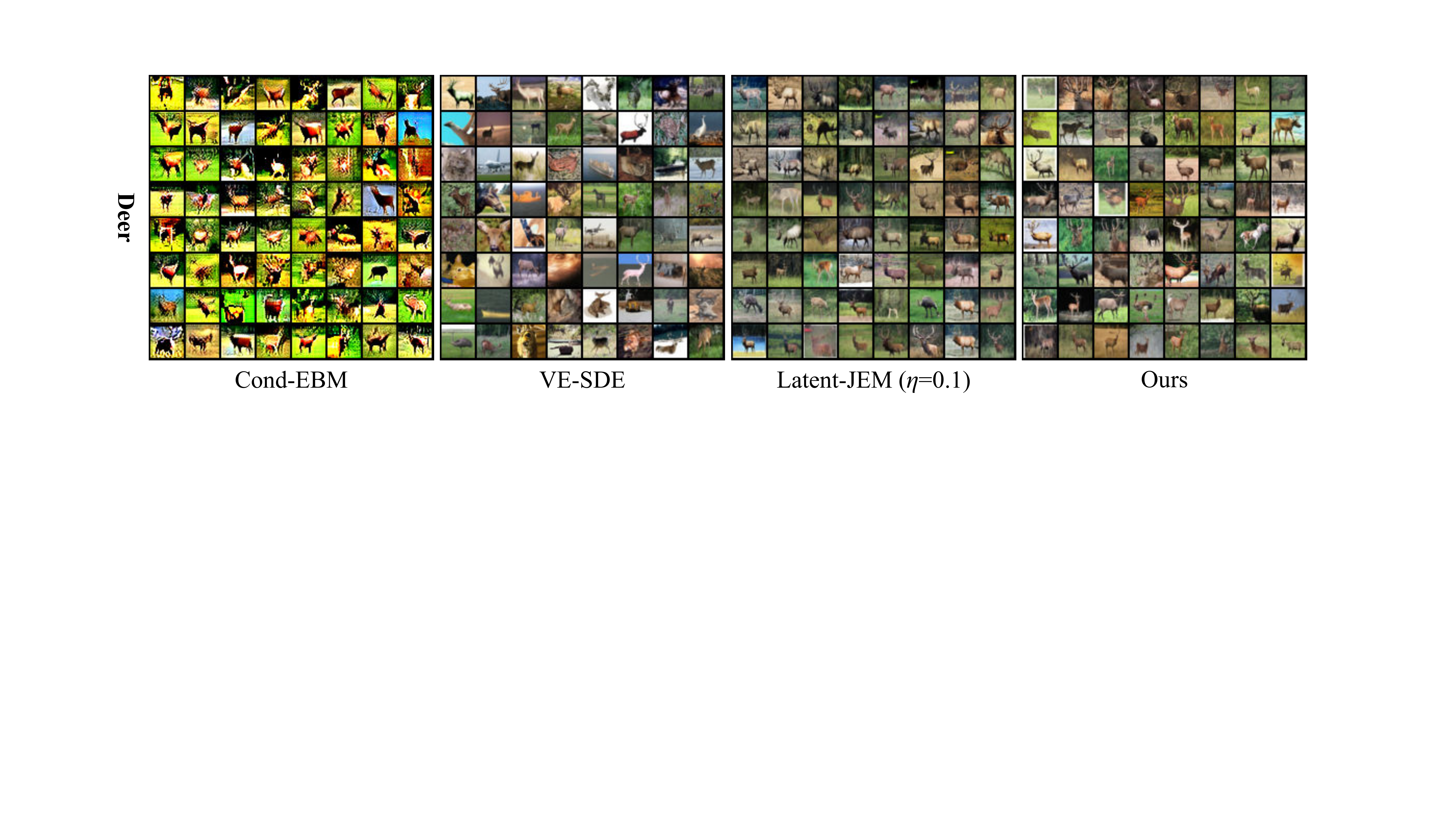}
\end{subfigure}
    
    \caption{\small Conditionally generated images of our method (LACE-ODE) and baselines on each class of CIFAR-10 (0-4): plane, car, bird, cat, and deer. We can see that our method can achieve good controllability with high image quality and diversity. On the contrary, Cond-EBM suffers from the poor image quality and diversity, VE-SDE suffers from the poor controllability (with many samples inconsistent with the given class label), and the proposed baseline Latent-JEM tends to have worse image diversity than ours.
    }
    \label{app:cifar10_visual_a}
    \vspace{-10pt}
\end{figure}

\begin{figure}[ht]
    \centering
\begin{subfigure}[c]{\textwidth}
    \centering
    \includegraphics[width=\linewidth]{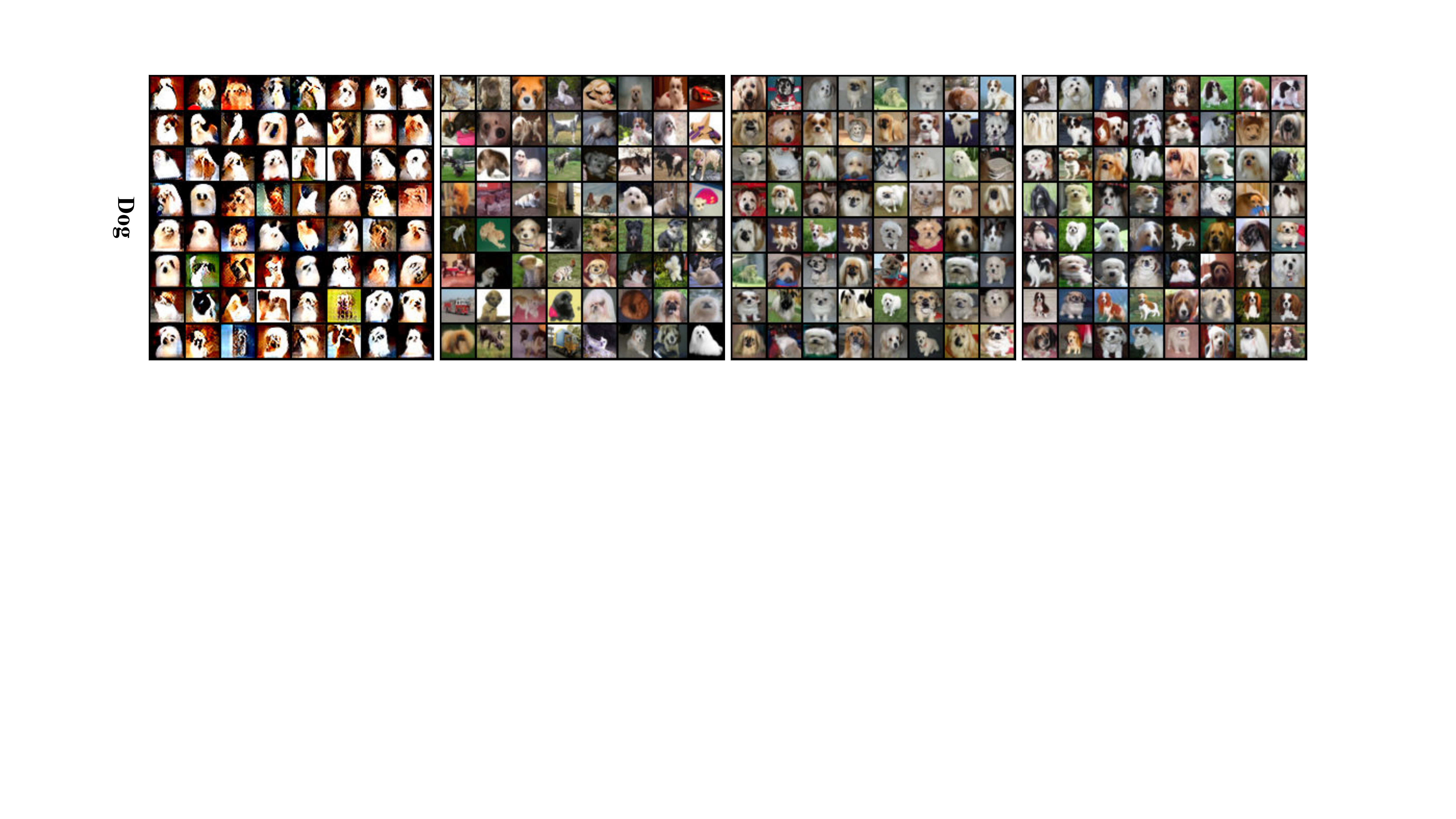}
    \includegraphics[width=\linewidth]{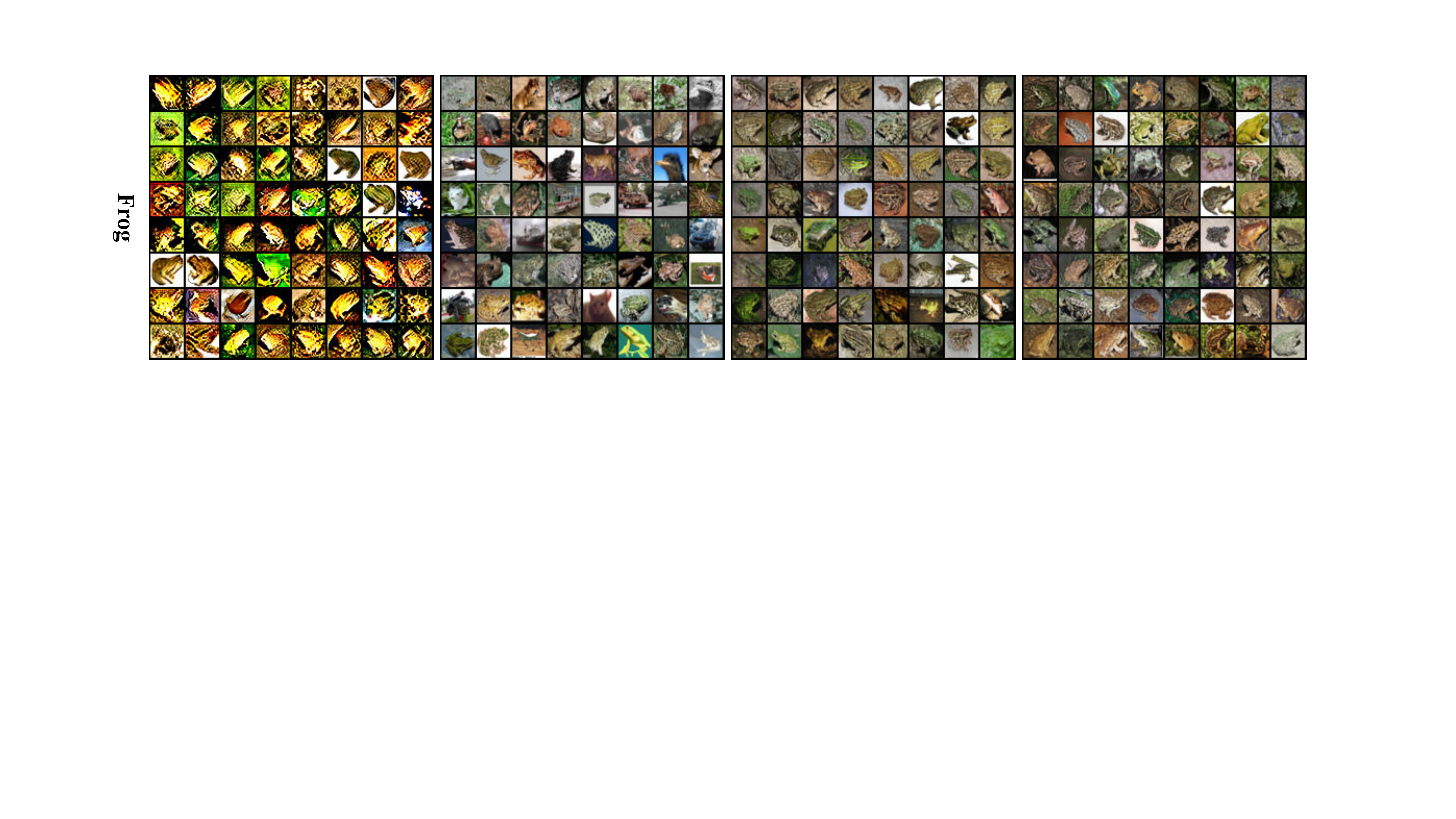}
    \includegraphics[width=\linewidth]{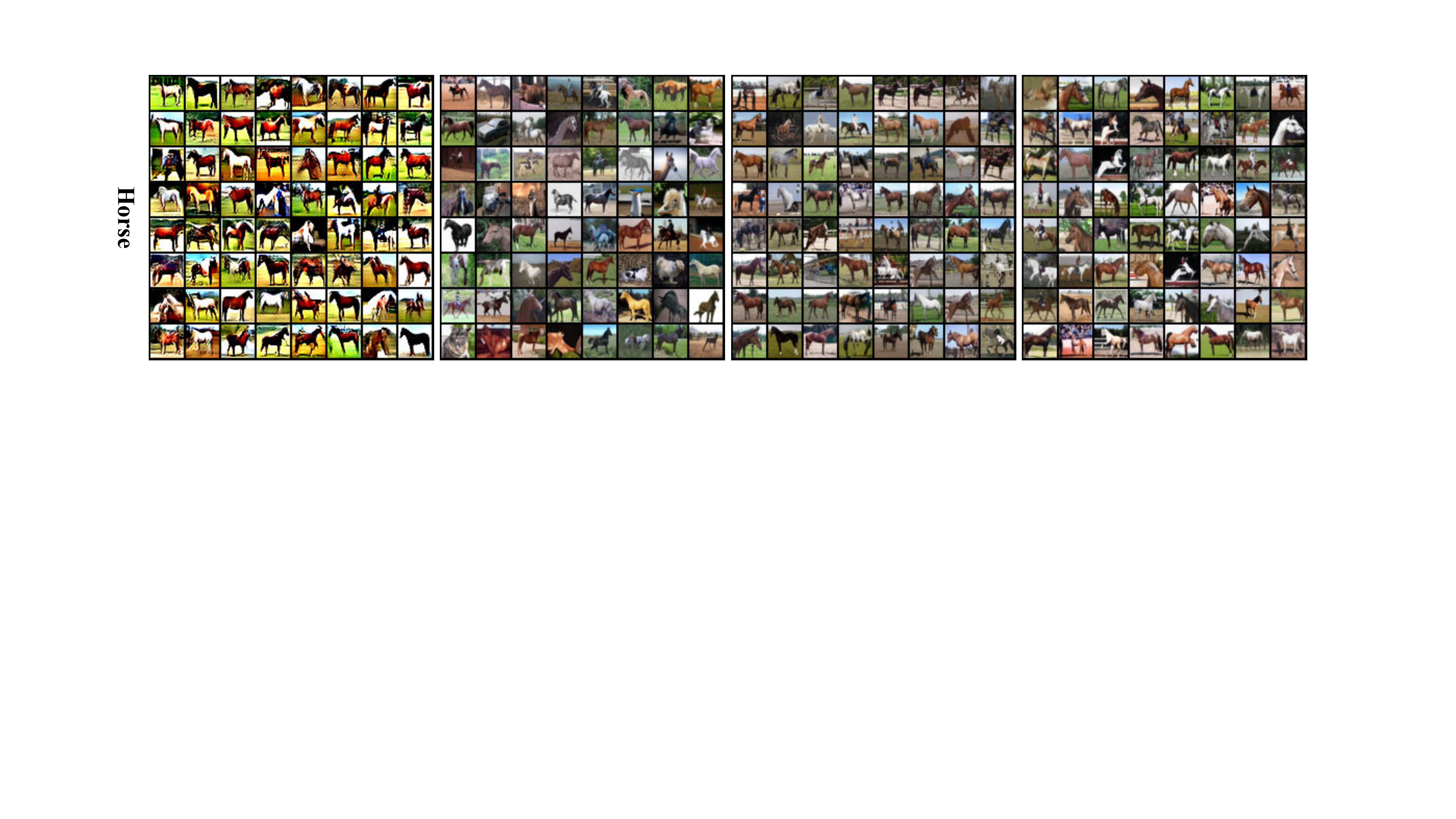}
    \includegraphics[width=\linewidth]{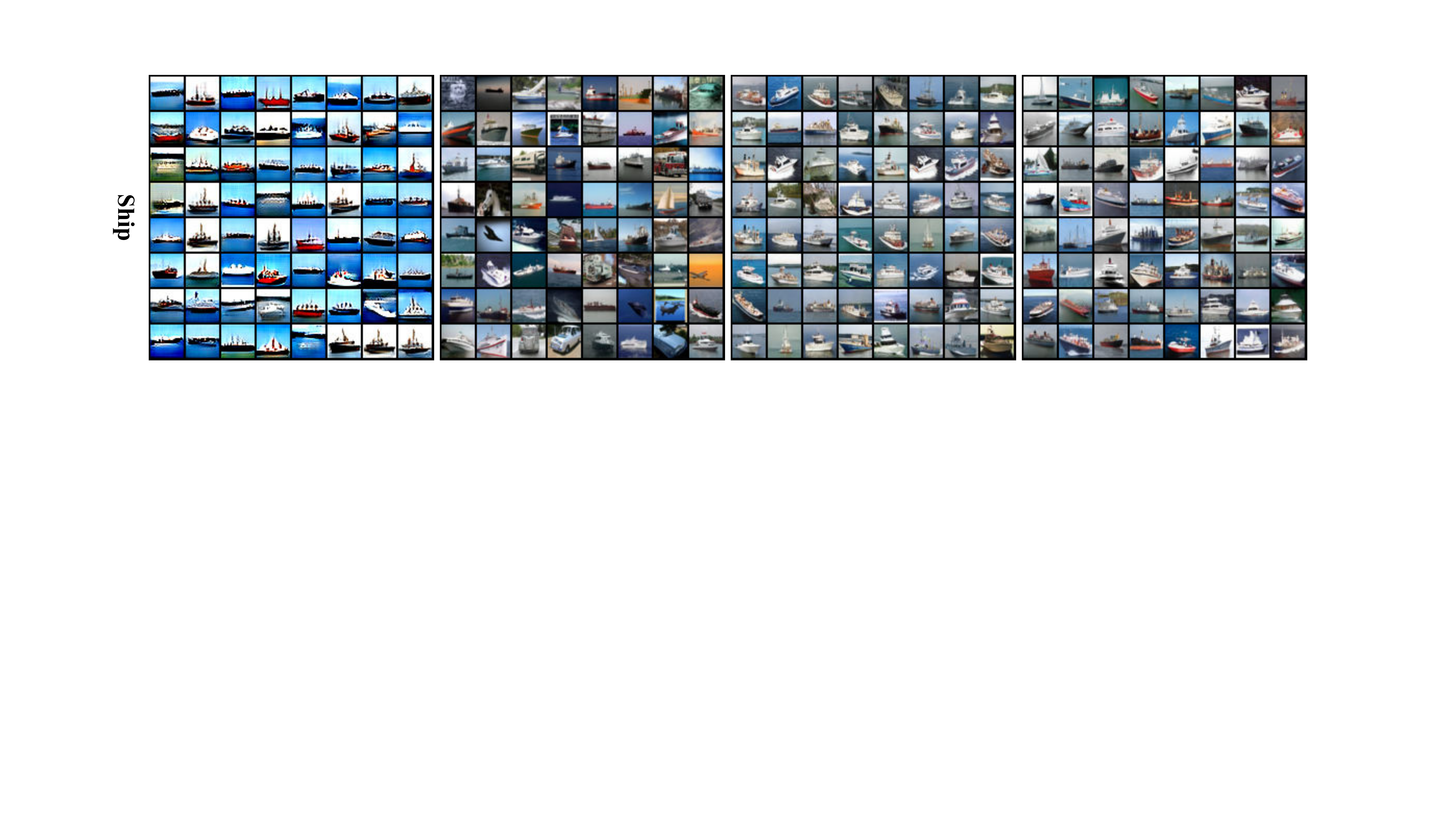}
    \includegraphics[width=\linewidth]{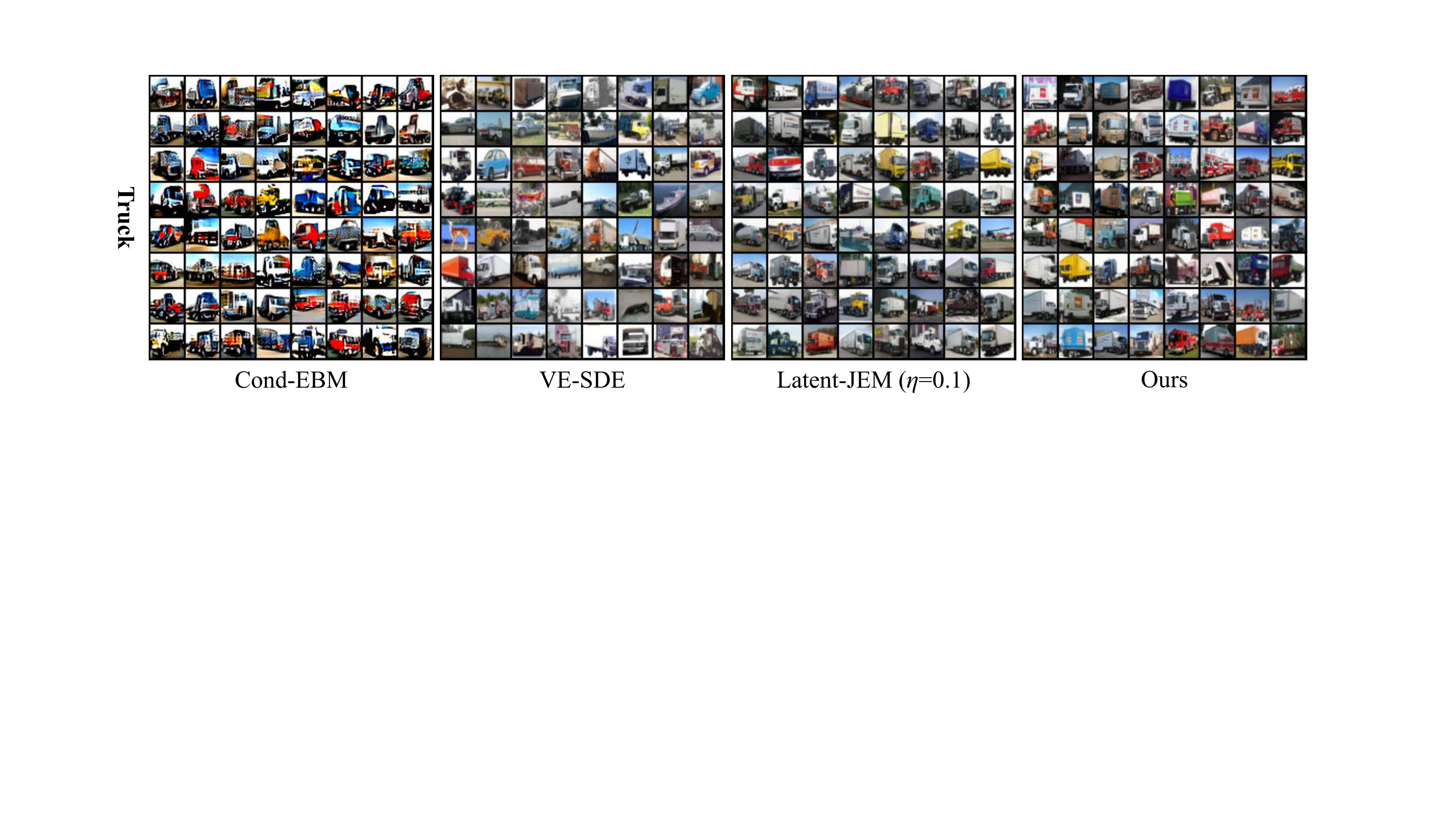}
\end{subfigure}
    
    \caption{\small Conditionally generated images of our method (LACE-ODE) and baselines on each class of CIFAR-10 (5-9): dog, frog, horse, ship, and truck. We can see that our method can achieve good controllability with high image quality and diversity. On the contrary, Cond-EBM suffers from the poor image quality and diversity, VE-SDE suffers from the poor controllability (with many samples inconsistent with the given class label), and the proposed baseline Latent-JEM tends to have worse image diversity than ours.
    }
    \label{app:cifar10_visual_b}
    \vspace{-12pt}
\end{figure}

The visual samples of our method (LACE-ODE) and baselines conditioned on each class of CIFAR-10 can be seen in Figure \ref{app:cifar10_visual_a} and Figure \ref{app:cifar10_visual_b}.

\begin{table}[ht]
\vspace{-10pt}
    \centering
    \captionof{table}{\small Comparison of our method and baselines for conditional sampling on CIFAR-10.
    For notations, Train -- training time, Infer -- inference time (m: minute, s: second), which refer to the single GPU time for generating a batch of 64 images,
    $\eta$ is the LD step size, and $N$ is the number of predictor steps in the PC sampler. 
    }
    \vspace{2pt}
    
\footnotesize\addtolength{\tabcolsep}{-4pt}
  \begin{subtable}[t]{\textwidth}
    \centering
    \begin{tabular}{c|cccc}
        \hline
         Methods & Train & Infer & FID$\downarrow$ & ACC$\uparrow$ \\
         \hline
          JEM \citep{grathwohl2019your} &
          2160m & 135s & 52.35\text{\tiny $\pm$.09} & 0.645\text{\tiny $\pm$.008}
          \\
          Cond-EBM \citep{NEURIPS2020_49856ed4} & 
          2280m & 24.5s & 41.72\text{\tiny $\pm$.01} & 0.792\text{\tiny $\pm$.003} 
          \\
          VP-SDE \citep{song2021scorebased} & 
          52800m & 438s & 19.13\text{\tiny $\pm$.04} & 0.643\text{\tiny $\pm$.003}
          \\
          VE-SDE \citep{song2021scorebased} & 
          52800m & 448s  & \textbf{2.97\text{\tiny $\pm$.04}} & 0.662\text{\tiny $\pm$.002} 
          \\
\hline
        Latent-JEM ($\eta$=0.1) & 
        21m & 0.63s 
        & 8.75\text{\tiny $\pm$.13} & 0.950\text{\tiny $\pm$.003}
        \\
        Latent-JEM ($\eta$=0.01) & 
        21m & 0.63s 
        & 5.65\text{\tiny $\pm$.09} & 0.821\text{\tiny $\pm$.001}
        \\
\hline
\hline
        LACE-PC ($N$=100) &
        \textbf{4m} & 0.84s & \textbf{2.99\text{\tiny $\pm$.01}} & 0.747\text{\tiny $\pm$.001}
        \\
        LACE-PC ($N$=200) &
        \textbf{4m} & 1.86s  & \textbf{2.94\text{\tiny $\pm$.02}} & 0.722\text{\tiny $\pm$.001}
        \\
        LACE-LD &
        \textbf{4m} & 0.68s & 4.30\text{\tiny $\pm$.05} & 0.939\text{\tiny $\pm$.002}
        \\
        LACE-ODE & 
        \textbf{4m} & \textbf{0.50s} & 6.63\text{\tiny $\pm$.06} & \textbf{0.972\text{\tiny $\pm$.001}}
        \\
         \hline
    \end{tabular}
\end{subtable}
      \label{app:baselines_cifar10}
      \vspace{-10pt}
\end{table}

\vspace{-3pt}
\subsection{More results of conditional sampling on FFHQ}
\label{app:cond_ffhq}
\vspace{-5pt}

\begin{figure}[ht]
    \centering
\begin{subfigure}[c]{\textwidth}
    \centering
        \includegraphics[width=\linewidth]{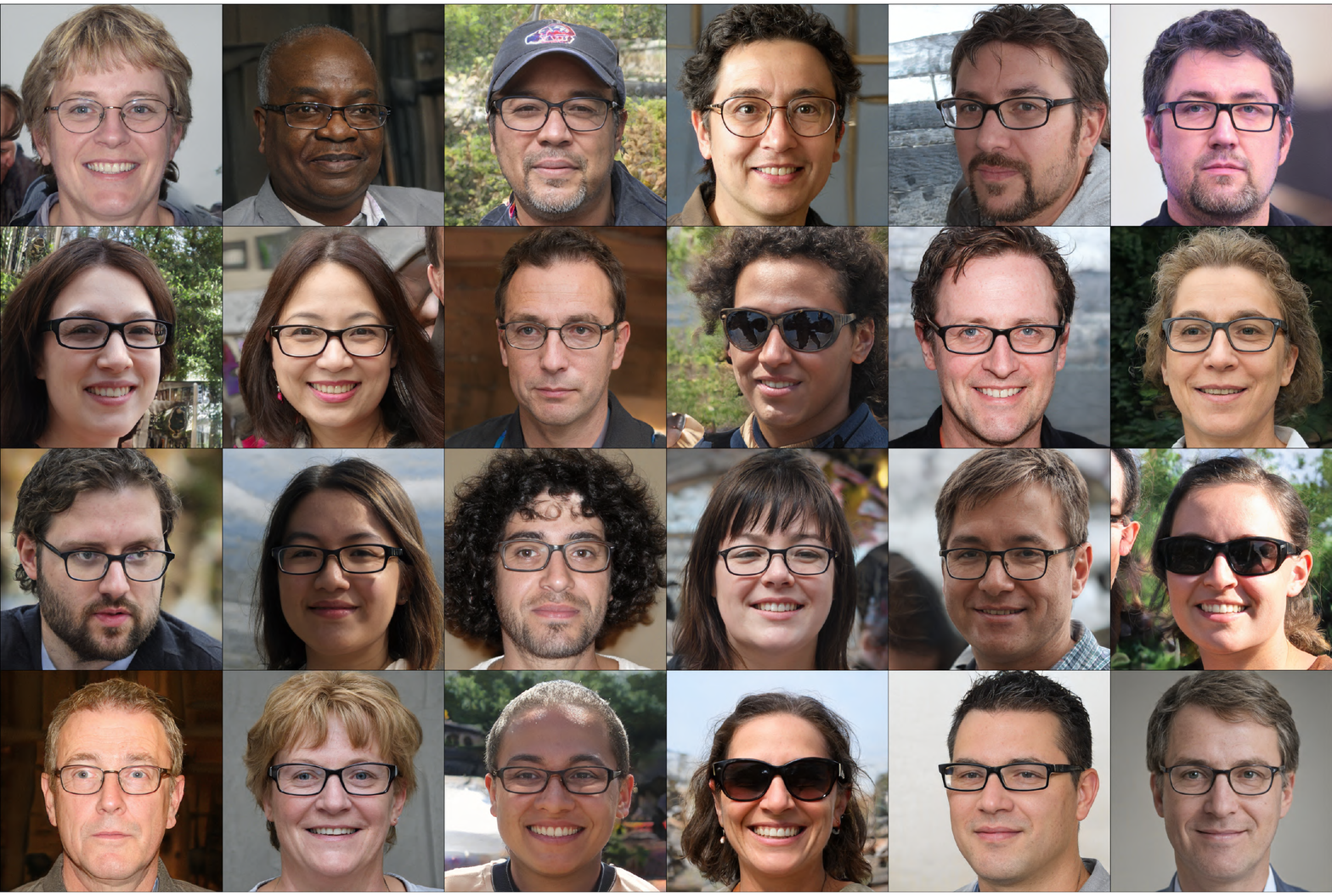}
    \caption{Ours  (\texttt{glasses=1})}
\end{subfigure}

\begin{subfigure}[c]{\textwidth}
    \centering
        \includegraphics[width=\linewidth]{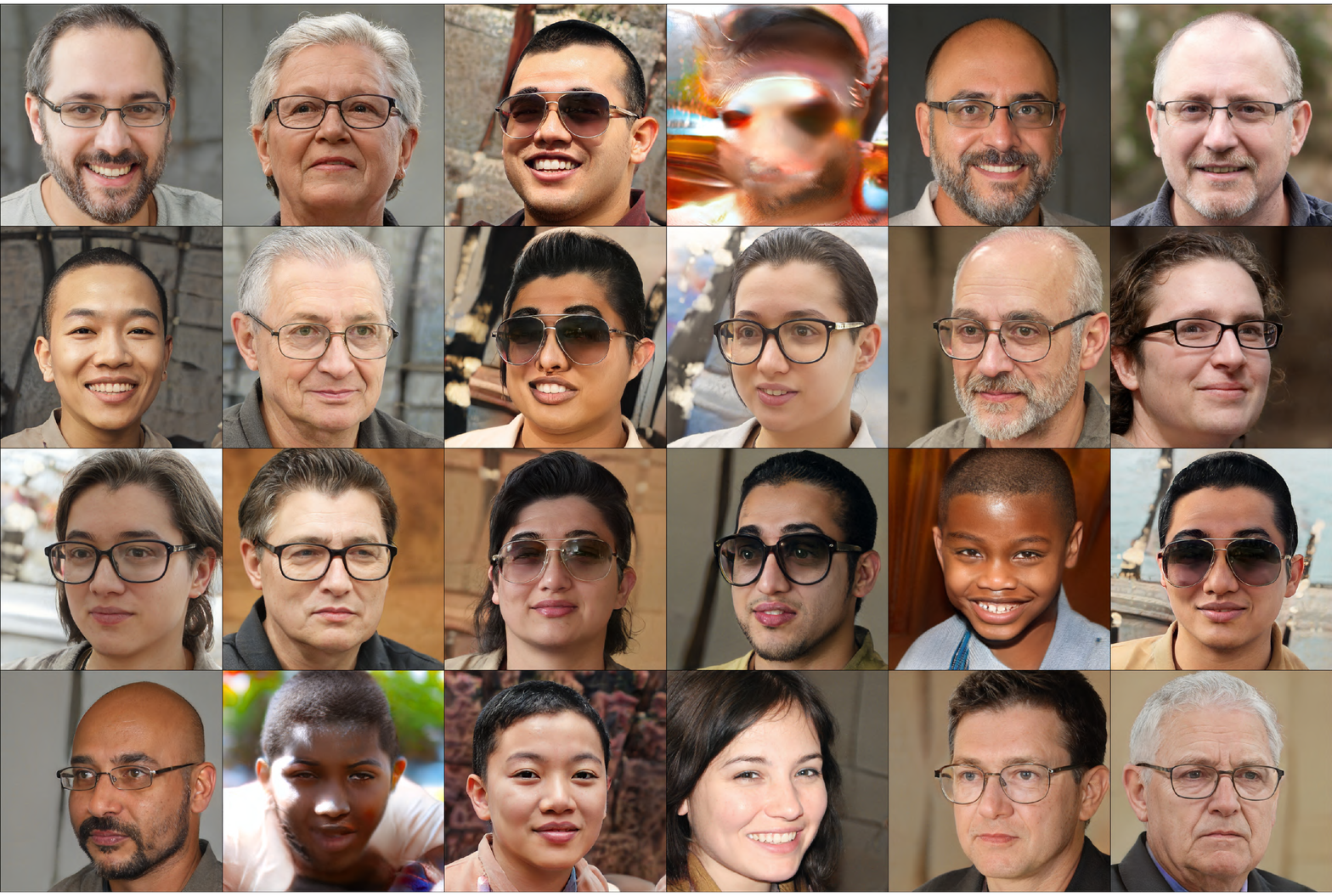}
    \caption{StyleFlow (\texttt{glasses=1})}
\end{subfigure}
    
    \caption{\small Uncurated 1024$\times$1024 conditional sampling results of our method (LACE-ODE) and StyleFlow on \{\texttt{glasses=1}\} of FFHQ, where our method outperforms StyleFlow in terms of image quality (or diversity) and controllability.
    }
    \label{app:ffhq_cond_glasses}
\end{figure}

\begin{figure}[ht]
    \centering
\begin{subfigure}[c]{\textwidth}
    \centering
        \includegraphics[width=\linewidth]{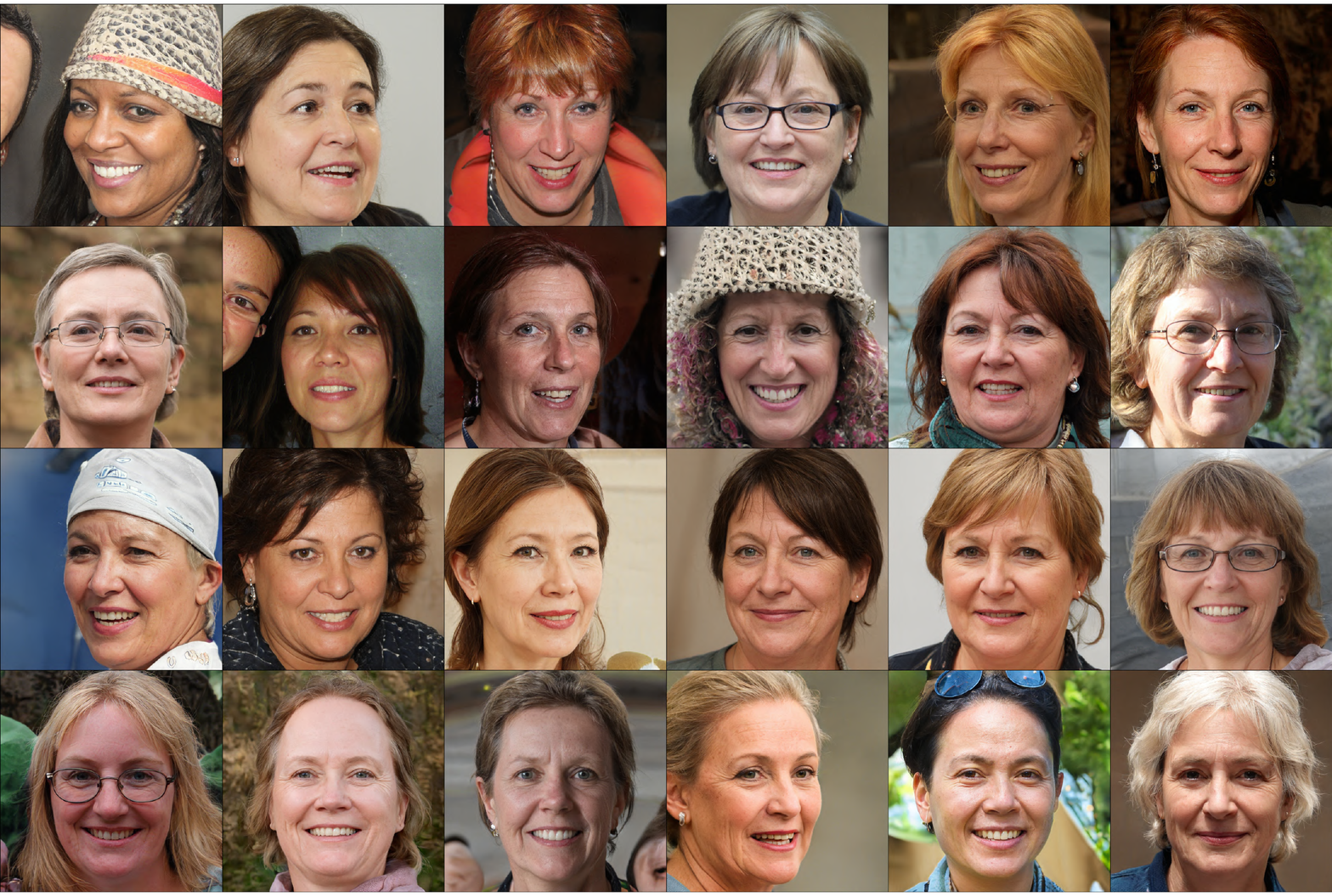}
    \caption{Ours  (\texttt{gender=female,smile=1,age=55})}
\end{subfigure}

\begin{subfigure}[c]{\textwidth}
    \centering
        \includegraphics[width=\linewidth]{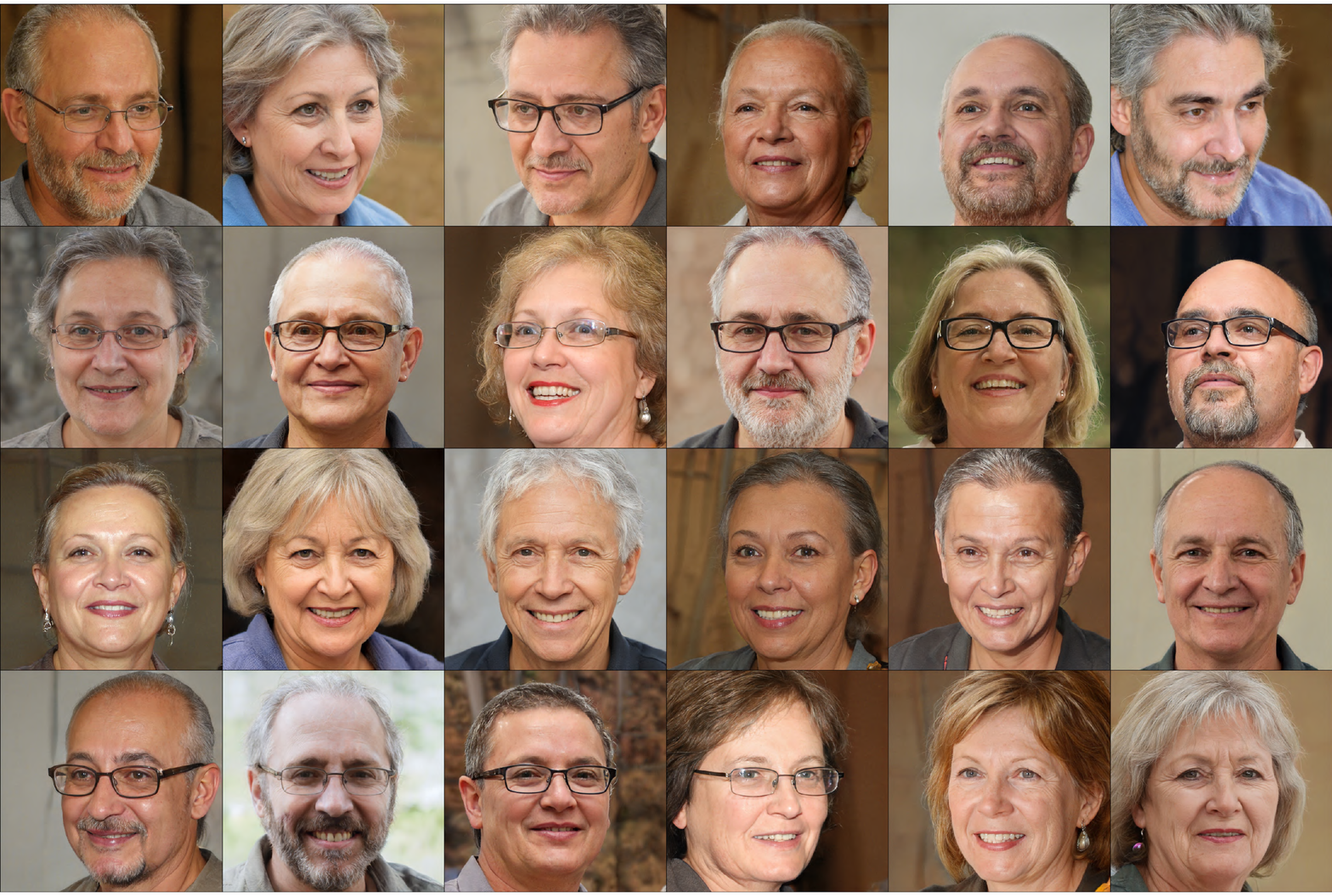}
    \caption{StyleFlow (\texttt{gender=female,smile=1,age=55})}
\end{subfigure}
    
    \caption{\small Uncurated 1024$\times$1024 conditional sampling results of our method (LACE-ODE) and StyleFlow on \{\texttt{gender=female,smile=1,age=55}\} of FFHQ, where our method outperforms StyleFlow in terms of image quality (or diversity) and controllability.
    }
    \label{app:ffhq_cond_gender_smile_age}
\end{figure}

We report the results of our method and baselines on the \texttt{glasses} and \texttt{gender\_smile\_age} of FFHQ with error bars in Table \ref{app:baselines_ffhq}.
The 1024$\times$1024 conditional sampling visual samples of our method (LACE-ODE) and StyleFlow conditioned on \{\texttt{glasses=1}\} and \{\texttt{gender=female,smile=1,age=55}\} of FFHQ can be seen in Figure \ref{app:ffhq_cond_glasses} and Figure \ref{app:ffhq_cond_gender_smile_age}, respectively.

\begin{table}[t]
\vspace{-5pt}
    \caption{\small Comparison between our method and baselines for conditional sampling on the \texttt{glasses} and \texttt{gender\_smile\_age} of FFHQ, respectively. For notations, Train -- training time, Infer -- inference time (m: minute, s: second), which refer to the single GPU time for generating a batch of 16 images, 
    $\eta$ is the LD step size, and $N$ is the number of predictor steps in the PC sampler. 
    }
    \vspace{2pt}
\begin{subtable}[t]{\textwidth}
    \centering
    \footnotesize\addtolength{\tabcolsep}{-4pt}
    \begin{tabular}{c|c|*{3}{c}|*{5}{c}}
        \hline
         
      \multirow{2}{*}{Methods} &
      \multirow{2}{*}{Train} &
      \multicolumn{3}{c|}{\texttt{glasses}} &
      \multicolumn{5}{c}{\texttt{gender\_smile\_age}} \\
      \cline{3-10} 
      & & Infer &
      FID$\downarrow$ &
      $\text{ACC}_{\text{gl}}$$\uparrow$ & Infer & FID$\downarrow$ &
      $\text{ACC}_{\text{ge}}$$\uparrow$ & $\text{ACC}_{\text{s}}$$\uparrow$ & $\text{ACC}_{\text{a}}$$\uparrow$  
      \\
         \hline
          StyleFlow \citep{abdal2020styleflow} &
          50m & \textbf{0.61s} & 42.08\text{\tiny $\pm$.38} & 0.899\text{\tiny $\pm$.007} 
          & \textbf{0.61s} & 43.88\text{\tiny $\pm$.73} & 0.718\text{\tiny $\pm$.031} & 0.870\text{\tiny $\pm$.010} & 0.874\text{\tiny $\pm$.009} 
          \\
        \hline
        Latent-JEM ($\eta$=0.1) & 
        15m & 0.69s & 22.83\text{\tiny $\pm$.19} & 0.765\text{\tiny $\pm$.012} & 
        0.93s & 22.74\text{\tiny $\pm$.08} & 0.878\text{\tiny $\pm$.001} & 0.953\text{\tiny $\pm$.005} & 0.843\text{\tiny $\pm$.001}
        \\
        Latent-JEM ($\eta$=0.01) & 
        15m & 0.69s & 21.58\text{\tiny $\pm$.10} & 0.750\text{\tiny $\pm$.004} 
        & 0.93s & \textbf{21.98\text{\tiny $\pm$.14}} & 0.755\text{\tiny $\pm$.003} & 0.946\text{\tiny $\pm$.009} & 0.831\text{\tiny $\pm$.002}
        \\
\hline
\hline
        LACE-PC ($N$=100) &
        \textbf{2m} & 1.29s & 21.48\text{\tiny $\pm$.26} & 0.943\text{\tiny $\pm$.005}  
        & 2.65s & 24.31\text{\tiny $\pm$.36} & 0.951\text{\tiny $\pm$.007} & 0.922\text{\tiny $\pm$.007} & 0.896\text{\tiny $\pm$.001}  
        \\
        LACE-PC ($N$=200) &
        \textbf{2m} & 2.20s & 21.38\text{\tiny $\pm$.37} & 0.925\text{\tiny $\pm$.003}
        & 4.62s & 23.86\text{\tiny $\pm$.08} & 0.949\text{\tiny $\pm$.004} & 0.914\text{\tiny $\pm$.008} & 0.894\text{\tiny $\pm$.001}
        \\
        LACE-LD &
        \textbf{2m} & 1.15s & \textbf{20.92\text{\tiny $\pm$.15}} & \textbf{0.998\text{\tiny $\pm$.001}} 
        & 2.40s & 22.97\text{\tiny $\pm$.14} & 0.955\text{\tiny $\pm$.004} & 0.960\text{\tiny $\pm$.002} & \textbf{0.913\text{\tiny $\pm$.001}}
        \\
        LACE-ODE & 
        \textbf{2m} & {0.68s} & \textbf{20.93\text{\tiny $\pm$.14}} & \textbf{0.998\text{\tiny $\pm$.001}} 
        & 4.81s & 24.52\text{\tiny $\pm$.94} & \textbf{0.969\text{\tiny $\pm$.004}} & \textbf{0.982\text{\tiny $\pm$.006}} & \textbf{0.914\text{\tiny $\pm$.001}}
        \\
         \hline
    \end{tabular}
\end{subtable}
    \label{app:baselines_ffhq}
    \vspace{-10pt}
\end{table}


\paragraph{On reducing inference time}

As we can see from Table \ref{app:baselines_ffhq}, the inference time of our method increases with the number of attributes. 
In our current setting, each attribute classifier is parametrized by a separate (384-256-128) MLP network (Table \ref{tab:clf_arc} in the Appendix). That is, when conditioning on $n$ attributes, we have $n$ separate MLP networks. We found that the inference time of our method largely depends on the number of MLP networks. Accordingly, if we use a single MLP network with the same size and multiple prediction heads, each of which corresponds to one attribute, we can reduce the inference time without sacrificing the controllable generation performance. 

We run our method for conditional sampling with the increasing number of attributes (1-5). Without loss of generality, we consider the test case: “\texttt{glasses}” (1), “\texttt{age, glasses}” (2), “\texttt{smile, age, glasses}” (3), “\texttt{gender, smile, age, glasses}” (4),  “\texttt{yaw, gender, smile, age, glasses}” (5). The inference time for different numbers of attributes is listed in Table \ref{app:infer_time_acc}(a). Note that “separate” denotes the current setting where we use $n$ separate MLP networks for $n$ attributes, and “single” denotes the new setting where we use a single MLP network with the same size and $n$ prediction heads for $n$ attributes. We can see that although the inference time increases with the number of attributes in both cases, the new setting (“single”) has much smaller inference time, and the advantage becomes larger with more attributes. 
Meanwhile, the performances remain similar. For instance, in the case of conditioning “\texttt{yaw, gender, smile, age, glasses}” (5), the ACCs of the two settings “separate” and “single” are shown in Table \ref{app:infer_time_acc}(b).

\begin{table}[ht]
\vspace{-6pt}
    \centering
    \captionof{table}{\small (a) Inference time (Infer) vs. number of attributes (\#attributes), and (b) ACCs of the two settings “separate” and “single” in the case of conditioning “\texttt{yaw, gender, smile, age, glasses}” (5), where “separate” means that we use $n$ separate MLP networks for $n$ attributes (i.e., the current setting), and “single” means that we use a single MLP network with the same size and $n$ prediction heads for $n$ attributes (i.e., the new setting). We can see that the new setting (“single”) has much smaller inference time.
    }
    \vspace{-4pt}
    
\footnotesize\addtolength{\tabcolsep}{-4pt}
  \begin{subtable}[t]{0.45\textwidth}
    \centering
    \caption{Inference time vs. number of attributes}
    \begin{tabular}{c|ccccc}
        \hline
          \#attributes & 1 & 2 & 3 & 4 & 5 \\
         \hline
          Infer ("separate") &
          0.68 & 2.34 & 4.36 & 6.65 & 7.84
          \\
          Infer ("single")  & 
          0.68 & 1.70 & 2.25 & 2.58 & 2.63 
        \\
\hline
    \end{tabular}
\end{subtable}
\begin{subtable}[t]{0.53\textwidth}
    \centering
    \caption{ACCs of the two settings: “separate” and “single”}
    \begin{tabular}{c|ccccc}
        \hline
          attribute name & ${\text{yaw}}$ & ${\text{gender}}$ & ${\text{smile}}$ & ${\text{age}}$ & ${\text{glasses}}$ \\
         \hline
          ACC ("separate") &
          0.927 & 0.956 & 0.953 & 0.897 & 0.994
          \\
          ACC ("single")  & 
          0.904 & 0.973 & 0.954 & 0.892 & 0.984 
        \\
\hline
    \end{tabular}
\end{subtable}
      \label{app:infer_time_acc}
      \vspace{-5pt}
\end{table}

\vspace{-5pt}
\subsection{More results of sequential editing}
\label{app:seq}
\vspace{-3pt}

In sequential editing, we apply the \textit{subsection selection} strategy as proposed in StyleFlow \citep{abdal2020styleflow} to alleviate the background change and the \textit{reweighting of energy functions} to improve the disentanglement quality. We now introduce them in the following.

\vspace{-4pt}
\paragraph{Subsection selection} 
By observing the hierarchical structure of StyleGAN2 \citep{Karras2020stylegan2}, we can apply the updated $w$ only to a subset (with different indices) of the $W\text{+} \in \mathbb{R}^{18\times512}$ space, depending on the nature of each edit \citep{abdal2020styleflow}. For instance, the head pose (such as \texttt{yaw} and \texttt{pitch}) is a coarse-grained feature and is expected to only affect the early layers of the StyleGAN2 generator. Thus, it will cause less unintentional changes (such as background and other fine-grained attributes) by applying the updated $w$ to the early layers only during editing the head pose. StyleFlow has empirically identified the index subsets of the edits that work the best for their method, including \texttt{smile} (4 -- 5), \texttt{yaw} (0 -- 3), \texttt{pitch} (0 -- 3), \texttt{age} (4 -- 7), \texttt{gender} (0 -- 7), \texttt{glasses} (0 -- 5), \texttt{bald} (0 -- 5) and
\texttt{beard} (5 -- 7 and 10). To keep it simple, we directly use the above index subsets of the edits for our method.

\vspace{-4pt}
\paragraph{Reweighting of energy functions}

In reweighting of energy functions, we slightly modify the term $E_\theta(z, \{c_j\}_{j=1}^i)$ in the joint energy function of the $i$-th edit into
\begin{align*}
    E_\theta(z, \{c_j\}_{j=1}^i) = \alpha_0 \sum_{j=1}^{i -1} E_\theta ( c_j | g(z) ) + \alpha_1 E_\theta ( c_i | g(z) ) + \frac{1}{2} \|z\|^2_2
\end{align*}
where we introduce two reweighting coefficients: $\alpha_0$ for previous edited attributes and $\alpha_1$ for the current $i$-th attribute. If we set $\alpha_0=\alpha_1=1$, the above equation reduces to Eq. (\ref{energy_final}). We find that slightly increasing $\alpha_1$ and decreasing $\alpha_0$ can make our method pay more attention the current edit while less modifying previously edited attributes. In experiments, we set $\alpha_0 = 0.2$, and we set $\alpha_1 = 10 $ for continuous attributes and $\alpha_1 = 5$ for discrete attributes.

We report both the final results \textit{after all edits} and the results of every individual edit
with error bars in Table \ref{app:seq_edit}, where we edit the attributes [\texttt{yaw}, \texttt{smile}, \texttt{age}, \texttt{glasses}] in a sequential order. 
Besides, we also perform ablation studies on the impact of subset selection and reweighting of energy functions (see Appendix \ref{app:exp_setting} for details) on our method. 
From Table \ref{app:seq_edit}, we can see that adding subset selection or reweighting of energy functions does not change much the final ACC and FID scores. However, the disentanglement quality (DES) and identity preservation (ID) both get improved, after adding both subset selection and reweighting of energy functions.

\begin{table}[ht]
    \caption{\small Comparison between our method and StyleFlow \citep{abdal2020styleflow} for sequential editing on FFHQ, where we edit each attribute of [\texttt{yaw}, \texttt{smile}, \texttt{age}, \texttt{glasses}] in a sequential order. Note that "w/o \texttt{ss}" means no subset selection, "w/o \texttt{rw}" means no reweighting of energy functions, and $\psi$ denotes the truncation coefficient of StyleGAN2.}
    \label{tab_seq_edit_all}
    \vspace{3pt}
\begin{subtable}[t]{\textwidth}
    \centering
    \footnotesize\addtolength{\tabcolsep}{-4pt}
    \begin{tabular}{c|cc|cc|cc}
        \hline
         
      \multirow{2}{*}{Methods} &
      \multicolumn{2}{c|}{+\texttt{yaw}} &
      \multicolumn{2}{c|}{+\texttt{smile}} &
      \multicolumn{2}{c}{+\texttt{age}}\\
      \cline{2-7} 
      & $\text{DES}_1$$\uparrow$ & ID$\downarrow$ & $\text{DES}_2$$\uparrow$ & ID$\downarrow$ & $\text{DES}_3$$\uparrow$ & ID$\downarrow$ \\
         \hline
          StyleFlow &
          {0.568\text{\tiny $\pm$.012}} & {0.188\text{\tiny $\pm$.014}} &
          0.570\text{\tiny $\pm$.029} & \textbf{0.062\text{\tiny $\pm$.001}} &
          0.398\text{\tiny $\pm$.004} & 0.327\text{\tiny $\pm$.015}
          \\
        LACE-ODE ($\psi$=0.5, w/o \texttt{ss}, w/o \texttt{rw}) & 
        0.534\text{\tiny $\pm$.016} & 0.179\text{\tiny $\pm$.011} & {0.745\text{\tiny $\pm$.017}} & 0.117\text{\tiny $\pm$.006} & {0.381\text{\tiny $\pm$.010}} & {0.175\text{\tiny $\pm$.008}} 
        \\
        LACE-ODE ($\psi$=0.5, w/o \texttt{rw}) & 
        0.475\text{\tiny $\pm$.012} & \textbf{0.141\text{\tiny $\pm$.012}} & {0.633\text{\tiny $\pm$.029}} & 0.103\text{\tiny $\pm$.006} & {0.260\text{\tiny $\pm$.039}} & \textbf{{0.151\text{\tiny $\pm$.008}}} 
        \\
        LACE-ODE ($\psi$=0.5) & 
        \textbf{0.623\text{\tiny $\pm$.014}} & 0.204\text{\tiny $\pm$.016} & \textbf{0.875\text{\tiny $\pm$.026}} & 0.082\text{\tiny $\pm$.007} & \textbf{0.453\text{\tiny $\pm$.016}} & {0.197\text{\tiny $\pm$.012}}
        \\
        LACE-ODE ($\psi$=0.7) & 
        0.559\text{\tiny $\pm$.015} & 0.211\text{\tiny $\pm$.011} & {0.825\text{\tiny $\pm$.024}} & 0.091\text{\tiny $\pm$.007} & {0.408\text{\tiny $\pm$.005}} & {0.216\text{\tiny $\pm$.011}} 
        \\
         \hline
    \end{tabular}
    \vspace{1pt}
\end{subtable}

\begin{subtable}[t]{\textwidth}
    \centering
    \footnotesize\addtolength{\tabcolsep}{-3.5pt}
    \begin{tabular}{cc|ccccccc}
        \hline
         
      \multicolumn{2}{c|}{+\texttt{glasses}} &
      \multicolumn{7}{c}{All} \\
      \cline{1-9} 
      $\text{DES}_4$$\uparrow$ & ID$\downarrow$ & $\text{DES}$$\uparrow$ & ID$\downarrow$ & FID$\downarrow$ & $\text{ACC}_{\texttt{y}}$$\uparrow$ & $\text{ACC}_{\texttt{s}}$$\uparrow$ & $\text{ACC}_{\texttt{a}}$$\uparrow$ & $\text{ACC}_{\texttt{g}}$$\uparrow$
      \\
         \hline
         0.741\text{\tiny $\pm$.033} & \textbf{0.188\text{\tiny $\pm$.006}} &
         0.569\text{\tiny $\pm$.009} & {0.549\text{\tiny $\pm$.016}} & 44.13\text{\tiny $\pm$1.62} &
         \textbf{0.947\text{\tiny $\pm$.004}} &
         0.773\text{\tiny $\pm$.022} &
         0.817\text{\tiny $\pm$.007} &
         0.876\text{\tiny $\pm$.009}
          \\
        {0.956\text{\tiny $\pm$.018}} & {0.213\text{\tiny $\pm$.016}} & {0.654\text{\tiny $\pm$.009}} & {0.523\text{\tiny $\pm$.005}}  & {27.46\text{\tiny $\pm$0.16}} & 0.941\text{\tiny $\pm$.003} &
        {0.968\text{\tiny $\pm$.017}} &  \textbf{0.897\text{\tiny $\pm$.003}} & {0.975\text{\tiny $\pm$.004}}
        \\
        {0.942\text{\tiny $\pm$.010}} & {0.205\text{\tiny $\pm$.019}} & {0.578\text{\tiny $\pm$.015}} & \textbf{0.492\text{\tiny $\pm$.008}}  & {27.90\text{\tiny $\pm$0.09}} & 0.940\text{\tiny $\pm$.004} &
        \textbf{0.969\text{\tiny $\pm$.009}} &  {0.884\text{\tiny $\pm$.005}} & {0.975\text{\tiny $\pm$.005}}
        \\
        \textbf{0.989\text{\tiny $\pm$.007}} & {0.216\text{\tiny $\pm$.019}} & \textbf{0.735\text{\tiny $\pm$.009}} & {0.501\text{\tiny $\pm$.009}}  & {27.94\text{\tiny $\pm$0.08}} & 0.938\text{\tiny $\pm$.004} &
        {0.956\text{\tiny $\pm$.013}} &  {0.881\text{\tiny $\pm$.006}} & \textbf{0.997\text{\tiny $\pm$.001}}
        \\
        {0.971\text{\tiny $\pm$.014}} & {0.209\text{\tiny $\pm$.015}} & {0.691\text{\tiny $\pm$.010}} & {0.532\text{\tiny $\pm$.006}}  & \textbf{21.90\text{\tiny $\pm$0.23}} & 0.933\text{\tiny $\pm$.004} &
        {0.941\text{\tiny $\pm$.015}} &  {0.871\text{\tiny $\pm$.008}} & {0.983\text{\tiny $\pm$.003}}
        \\
         \hline
    \end{tabular}
\end{subtable}
\label{app:seq_edit}
\vspace{-10pt}
\end{table}

\begin{figure}[ht]
    \centering
\begin{subfigure}[c]{0.9\textwidth}
    \centering
        \includegraphics[width=\linewidth]{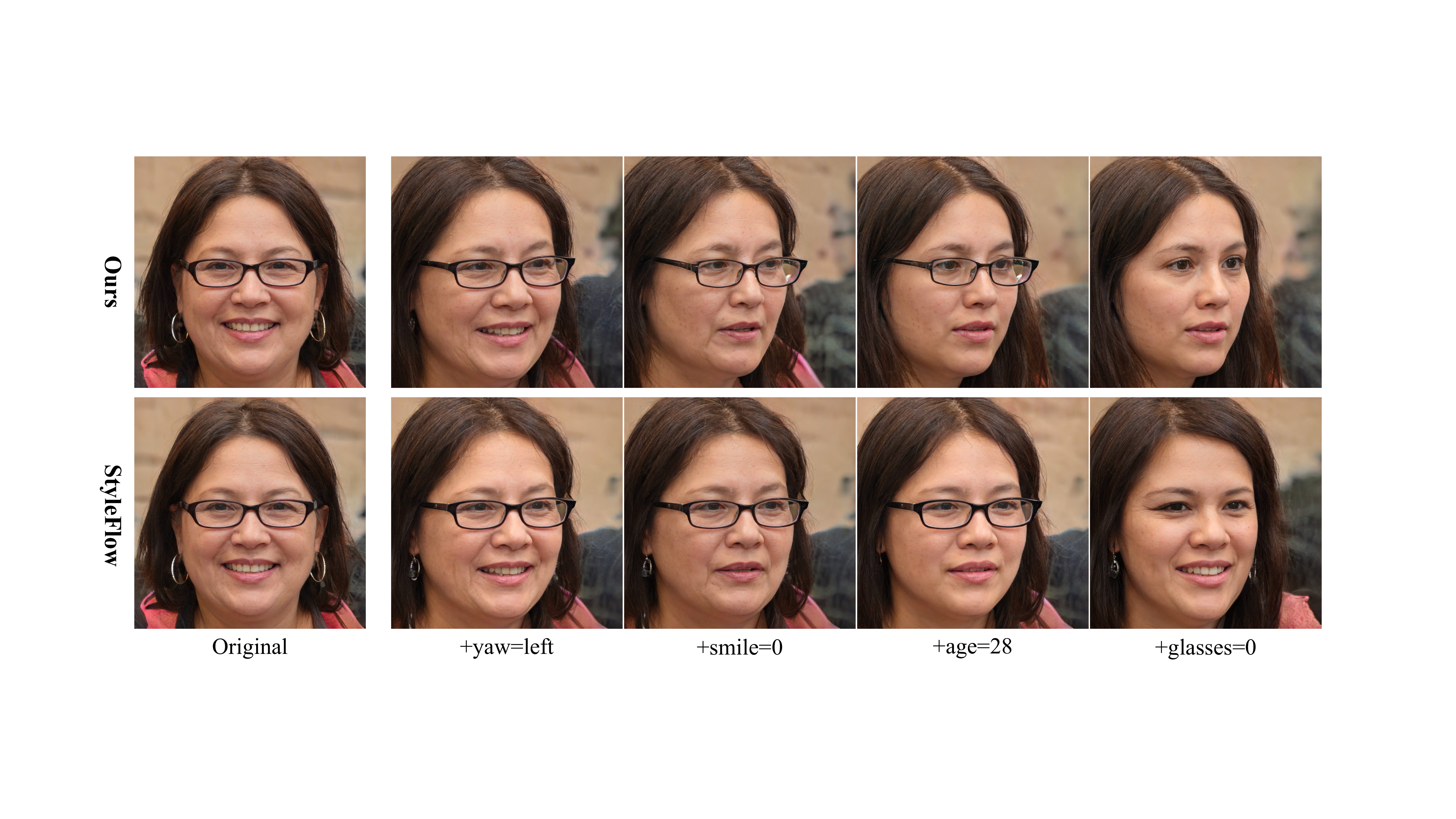}
        \includegraphics[width=\linewidth]{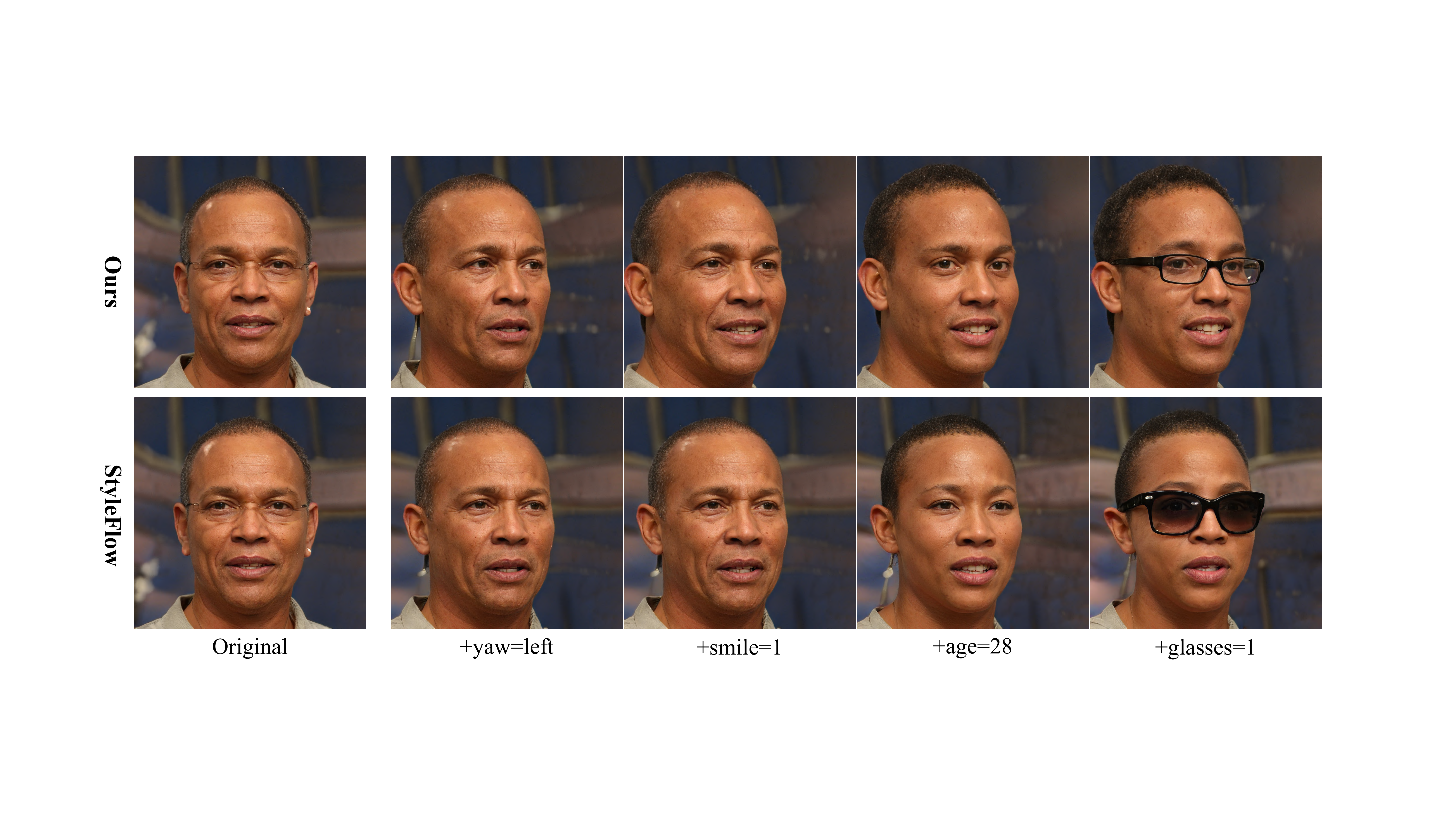}
\end{subfigure}

\begin{subfigure}[c]{0.9\textwidth}
    \centering
        \includegraphics[width=\linewidth]{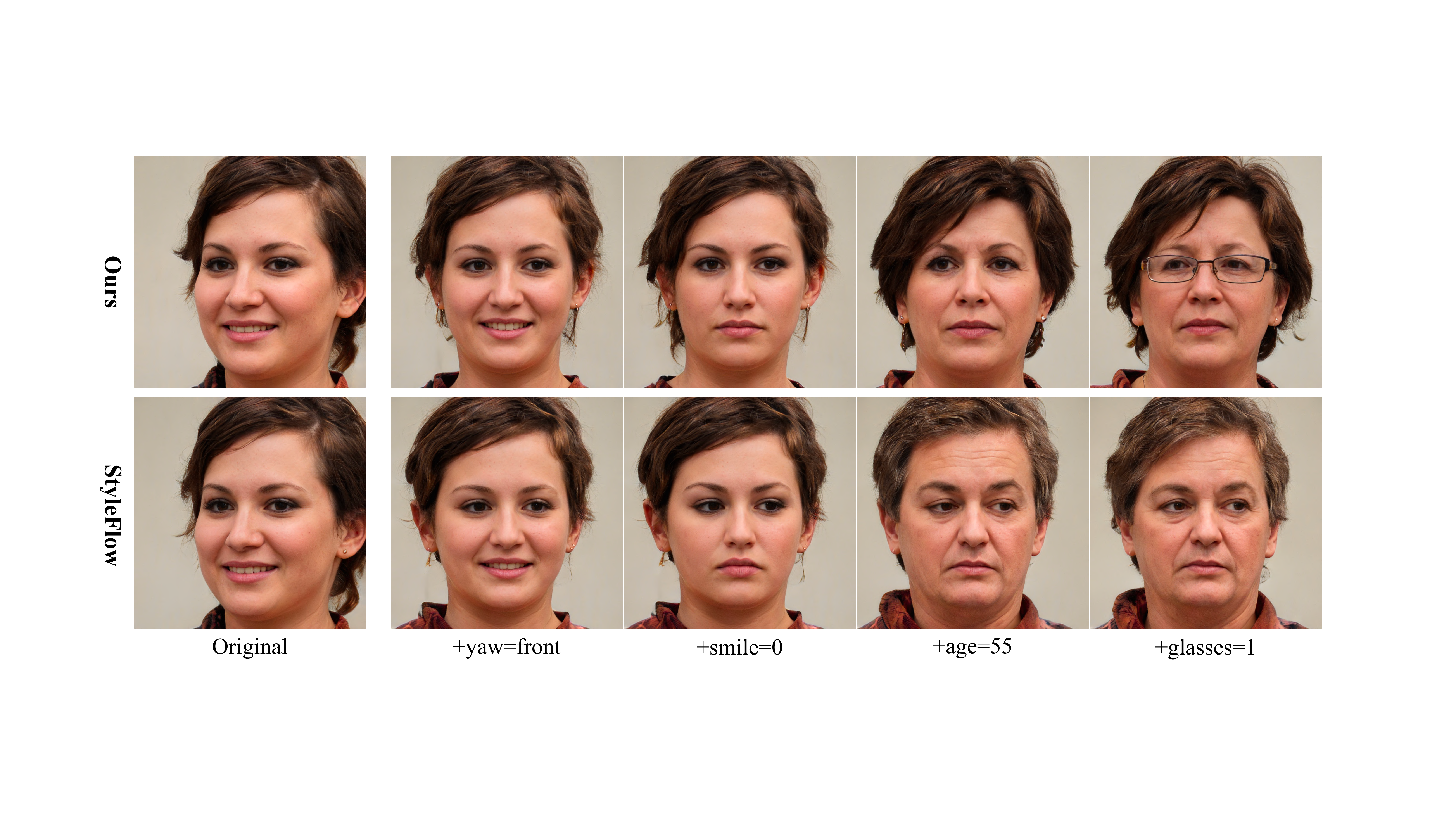}
        \includegraphics[width=\linewidth]{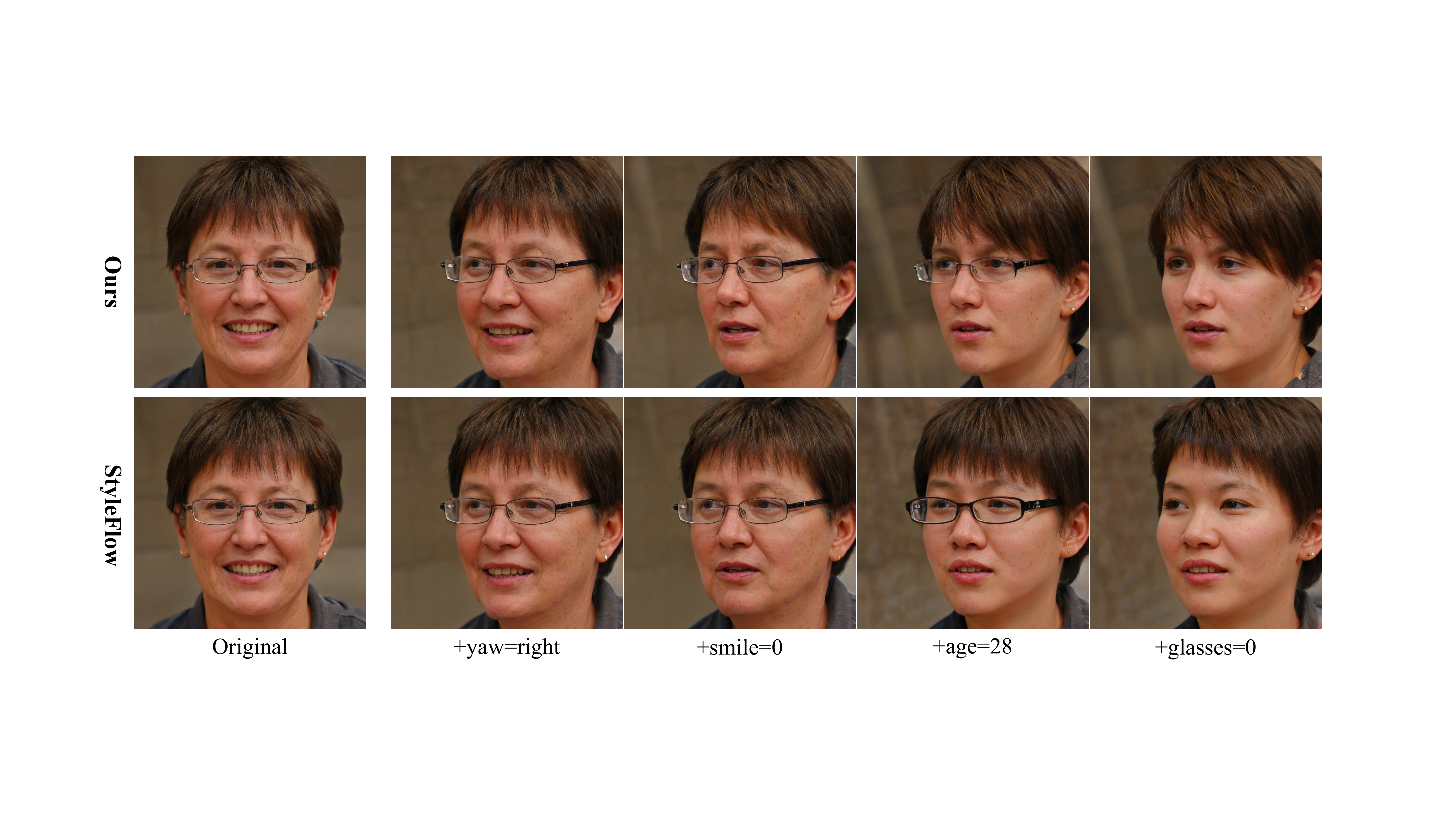}
\end{subfigure}
    
    \caption{\small The sequential editing results of our method (LACE-ODE) and StyleFlow on FFHQ with a sequence of [\texttt{yaw,smile,age,glasses}]. Note that `+' means the current editing is built upon the last edited images. For instance, \texttt{+smile=0} refers to changing the last edited images to make them not smile. Overall, compared to StyleFlow, our method can successfully perform each edit while less affecting other attributes and face identities. 
    }
    \label{app:ffhq_seq_edit}
\end{figure}

The 1024$\times$1024 sequential editing visual samples of our method (LACE-ODE) and StyleFlow on FFHQ with a sequence of [\texttt{yaw,smile,age,glasses}] can be seen in Figure \ref{app:ffhq_seq_edit}.

\paragraph{Randomize ordering of attributes} We also randomly perturb the ordering of attributes and report the quantitative results in Table \ref{app:seq_edit_v2}, where we edit the attributes [\texttt{age}, \texttt{yaw}, \texttt{glasses}, \texttt{smile}] in a sequential order. The results 
remains similar to Table \ref{app:seq_edit}: 1) our method largely outperform StyleFlow regarding editing quality and image quality, and 2) adding both subset selection and reweighting of energy functions can largely the disentanglement quality (DES).

\begin{table}[ht]
    \caption{\small Comparison between our method and StyleFlow \citep{abdal2020styleflow} for sequential editing on FFHQ, where we edit each attribute of [\texttt{age}, \texttt{yaw}, \texttt{glasses}, \texttt{smile}] in a sequential order. Note that "w/o \texttt{ss}" means no subset selection, "w/o \texttt{rw}" means no reweighting of energy functions, and $\psi$ denotes the truncation coefficient of StyleGAN2.}
    \label{tab_seq_edit_all_v2}
    \vspace{3pt}
\begin{subtable}[t]{\textwidth}
    \centering
    \footnotesize\addtolength{\tabcolsep}{-4pt}
    \begin{tabular}{c|cc|cc|cc}
        \hline
         
      \multirow{2}{*}{Methods} &
      \multicolumn{2}{c|}{+\texttt{age}} &
      \multicolumn{2}{c|}{+\texttt{yaw}} &
      \multicolumn{2}{c}{+\texttt{glasses}}\\
      \cline{2-7} 
      & $\text{DES}_1$$\uparrow$ & ID$\downarrow$ & $\text{DES}_2$$\uparrow$ & ID$\downarrow$ & $\text{DES}_3$$\uparrow$ & ID$\downarrow$ \\
         \hline
          StyleFlow &
          {0.402\text{\tiny $\pm$.003}} & {0.329\text{\tiny $\pm$.011}} &
          0.599\text{\tiny $\pm$.010} & \textbf{0.187\text{\tiny $\pm$.004}} &
          0.727\text{\tiny $\pm$.032} & \textbf{0.187\text{\tiny $\pm$.006}}
          \\
        LACE-ODE ($\psi$=0.5, w/o \texttt{ss}, w/o \texttt{rw}) & 
        0.491\text{\tiny $\pm$.013} & \textbf{0.167\text{\tiny $\pm$.009}} & {0.498\text{\tiny $\pm$.007}} & 0.192\text{\tiny $\pm$.011} & {0.889\text{\tiny $\pm$.014}} & {0.219\text{\tiny $\pm$.009}} 
        \\
        LACE-ODE ($\psi$=0.5, w/o \texttt{rw}) & 
        0.497\text{\tiny $\pm$.012} & \textbf{0.167\text{\tiny $\pm$.009}} & {0.499\text{\tiny $\pm$.014}} & 0.192\text{\tiny $\pm$.012} & {0.882\text{\tiny $\pm$.016}} & {{0.220\text{\tiny $\pm$.009}}} 
        \\
        LACE-ODE ($\psi$=0.5) & 
        \textbf{0.558\text{\tiny $\pm$.009}} & 0.222\text{\tiny $\pm$.011} & \textbf{0.693\text{\tiny $\pm$.011}} & 0.273\text{\tiny $\pm$.016} & \textbf{1.003\text{\tiny $\pm$.010}} & {0.196\text{\tiny $\pm$.012}}
        \\
        LACE-ODE ($\psi$=0.7) & 
        0.504\text{\tiny $\pm$.005} & 0.254\text{\tiny $\pm$.007} & {0.624\text{\tiny $\pm$.014}} & 0.281\text{\tiny $\pm$.012} & {0.974\text{\tiny $\pm$.011}} & {0.198\text{\tiny $\pm$.008}} 
        \\
         \hline
    \end{tabular}
   \vspace{1pt}
\end{subtable}

\begin{subtable}[t]{\textwidth}
    \centering
    \footnotesize\addtolength{\tabcolsep}{-3.5pt}
    \begin{tabular}{cc|ccccccc}
        \hline
         
      \multicolumn{2}{c|}{+\texttt{smile}} &
      \multicolumn{7}{c}{All} \\
      \cline{1-9} 
      $\text{DES}_4$$\uparrow$ & ID$\downarrow$ & $\text{DES}$$\uparrow$ & ID$\downarrow$ & FID$\downarrow$ & $\text{ACC}_{\texttt{y}}$$\uparrow$ & $\text{ACC}_{\texttt{s}}$$\uparrow$ & $\text{ACC}_{\texttt{a}}$$\uparrow$ & $\text{ACC}_{\texttt{g}}$$\uparrow$
      \\
         \hline
         0.533\text{\tiny $\pm$.007} & \textbf{0.055\text{\tiny $\pm$.002}} &
         0.565\text{\tiny $\pm$.011} & {0.550\text{\tiny $\pm$.010}} & 44.02\text{\tiny $\pm$1.45} &
         {0.821\text{\tiny $\pm$.008}} &
         \textbf{0.948\text{\tiny $\pm$.001}} &
         0.870\text{\tiny $\pm$.008} &
         0.764\text{\tiny $\pm$.012}
          \\
        {0.800\text{\tiny $\pm$.048}} & {0.099\text{\tiny $\pm$.003}} & {0.669\text{\tiny $\pm$.013}} & \textbf{0.537\text{\tiny $\pm$.011}}  & {27.33\text{\tiny $\pm$0.04}} & \textbf{0.910\text{\tiny $\pm$.006}} &
        {0.937\text{\tiny $\pm$.003}} &  \textbf{0.995\text{\tiny $\pm$.002}} & {0.920\text{\tiny $\pm$.006}}
        \\
        {0.787\text{\tiny $\pm$.051}} & {0.098\text{\tiny $\pm$.004}} & {0.666\text{\tiny $\pm$.009}} & \textbf{0.537\text{\tiny $\pm$.011}}  & {27.28\text{\tiny $\pm$0.16}} & \textbf{0.910\text{\tiny $\pm$.007}} &
        {0.938\text{\tiny $\pm$.003}} &  \textbf{0.995\text{\tiny $\pm$.002}} & {0.918\text{\tiny $\pm$.005}}
        \\
        \textbf{0.929\text{\tiny $\pm$.030}} & {0.091\text{\tiny $\pm$.007}} & \textbf{0.796\text{\tiny $\pm$.007}} & {0.541\text{\tiny $\pm$.010}}  & {27.22\text{\tiny $\pm$0.22}} & 0.905\text{\tiny $\pm$.007} &
        {0.937\text{\tiny $\pm$.003}} &  {0.992\text{\tiny $\pm$.002}} & \textbf{0.964\text{\tiny $\pm$.005}}
        \\
        {0.853\text{\tiny $\pm$.032}} & {0.100\text{\tiny $\pm$.007}} & {0.739\text{\tiny $\pm$.013}} & {0.570\text{\tiny $\pm$.013}}  & \textbf{21.76\text{\tiny $\pm$0.30}} & 0.897\text{\tiny $\pm$.005} &
        {0.929\text{\tiny $\pm$.003}} &  {0.980\text{\tiny $\pm$.009}} & {0.939\text{\tiny $\pm$.001}}
        \\
         \hline
    \end{tabular}
\end{subtable}
\label{app:seq_edit_v2}
\vspace{-8pt}
\end{table}


    

\vspace{-3pt}
\subsection{More results of compositional generation}
\label{app:composition}

\subsubsection{Zero-shot generation}

\begin{figure}[ht]
    \centering
\begin{subfigure}[c]{\textwidth}
    \centering
        \includegraphics[width=\linewidth]{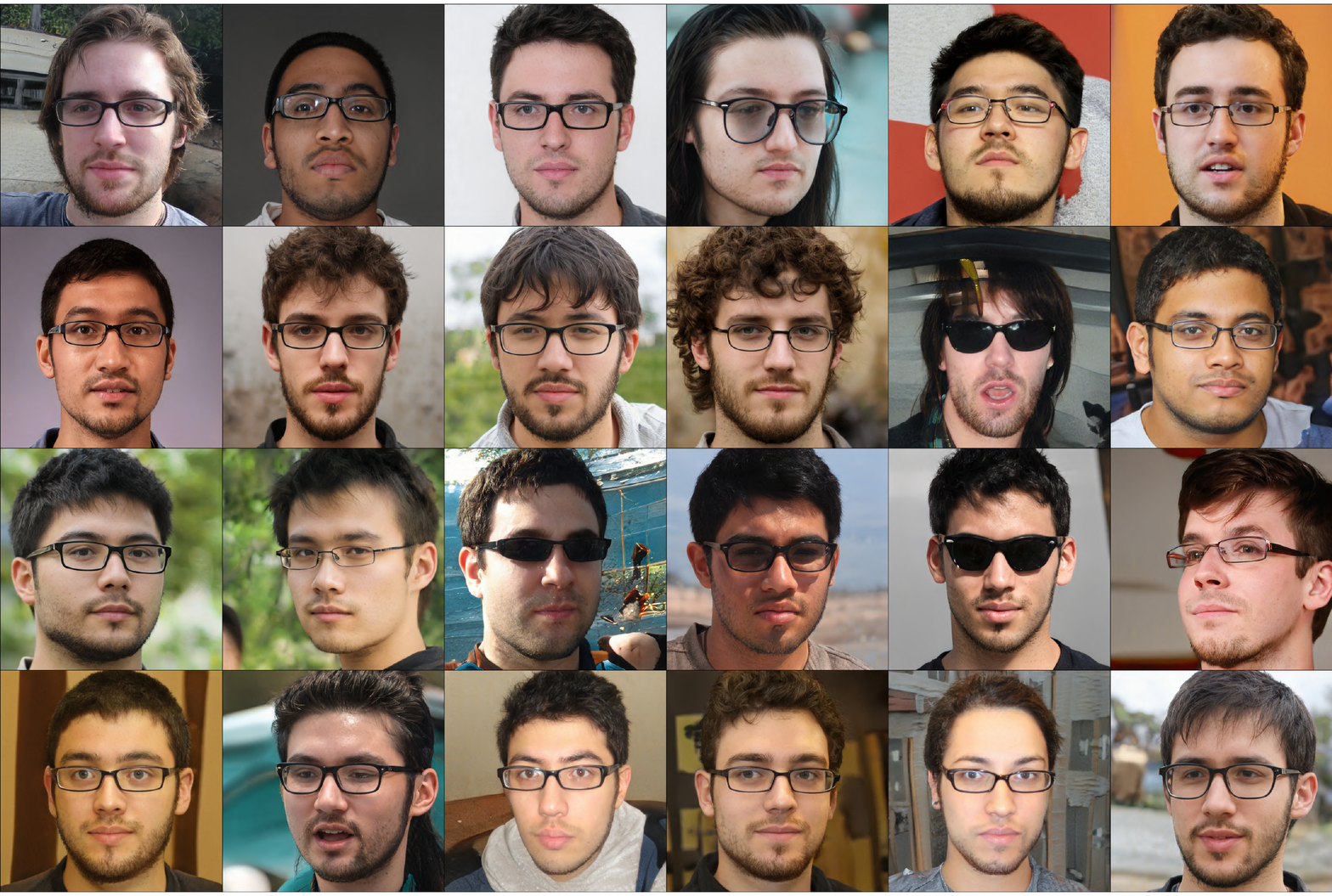}
    \caption{Ours  (\texttt{beard=1,smile=0,glasses=1,age=15})}
\end{subfigure}

\begin{subfigure}[c]{\textwidth}
    \centering
        \includegraphics[width=\linewidth]{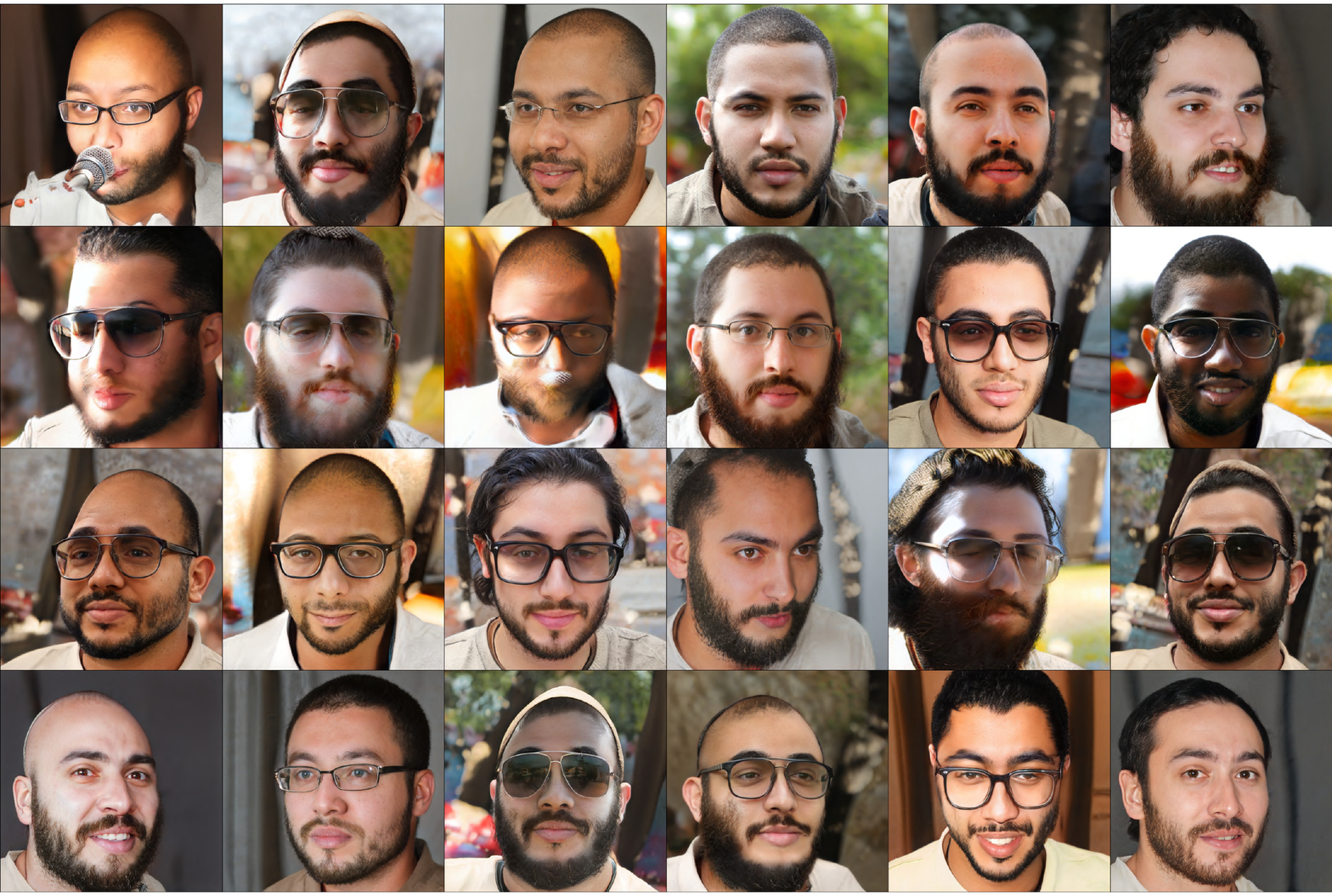}
    \caption{StyleFlow (\texttt{beard=1,smile=0,glasses=1,age=15})}
\end{subfigure}
    
    \caption{\small Uncurated 1024$\times$1024 zero-shot generation results of our method (LACE-ODE) and StyleFlow on the unseen attribute combinations \{\texttt{beard=1,smile=0,glasses=1,age=15}\} of FFHQ, where where our method excels at zero-shot generation while StyleFlow performs significantly worse by either generating low-quality images or completely missing the conditional information.
    }
    \label{app:ffhq_zero_shot_a}
\end{figure}

\begin{figure}[ht]
    \centering
\begin{subfigure}[c]{\textwidth}
    \centering
        \includegraphics[width=\linewidth]{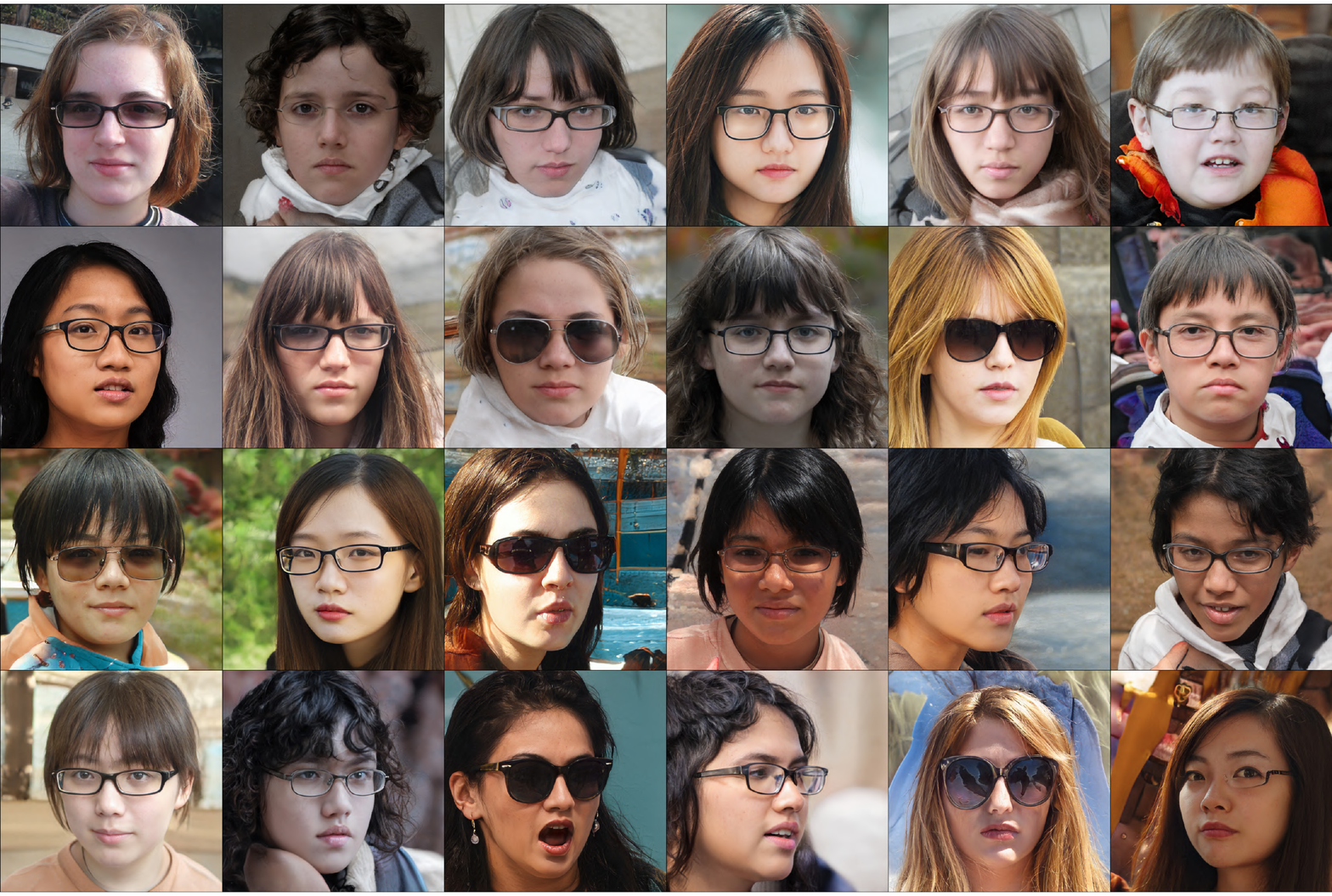}
    \caption{Ours  (\texttt{gender=female,smile=0,glasses=1,age=10})}
\end{subfigure}

\begin{subfigure}[c]{\textwidth}
    \centering
        \includegraphics[width=\linewidth]{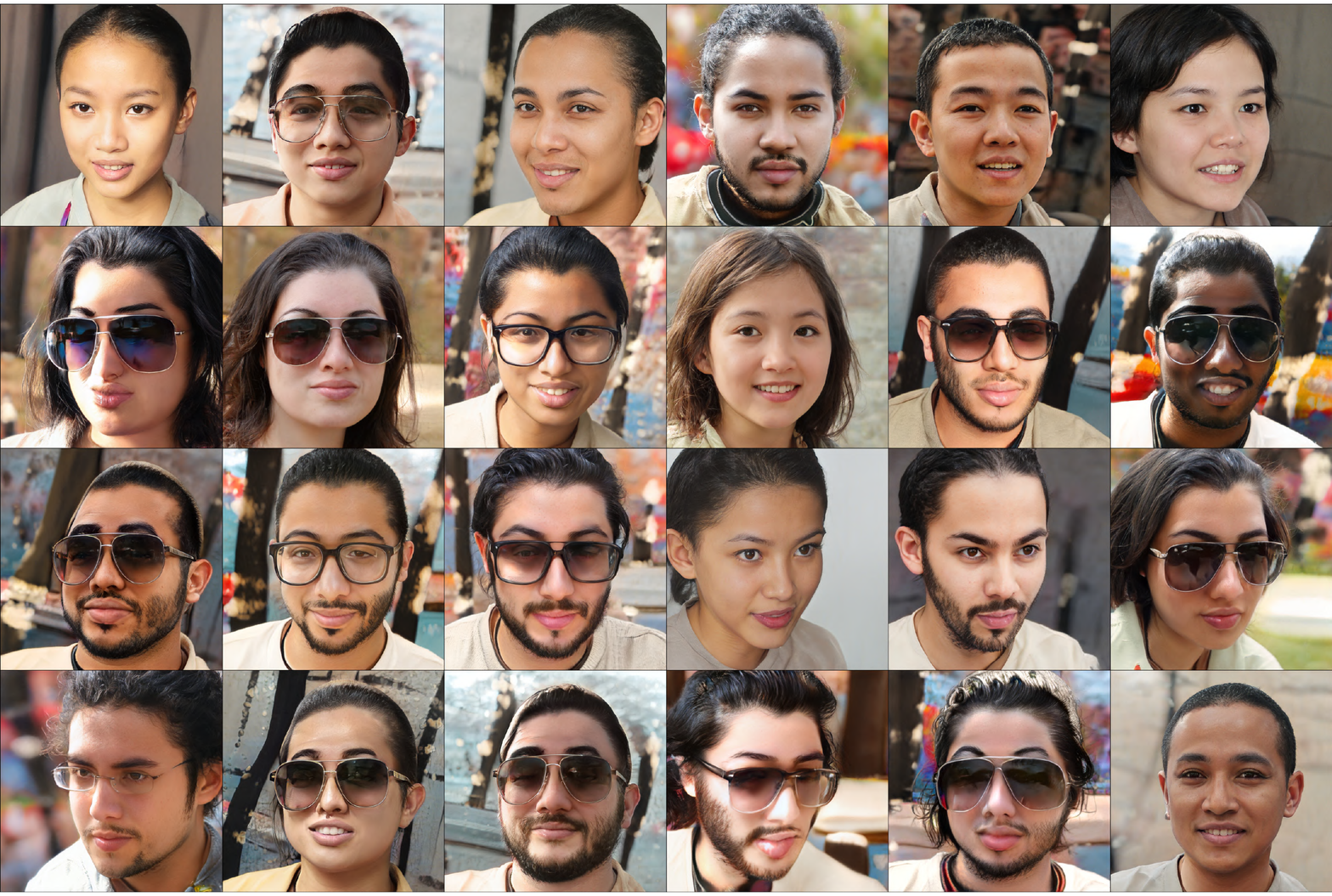}
    \caption{StyleFlow (\texttt{gender=female,smile=0,glasses=1,age=10})}
\end{subfigure}
    
    \caption{\small Uncurated 1024$\times$1024 zero-shot generation results of our method (LACE-ODE) and StyleFlow on unseen attribute combinations \{\texttt{gender=female,smile=0,glasses=1,age=10}\} of FFHQ, where our method excels at zero-shot generation while StyleFlow performs significantly worse by either generating low-quality images or completely missing the conditional information.
    }
    \label{app:ffhq_zero_shot_b}
\end{figure}

The 1024$\times$1024  visual samples of our method (LACE-ODE) and StyleFlow in zero-shot generation on the unseen attribute combinations \{\texttt{beard=1,smile=0,glasses=1,age=15}\} and \{\texttt{gender=female,smile=0,glasses=1,age=10}\} of FFHQ can be seen in Figure \ref{app:ffhq_zero_shot_a} and Figure \ref{app:ffhq_zero_shot_b}, respectively.


\vspace{-6pt}
\subsubsection{Compositions of energy functions}

\begin{figure}[ht]
    \centering
\begin{subfigure}[c]{\textwidth}
    \centering
        \includegraphics[width=\linewidth]{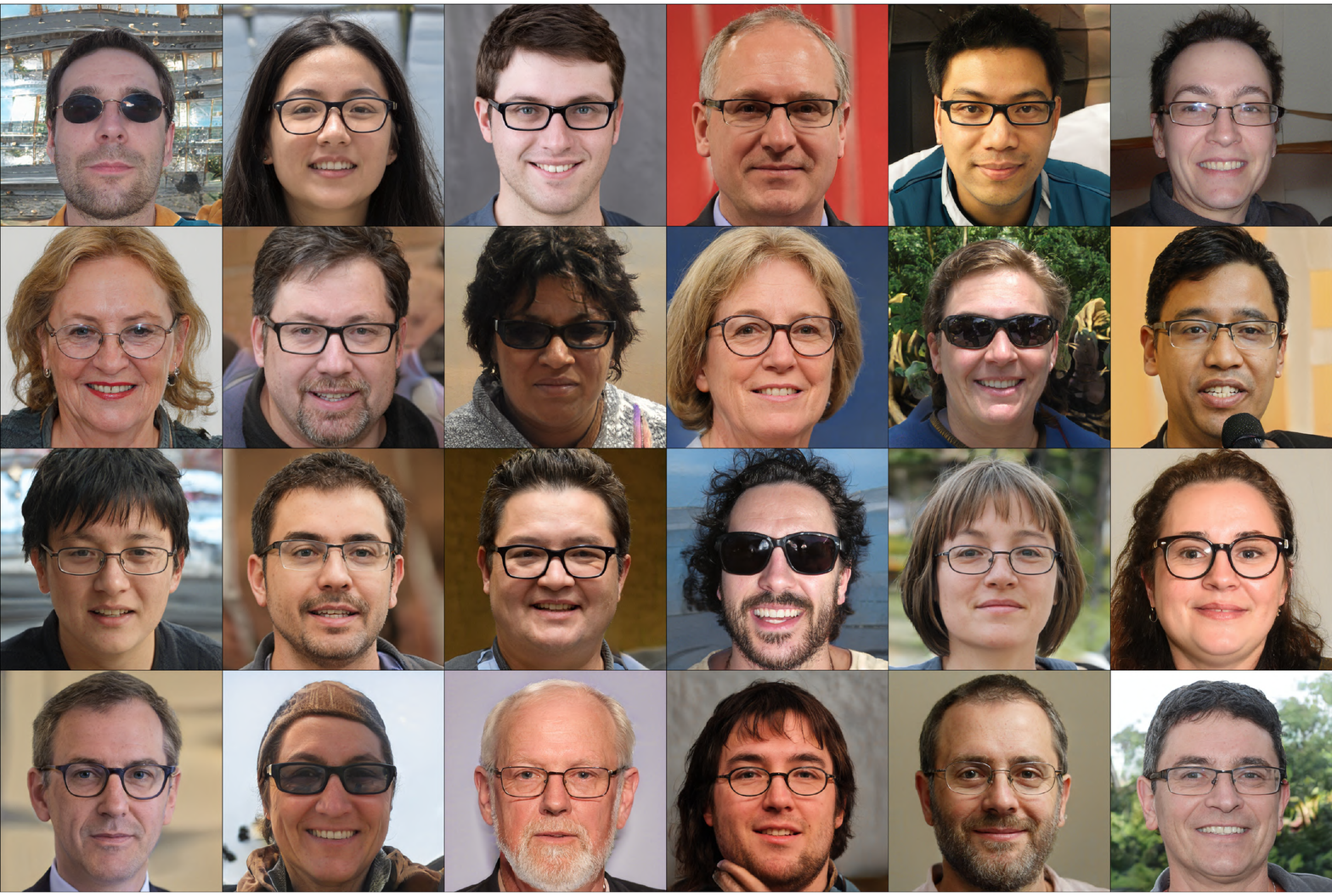}
    \caption{glasses=1 AND yaw=front}
\end{subfigure}

\begin{subfigure}[c]{\textwidth}
    \centering
        \includegraphics[width=\linewidth]{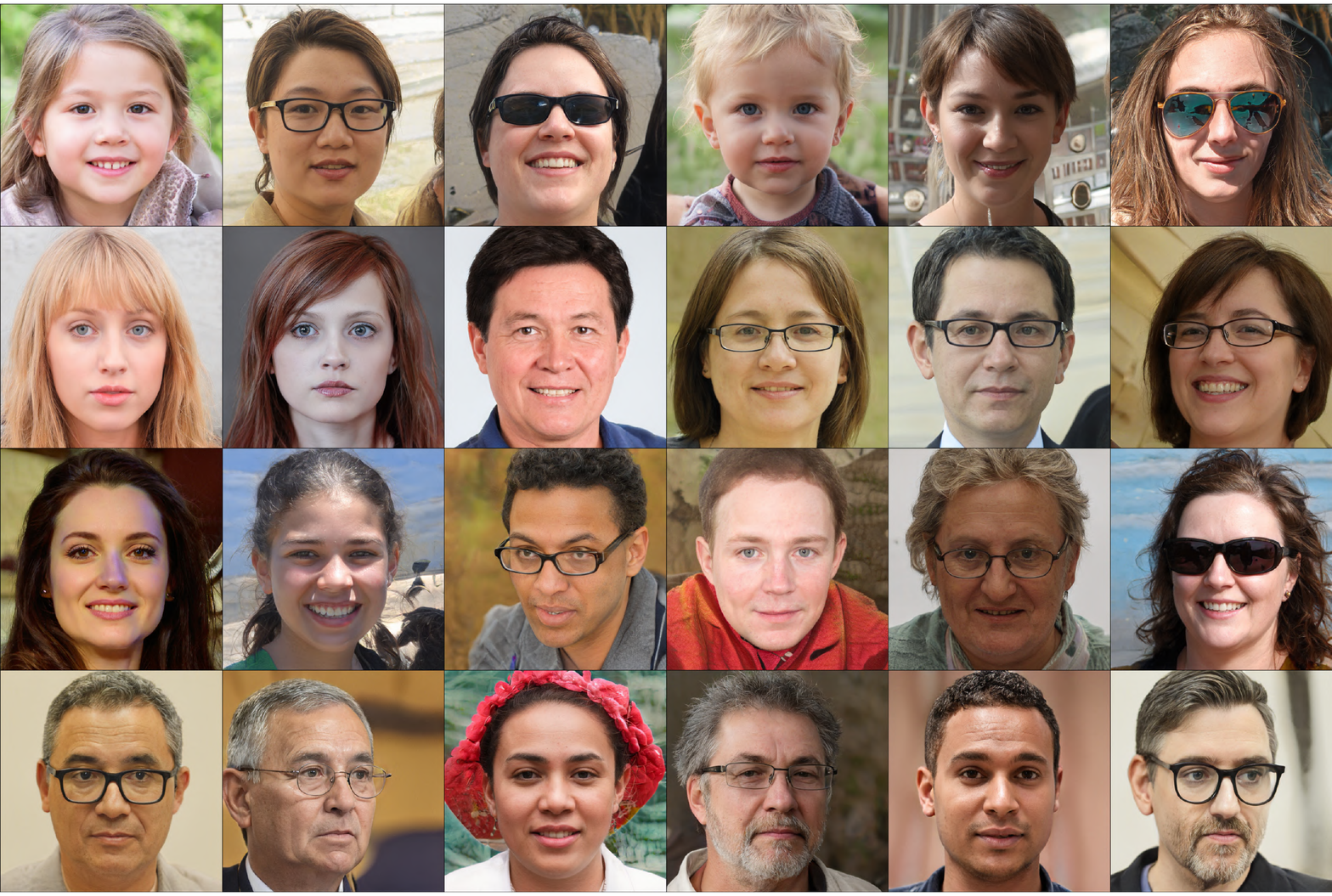}
    \caption{glasses=1 OR yaw=front}
\end{subfigure}
    
    \caption{\small Uncurated 1024$\times$1024 generation results of our method (LACE-ODE) in compositions of energy functions with different logical operations: (a) conjunction (AND), and (b) disjunction (OR). We can see that our method closely follows the rule of the given logical operators, and also produces diverse images that cover all possible logical cases. 
    }
    \label{app:ffhq_comp_energy_a}
\end{figure}

\begin{figure}[ht]
    \centering
\begin{subfigure}[c]{\textwidth}
    \centering
        \includegraphics[width=\linewidth]{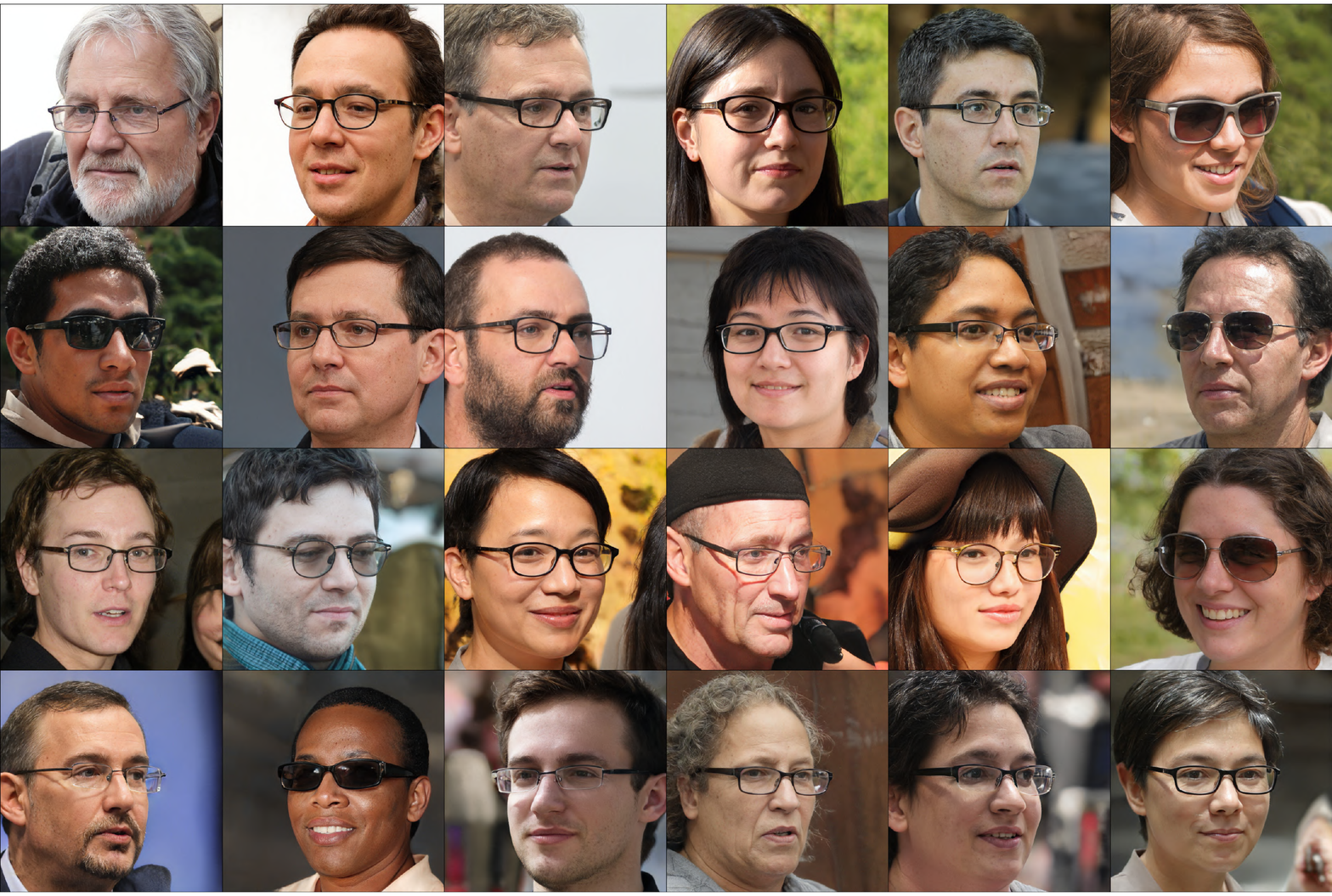}
    \caption{glasses=1 AND (NOT yaw=front)}
\end{subfigure}

\begin{subfigure}[c]{\textwidth}
    \centering
        \includegraphics[width=\linewidth]{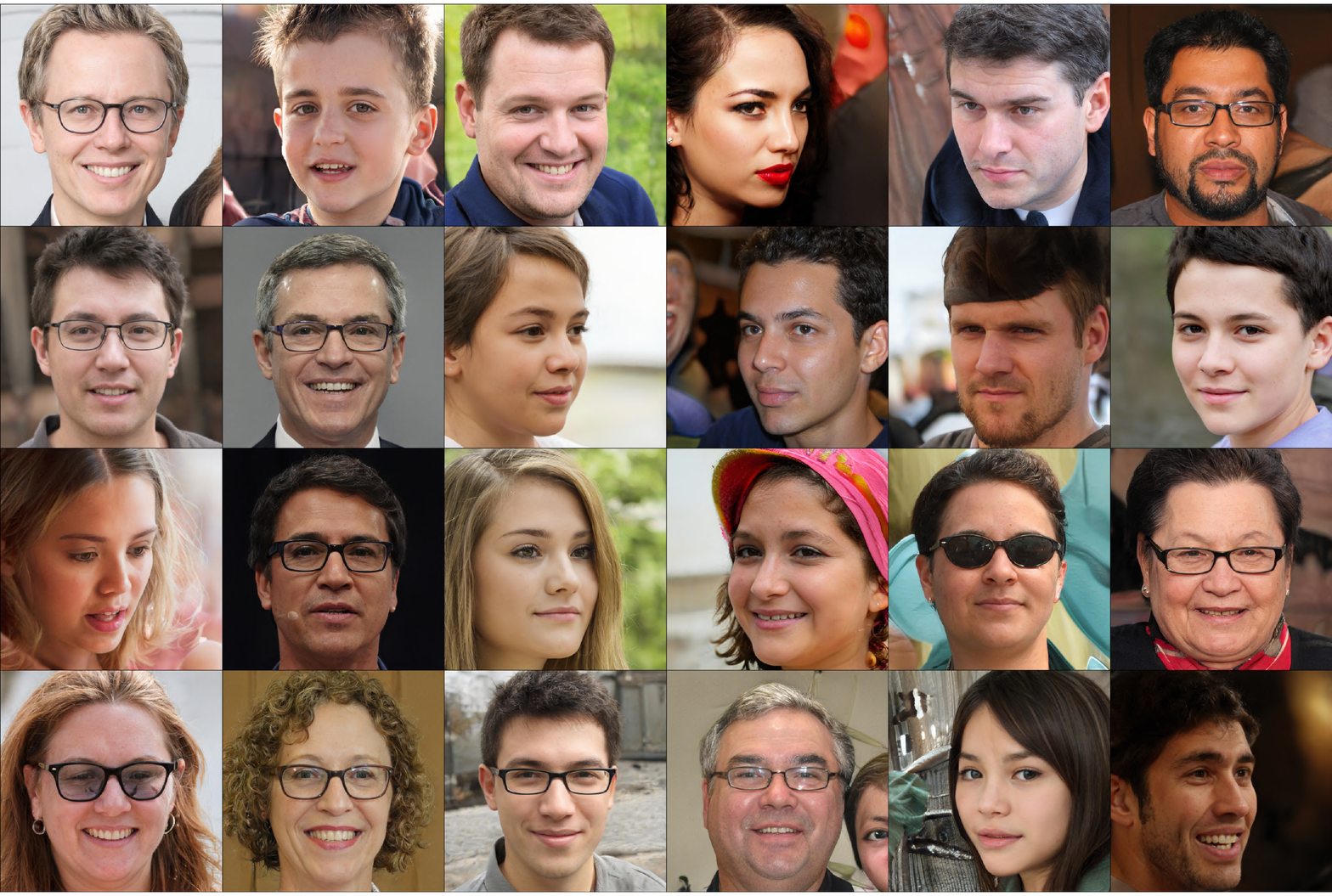}
    \caption{(glasses=1 AND yaw=front) OR (glasses=0 AND (NOT yaw=front))}
\end{subfigure}
    
    \caption{\small Uncurated 1024$\times$1024 generation results of our method (LACE-ODE) in compositions of energy functions with different logical operations: (a) negation (NOT), and (b) recursive combinations of logical operations on FFHQ. We can see that our method closely follows the rule of the given logical operators, and also produces diverse images that cover all possible logical cases. 
    }
    \label{app:ffhq_comp_energy_b}
\end{figure}

The 1024$\times$1024  visual samples of our method (LACE-ODE) in compositions of energy functions with different logical operations: conjunction (AND), disjunction (OR), negation (NOT), and their recursive combinations on FFHQ can be seen in Figure \ref{app:ffhq_comp_energy_a} and Figure \ref{app:ffhq_comp_energy_b}, respectively.


\vspace{-3pt}
\subsection{Continuous control on discrete attributes}

When the controlling attributes are discrete or binary, a smooth interpolation between discrete or binary attribute values could be a challenge of methods in controllable generation \citep{choi2018stargan}. For example, can we smoothly control the amount of beard in the generated images even if its provided ground-truth labels are binary (0: without beard, 1: with beard)? To this end, we can add a temperature variable $T$ to the energy function of discrete attributes defined in Eq. (\ref{e_cond_dis_cont}), which becomes
\begin{align*}
    E_\theta(c_i |x) = -\log \text{softmax} \left(\frac{ f_i(x;\theta)[c_i] }{T} \right)
    := - \frac{ f_i(x;\theta)[c_i] }{T}
    + \log\sum_{c_i}\exp{ \left( \frac{ f_i(x;\theta)[c_i] }{T} \right)} 
\end{align*}
where $c_i$ is a discrete attribute. 
Thus, the temperature $T \in (-\infty, 0)$ can be varied to adjust the impact of the attribute signal on the energy function. Its impact on the energy function becomes larger with a smaller value of $T$, resulting in a more significant visual appearance of the attribute value in the generated images, such as the increasing amount of beard on faces. 

\begin{figure}[ht]
    \centering
\begin{subfigure}[c]{\textwidth}
    \centering
        \includegraphics[width=\linewidth]{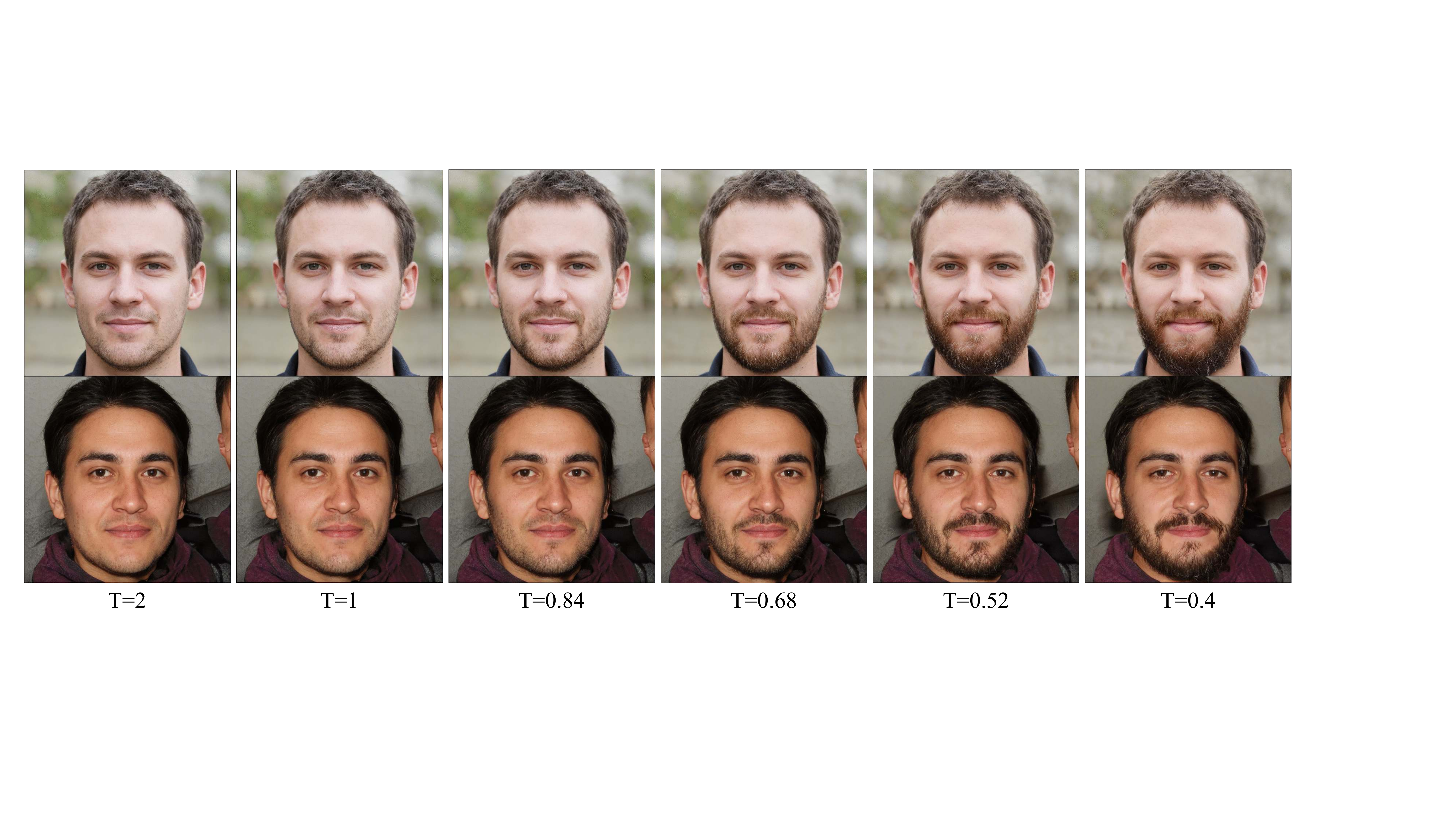}
        \caption{ \texttt{beard}=1 with the temperature $T$ varying from 2 to 0.4. }
\end{subfigure}
\begin{subfigure}[c]{\textwidth}
    \centering
        \includegraphics[width=\linewidth]{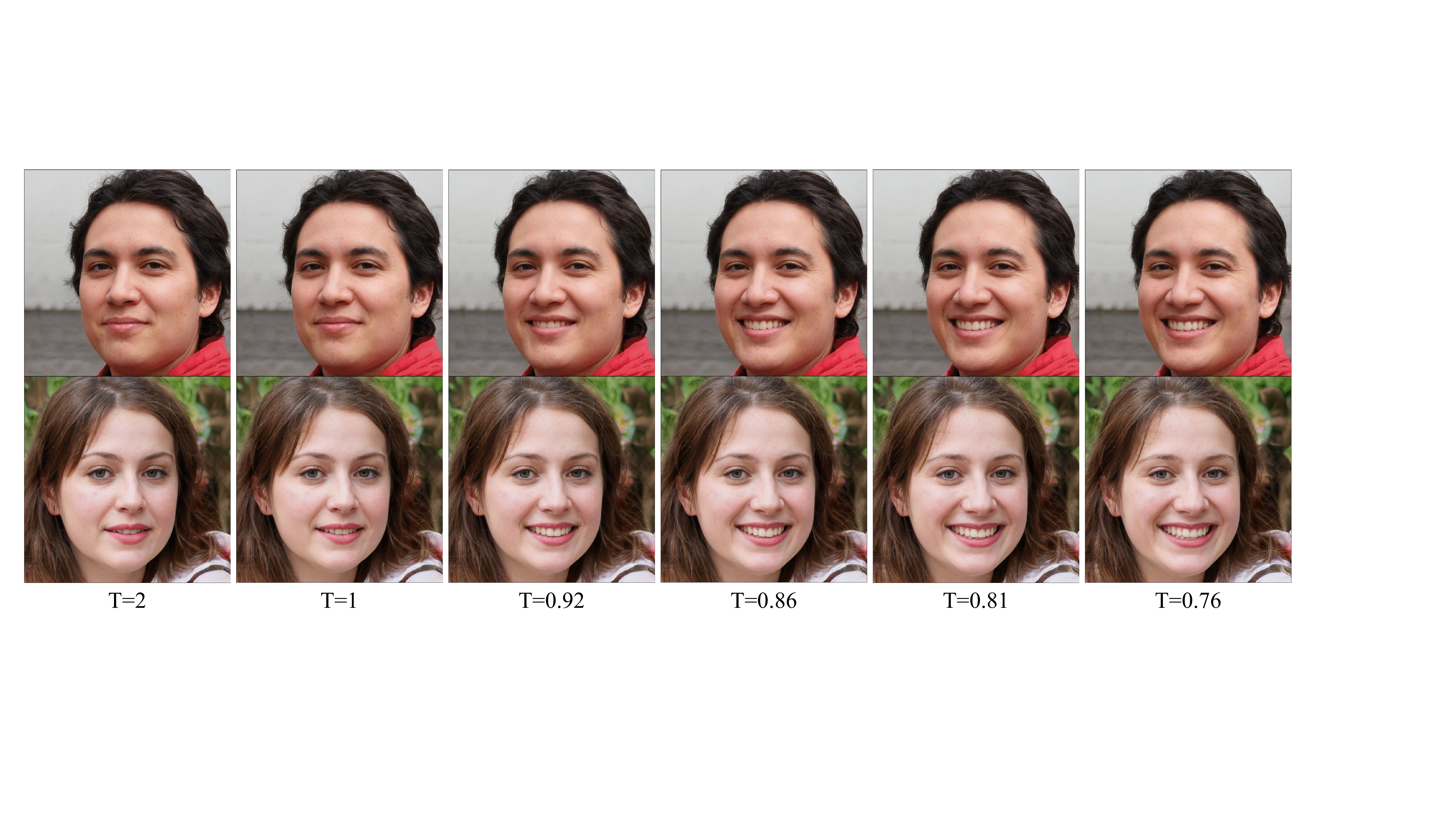}
        \caption{ \texttt{smile}=1 with the temperature $T$ varying from 2 to 0.76.}
\end{subfigure}
    
    \caption{\small Continuous control of discrete attributes (a) \texttt{beard} and (b) \texttt{smile} by varying the temperature $T$. Even if the ground-truth labels of the above discrete attributes are binary only, our method can learn to smoothly perform continuous control of them. As we decrease the temperature $T$, the visual appearance of the controlling attributes becomes more significant in the generated images.
    }
    \label{app:ffhq_temperature}
\end{figure}

In Figure \ref{app:ffhq_temperature}, we show the 1024$\times$1024 visual examples of continuous control on two binary attributes: (a) \texttt{beard} and (b) \texttt{smile} on the FFHQ data, where the visual appearance of both two attributes smoothly increases as we gradually decrease the temperature $T$.

\vspace{-6pt}
\subsection{More results of ODE sampling vs. LD sampling}
\label{app:ode_ld}

\subsubsection{Hyperparameter settings}

To compare the ODE and LD sampler more thoroughly, we perform a grid search in a large range of hyperparameter settings in each sampling method. In particular, The ODE sampler has two hyperparameters: (\texttt{atol}, \texttt{rtol}), which stand for the absolute and relative tolerances, respectively. The LD sampler has three hyperparameters: ($N$, $\eta$, $\sigma$), which denote the number of steps, step size and standard deviation of the noise in LD, respectively. 

For ODE, the grid search is performed with \texttt{atol} $\in$ [1e-1, 5e-2, 1e-2, 5e-3, 1e-3, 5e-4, 1e-4, 5e-5, 1e-5] and \texttt{rtol} $\in$ [1e-1, 5e-2, 1e-2, 5e-3, 1e-3, 5e-4, 1e-4, 5e-5, 1e-5]. Thus, there are 81 hyperparameter settings for the ODE sampler. 
For LD, the grid search is performed with $N \in [50, 100, 200, 300, 400, 500, 600, 1000]$, $\eta \in [0.1, 0.05, 0.01, 0.005, 0.001]$ and $\sigma \in [0.1, 0.05, \eta]$. Thus, there are 104 hyperparameter settings for the LD sampler.

\begin{figure}[ht]
  \centering
  \begin{subfigure}[c]{0.75\textwidth}
		\centering
		\includegraphics[width=\linewidth]{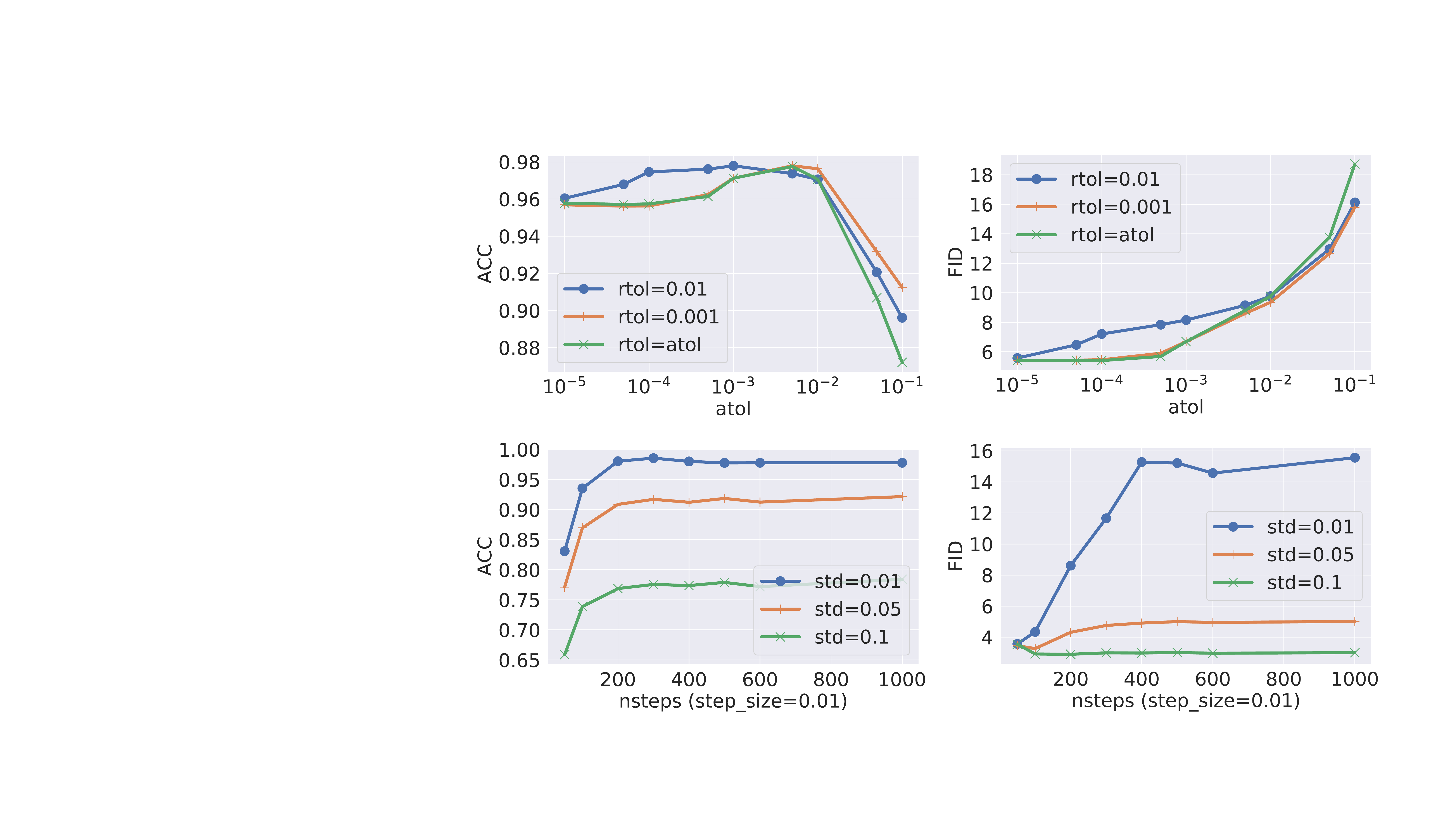}
  \end{subfigure}
  \vspace{2pt}
  \caption{\small The impact of hyperparameters (top: ODE, bottom: LD). We can see that we could always use small values of (\texttt{atol}, \texttt{rtol}) to get both good generation quality and controllability in the ODE sampler, which implies the hyperparameters in the ODE sampler are easy to tune. On the contrary, there exists a clear ACC-FID trade-off controlled by the standard deviation of the noise $\sigma$ and the number of steps $N$, which implies it tends to be more difficult to find the optimal hyperparameter setting for the LD sampler. }
  \vspace{-3mm}
  \label{app:which_sampler_hyper}
\end{figure}

\vspace{-3pt}
\subsubsection{Impact of each individual hyperparameter}

To dissect how sensitive the samplers are to each individual hyperparameter, Figure \ref{app:which_sampler_hyper} shows the impact of (\texttt{atol}, \texttt{rtol}) in the ODE sampler (top row) and the impact of ($N$, $\eta$, $\sigma$) in the LD sampler. We can see that with different \texttt{rtol} values, a smaller \texttt{atol} tends to have a higher ACC score (though it slightly decreases after $\texttt{atol} < 10^{-3}$) and a lower FID score. Hence, we could always use small values of (\texttt{atol}, \texttt{rtol}) to get both good generation quality and controllability, which implies the hyperparameters in the ODE sampler are easy to tune. 

In the LD sampler, however, there exists a clear ACC-FID trade-off controlled by the standard deviation of the noise $\sigma$: a smaller value of $\sigma$ results in a better ACC score but a worse FID score. Meanwhile, increasing the number of steps $N$ will also cause a better ACC score but a worse FID score when the value of $\sigma$ is small. Therefore, both values of $N$ and $\sigma$ in the LD sampler should not be too large or too small, and a sweet pot of these hyperparameters varies with different downstream tasks as we see in our experiments, which implies it tends to be more difficult to find the optimal hyperparameter setting for the LD sampler.

\subsubsection{Ablation on a simple Euler method}
\label{app:abl_euler}

By default, LACE-ODE applies the adaptive-step ``dopri5'' solver (i.e., Runge-Kutta of order 5) because of its adaptivity in step size for better efficiency.  But how does our ODE formulation in Eq. (\ref{ode_final}) work with a simple Euler discretization method?
To this end, we run our ODE sampler with the Euler method (called LACE-euler) on CIFAR-10, with an extra hyperparameter ``step\_size'' being set to 1e-2 or 1e-3. The results are shown in Table \ref{app:table_euler}.
We can see that 1) LACE-ODE with the default “dopri5” method is faster than LACE-euler (0.50s vs 0.68s) for getting similar performance, which confirms our intuition of adaptive step size vs. fixed step size, and 2) our method also works decently well with the Euler method, and its performance lies in-between that of LACE-LD and LACE-ODE.

\begin{table}[t]
    \centering
    \footnotesize
    \caption{Ablation results of LACE-euler (i.e., the Euler discretization method) with its hyperparameter ``step\_size'' being set to 1e-2 or 1e-3. For notations, Infer -- inference time (s: second), which refers to the single GPU time for generating a batch of 64 images.}\label{app:table_euler}
    \begin{tabular}{c|ccc}
        \hline
          Methods & Infer$\downarrow$ & FID$\downarrow$ & ACC$\uparrow$ \\
         \hline
          LACE-LD & 0.68s & \textbf{4.30} & 0.939
          \\
          LACE-euler (step\_size=1e-2) & 0.68s & {6.31} & 0.969
        \\
        LACE-euler (step\_size=1e-3) & 6.80s & {5.36} & 0.964
        \\
        LACE-ODE & \textbf{0.50s} & {6.63} & \textbf{0.972}
        \\
\hline
    \end{tabular}
    \vspace{-8pt}
\end{table}

\end{document}